\documentclass[10pt,twocolumn,letterpaper]{article}

\usepackage[pagenumbers]{cvpr}      % To produce the REVIEW version
\usepackage{graphicx}
\usepackage[dvipsnames]{xcolor}
\usepackage{amsmath}
\usepackage{amssymb}
\usepackage{booktabs}
\usepackage{overpic}
\usepackage{float}
\usepackage[linesnumbered, boxed, ruled]{algorithm2e}
\usepackage{multirow}
\usepackage{array}
\usepackage{soul}
\usepackage[pagebackref,breaklinks,colorlinks]{hyperref}

\usepackage[capitalize]{cleveref}
\crefname{section}{Sec.}{Secs.}
\Crefname{section}{Section}{Sections}
\Crefname{table}{Table}{Tables}
\crefname{table}{Tab.}{Tabs.}

\newcommand*{\affaddr}[1]{#1} 
\newcommand*{\affmark}[1][*]{\textsuperscript{#1}}

\usepackage[symbol]{footmisc}

\def\cvprPaperID{5647} % *** Enter the CVPR Paper ID here
\def\confName{CVPR}
\def\confYear{2023}

\begin{document}


% % \usepackage{hyperref}
% \usepackage{amsmath}
% \usepackage{graphicx}
% \usepackage{comment}
% %\usepackage{kbordermatrix}
% \usepackage{multirow,bigdelim}
% \usepackage{lipsum}
% \usepackage{array}
% \usepackage{wrapfig}
% \usepackage{booktabs}  
% % \usepackage{csvsimple}
% \usepackage{adjustbox}

% \usepackage{makecell}
% % \usepackage{float}
% \usepackage[normalem]{ulem}
% \usepackage{blindtext}
% \usepackage{xcolor}
% \usepackage{soul}
% \usepackage{cleveref}
% \usepackage{amsfonts}
% \usepackage{arydshln}
% \usepackage{listings}
% \usepackage[linesnumbered, boxed, ruled]{algorithm2e}
% \newcolumntype{C}[1]{>{\centering\let\newline\\\arraybackslash\hspace{0pt}}m{#1}}

\newcommand{\hlc}[2][color]{{\sethlcolor{#1}\hl{#2}}}

\newif\ifdraft
%\drafttrue
\draftfalse

\ifdraft
%Our comments:
\definecolor{darkg}{rgb}{0,0.4,0}
\newcommand{\dcc}[1]{{\color{red}[\textbf{DC:} #1]}}
\newcommand{\ahc}[1]{{\color{purple}[\textbf{AH:} #1]}}
\newcommand{\jtc}[1]{{\color{blue}[\textbf{JT:} #1]}}
\newcommand{\kac}[1]{{\color{teal}[\textbf{KA:} #1]}}
\newcommand{\ypc}[1]{{\color{violet}[\textbf{YP} #1]}}
\newcommand{\rmc}[1]{{\color{darkg}[\textbf{RM:} #1]}}

%Noticable new adds:
\newcommand{\dc}[1]{{\color{red}#1}}
\newcommand{\ah}[1]{{\color{purple}#1}}
\newcommand{\jt}[1]{{\color{blue}#1}}
\newcommand{\ka}[1]{{\color{teal}#1}}
\newcommand{\yp}[1]{{\color{violet}#1}}
\newcommand{\ron}[1]{{\color{darkg}#1}}

\newcommand{\rev}[1]{{\color{brown}#1}}

\newcommand{\drop}[1]{}

% removal candidates
% \newcommand{\nuke}[1]{{\color{brown}#1}}
% editing candidates
% \newcommand{\tune}[1]{{\hlc[pink]{#1}}}

% mark content changed for camera ready
% \newcommand{\cradd}[1]{{\color{blue}#1}}
% \newcommand{\crmv}[1]{{\color{red}#1}}

\newcommand{\cradd}[1]{{\color{black}#1}}
\newcommand{\crmv}[1]{{\color{black}#1}}

\else
\newcommand{\ahc}[1]{}
\newcommand{\dcc}[1]{}
\newcommand{\jtc}[1]{}
\newcommand{\kac}[1]{}
\newcommand{\ypc}[1]{}
\newcommand{\rmc}[1]{}
\newcommand{\dc}[1]{{\color{black}#1}}
\newcommand{\ah}[1]{{\color{black}#1}}
\newcommand{\jt}[1]{{\color{black}#1}}
\newcommand{\ka}[1]{{\color{black}#1}}
\newcommand{\yp}[1]{{\color{black}#1}}
\newcommand{\ron}[1]{{\color{black}#1}}
\fi

\newcommand{\Mi}{\hat{M}_{l-1}}
\newcommand{\Mo}{\hat{M}_{l}}
\newcommand{\dVi}{\Delta \hat{V}_{l-1}}
\newcommand{\dVo}{\Delta \hat{V}_{l}}
\newcommand{\Vi}{\hat{V}_{l-1}}
\newcommand{\Vo}{\hat{V}_{l}}
\newcommand{\dEo}{\Delta \hat{E}_{l}}
\newcommand{\Ci}{\hat{C}_{l-1}}
\newcommand{\Wo}{W_{l}}
\newcommand{\partmesh}{PartMesh}
\newcommand{\eps}{\varepsilon}
\newcommand{\norm}[1]{\left\Vert #1 \right\Vert_2}
\newcommand{\localalpha}{\alpha}
\newcommand{\localalphaf}{B}

\newcommand{\ourmethod}{StyleGAN-NADA}

\def\naive{na\"{\i}ve\xspace}
\def\Naive{Na\"{\i}ve\xspace}

\newcommand{\w}{$\mathcal{W}$\xspace}
\newcommand{\wplus}{$\mathcal{W}+$\xspace}

\makeatletter
\DeclareRobustCommand\onedot{\futurelet\@let@token\@onedot}
\def\@onedot{\ifx\@let@token.\else.\null\fi\xspace}

\def\eg{\emph{e.g}\onedot}
\def\Eg{\emph{E.g}\onedot}
\def\ie{\emph{i.e}\onedot}
\def\Ie{\emph{I.e}\onedot}
\def\etc{\emph{etc}\onedot}
\def\etal{\emph{et al}\onedot}
\makeatother
% % for tikz package
% \usepackage{pgfplots}
% \pgfplotsset{compat=newest}
% \usepgfplotslibrary{groupplots}
% \usepgfplotslibrary{dateplot}
% % end tikz package

% \usepackage[bottom]{footmisc}
\raggedbottom

\makeatletter
\def\blfootnote{\xdef\@thefnmark{}\@footnotetext}
\makeatother

% licenses
\newcommand{\ccbync}{\href{https://creativecommons.org/licenses/by-nc/4.0/legalcode}{CC BY-NC 4.0}}

\newcommand{\cczero}{\href{https://creativecommons.org/publicdomain/zero/1.0/}{CC0 1.0}}

\newcommand{\ccbyncsa}{\href{https://creativecommons.org/licenses/by-nc-sa/4.0/}{CC BY-NC-SA 4.0}}

\newcommand{\nvsrc}{\href{https://nvlabs.github.io/stylegan2/license.html
}{Nvidia Source Code License-NC}}

\newcommand{\bsd}{\href{https://opensource.org/licenses/BSD-3-Clause}{BSD 3-Clause}}

\newcommand{\mitlic}{\href{https://opensource.org/licenses/MIT}{MIT License}}

\newcommand{\adblic}{\href{https://github.com/utkarshojha/few-shot-gan-adaptation/blob/main/LICENSE.txt}{Adobe Research License}}

%\DeclareMathOperator*{\argmin}{arg\,min}

%real numbers
\newcommand{\naturals}{\mathbb{N}}
\newcommand{\reals}{\mathbb{R}}
\newcommand{\attmask}{M}
\newcommand{\pixelmod}{i}
\newcommand{\textmod}{p}
\newcommand{\textemb}{\mathcal{C}}
\newif\ifwatermark
\watermarktrue
\draftfalse

\title{Null-text Inversion for Editing Real Images using Guided Diffusion Models}

% \author[1,2]{Ron Mokady\footnote{}~~}
% \author[1,2]{Amir Hertz\samethanks ~~}
% \author[1]{Jay Tenenbaum}
% \author[1]{Kfir Aberman}
% \author[1]{Yael Pritch}
% \author[1,2]{Daniel Cohen-Or\samethanks ~~}
% \affil[1]{ Google Research}
% \affil[2]{The Blavatnik School of Computer Science, Tel Aviv University}

\author{%
Ron Mokady\footnote{}~~\footnote{}~~\affmark[1,]\affmark[2], Amir Hertz\footref{note1}~~\footref{note2}~~\affmark[1,]\affmark[2], Kfir Aberman\affmark[1], Yael Pritch\affmark[1], and Daniel Cohen-Or\footref{note2}~~\affmark[1,]\affmark[2]\\
\small{\affaddr{\affmark[1]Google Research,~~}}\small{\affaddr{\affmark[2]The Blavatnik School of Computer Science, Tel Aviv University}}\
}

%%%%%%%%% ABSTRACT

\twocolumn[{
\renewcommand\twocolumn[1][]{#1}

\vspace{-1.5cm}
\maketitle
%\vspace{-0.6cm}
\begin{center}
    \centering
    \vspace{-0.25cm}
    \includegraphics[width=\textwidth]{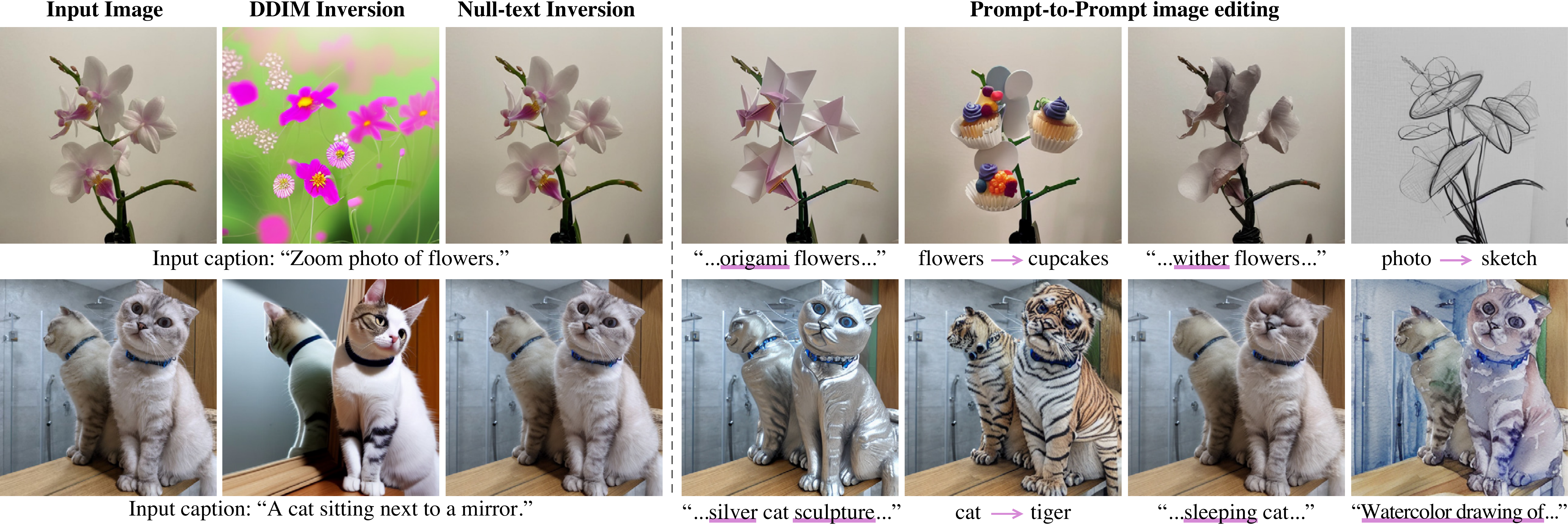}
    \vspace{-0.7cm}
    \captionof{figure}{{\bf Null-text inversion for real image editing.} {\it Our method takes as input a real image (leftmost column) and an associated caption. The image is inverted with a DDIM diffusion model to yield a diffusion trajectory (second column to the left). Once inverted, we use the initial trajectory as a pivot for null-text optimization that accurately reconstructs the input image (third column to the left). Then, we can edit the inverted image by modifying only the input caption using the editing technique of Prompt-to-Prompt \cite{hertz2022prompt} .} % \ypc{using "prompt to prompt"} (rightmost columns).} %\kac{the right most title "prompt-to-prompt Image Editing" should be centered at the middle of the right part of the figure (exactly in the middle of the second and third images) it is skewed to the left at the moment} \dcc{make a huge effort to fit entirely the abstract on the left column.} 
    }
    %\vspace{-0.2cm}
\label{fig:teaser}
\end{center}
}]

\vspace{-0.5cm}
\begin{abstract}
\vspace{-0.2cm}

Recent text-guided diffusion models provide powerful image generation capabilities. Currently, a massive effort is given to enable the modification of these images using text only as means to offer intuitive and versatile editing.
To edit a real image using these state-of-the-art tools, one must first invert the image with a meaningful text prompt into the pretrained model's domain. 
In this paper, we introduce an accurate inversion technique and thus facilitate an intuitive text-based modification of the image.
Our proposed inversion consists of two novel key components: 
(i)~\emph{Pivotal inversion for diffusion models.} While current methods aim at mapping random noise samples to a single input image, we use a single pivotal noise vector for each timestamp and optimize around it. We demonstrate that a direct inversion is inadequate on its own, but does provide a good anchor for our optimization.
(ii)~\emph{null-text optimization,} where we only modify the unconditional textual embedding that is used for classifier-free guidance, rather than the input text embedding. This allows for keeping both the model weights and the conditional embedding intact and hence enables applying prompt-based editing while avoiding the cumbersome tuning of the model's weights.
Our null-text inversion, based on the publicly available Stable Diffusion model, is extensively evaluated on a variety of images and prompt editing, showing high-fidelity editing of real images.

\end{abstract}

\footnotetext[1]{\label{note1} Equal contribution.}
\footnotetext[2]{\label{note2}Performed this work while working at Google.}

\vspace{-0.2cm}
\section{Introduction}
%\vspace{-0.1cm}

The progress in image synthesis using text-guided diffusion models has attracted much attention due to their exceptional realism and diversity. Large-scale models \cite{ramesh2022hierarchical,saharia2022photorealistic,rombach2021highresolution} have ignited the imagination of multitudes of users, enabling image generation with unprecedented creative freedom. Naturally, this has initiated ongoing research efforts, investigating how to harness these powerful models for image editing. Most recently, intuitive text-based editing was demonstrated over synthesized images, allowing the user to easily manipulate an image using text only \cite{hertz2022prompt}.

However, text-guided editing of a real image with these state-of-the-art tools requires \textit{inverting} the given image and textual prompt. That is, finding an initial noise vector that produces the input image when fed with the prompt into the diffusion process while preserving the editing capabilities of the model. The inversion process has recently drawn considerable attention for GANs \cite{xia2021gan,bermano2022state}, but has not yet been fully addressed for text-guided diffusion models. Although an effective DDIM inversion \cite{dhariwal2021diffusion,song2020denoising} scheme was suggested for unconditional diffusion models, it is found lacking for text-guided diffusion models when classifier-free guidance \cite{ho2021classifier}, which is necessary for meaningful editing, is applied.

In this paper, we introduce an effective inversion scheme, achieving near-perfect reconstruction, while retaining the rich text-guided editing capabilities of the original model (see \cref{fig:teaser}). Our approach is built upon the analysis of two key aspects of guided diffusion models: classifier-free guidance and DDIM inversion.

In the widely used classifier-free guidance, in each diffusion step, the prediction is performed twice: once unconditionally and once with the text condition. These predictions are then extrapolated to amplify the effect of the text guidance. While all works concentrate on the conditional prediction, we recognize the substantial effect induced by the unconditional part. Hence, we optimize the embedding used in the unconditional part in order to invert the input image and prompt. We refer to it as \emph{null-text optimization}, as we replace the embedding of the empty text string with our optimized embedding.

DDIM Inversion consists of performing DDIM sampling in reverse order. Although a slight error is introduced in each step, this works well in the unconditional case. However, in practice, it breaks for text-guided synthesis, since classifier-free guidance magnifies its accumulated error. We observe that it can still offer a  promising starting point for the inversion. Inspired by GAN literature, we use the sequence of noised latent codes, obtained from an initial DDIM inversion, as pivot \cite{roich2021pivotal}. We then perform our optimization around this pivot to yield an improved and more accurate inversion. We refer to this highly efficient optimization as \emph{Diffusion Pivotal Inversion}, which stands in contrast to existing works that aim to map all possible noise vectors to a single image.

To the best of our knowledge, our approach is the first to enable the text editing technique of Prompt-to-Prompt \cite{hertz2022prompt} on real images. %This virtue is due to the keeping of the conditional embedding intact. 
Moreover, unlike recent approaches \cite{Kawar2022ImagicTR, valevski2022unitune}, we do not tune the model weights, thus avoiding damaging the prior of the trained model and duplicating the entire model for each image. Throughout comprehensive ablation study and comparisons, we demonstrate the contribution of our key components to achieving a high-fidelity reconstruction of the given real image, while allowing meaningful and intuitive editing abilities. %\ypc{We plan to release our code, which is built..} 
For our code, built upon the publicly available Stable Diffusion model, please visit our project page \url{https://null-text-inversion.github.io/}.

%\vspace{-0.1cm}
\section{Related Work}
%\vspace{-0.1cm}

 Large-scale diffusion models, such as Imagen~\cite{saharia2022photorealistic}, DALL-E 2~\cite{ramesh2022hierarchical}, and Stable Diffusion~\cite{rombach2021highresolution}, have recently raised the bar for the task of generating images conditioned on plain text, known as text-to-image synthesis. Exploiting the powerful architecture of diffusion models \cite{ho2020denoising,sohl2015deep,song2019generative,ho2020denoising,song2020denoising,rombach2021highresolution}, these models can generate practically any image by simply feeding a corresponding text, and so have changed the landscape of artistic applications.

\begin{figure}
\centering 
\includegraphics[width=\columnwidth]{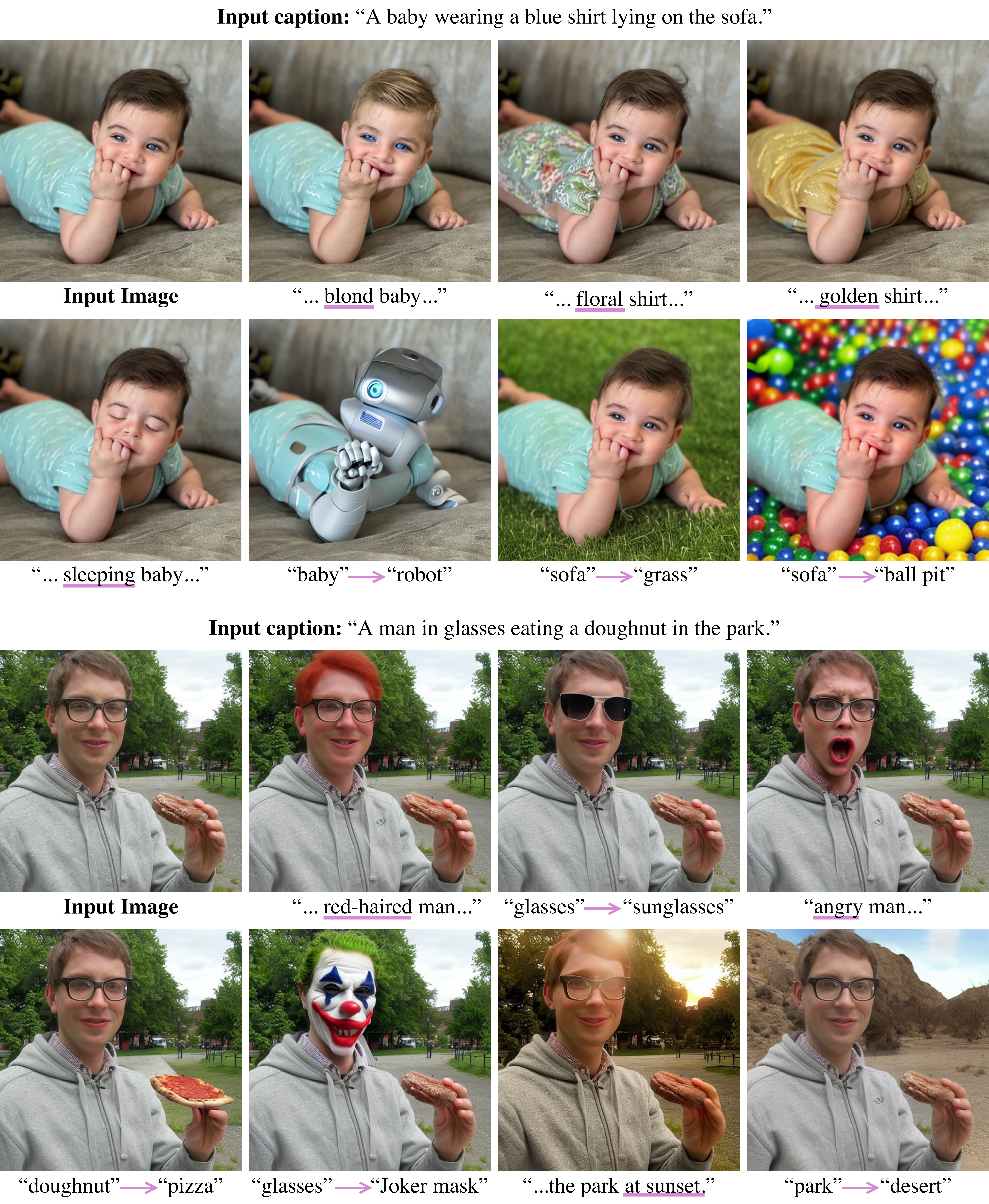} 

\vspace{-0.2cm}
\caption{{\bf Real image editing using our method.} {\it We first apply a single null-text inversion over the real input image, achieving high-fidelity reconstruction. Then, various Prompt-to-Prompt text-based editing operations are applied. As can be seen, our inversion scheme provides high fidelity while retaining high editability. See additional examples in \cref{sec:supp_results} (\cref{fig:supp_ours}). } } 
\vspace{-0.25cm}
\label{fig:exp_edit_people} 

\end{figure}
However, synthesizing very specific or personal objects which are not widespread in the training data has been challenging. 
This requires an \textit{inversion} process that given input images would enable regenerating the depicted object using a text-guided diffusion model. Inversion has been studied extensively for GANs \cite{zhu2016generative,lipton2017precise,creswell2018inverting, yeh2017semantic, xia2021gan, bermano2022state}, ranging from latent-based optimization \cite{abdal2019image2stylegan,abdal2020image2stylegan++} and encoders \cite{richardson2020encoding,tov2021designing} to feature space encoders \cite{Wang2021HighFidelityGI} and fine-tuning of the model \cite{roich2021pivotal, alaluf2021hyperstyle}. Motivated by this, Gal et al.~\cite{gal2022image} suggest a textual inversion scheme for diffusion models that enables regenerating a user-provided concept out of $3-5$ images. Concurrently, Ruiz et al.\cite{ruiz2022dreambooth} tackled the same task with model-tuning. However, these works struggle to edit a given real image while accurately reproducing the unedited parts.

Naturally, recent works have attempted to adapt text-guided diffusion models to the fundamental challenge of single-image editing, aiming to exploit their rich and diverse semantic knowledge.
Meng et al.~\cite{meng2021sdedit} add noise to the input image and then perform a text-guided denoising process from a predefined step. Yet, they struggle to accurately preserve the input image details. To overcome this, several works~\cite{nichol2021glide, avrahami2022blended, avrahami2022blendedlatent} assume that the user provides a mask to restrict the region in which the changes are applied, achieving both meaningful editing and background preservation.

%Nevertheless, performing accurate editing using a mask is less intuitive as the user is confined to providing a precise mask. 
However, requiring that users provide a precise mask is burdensome. Furthermore, masking the image content removes important information, which is mostly ignored in the inpainting process. 
While some text-only editing approaches are bound to global editing  \cite{crowson2022vqgan, kwon2021clipstyler, kim2022diffusionclip}, Bar-Tal et al.~\cite{bar2022text2live} propose a text-based localized editing technique without using any mask. Their technique allows high-quality texture editing, but not modifying complex structures, since only CLIP \cite{radford2021learning} is employed as guidance instead of a generative diffusion model.

 Hertz et al.~\cite{hertz2022prompt} suggest an intuitive editing technique, called Prompt-to-Prompt, of manipulating local or global details by modifying only the text prompt when using text-guided diffusion models. By injecting internal cross-attention maps, they preserve the spatial layout and geometry which enable the regeneration of an image while modifying it through prompt editing. Still, without an inversion technique, their approach is limited to synthesized images. Sheynin et al.~\cite{sheynin2022knn} suggest training the model for local editing without the inversion requirement, but their expressiveness and quality are inferior compared to current large-scale diffusion models. 
 Concurrent to our work, DiffEdit \cite{couairon2022diffedit} uses DDIM inversion for image editing, but avoids the emerged distortion by automatically producing a mask that allows background preservation.

Also concurrent, Imagic \cite{Kawar2022ImagicTR} and UniTune\cite{valevski2022unitune} have demonstrated impressive editing results using the powerful Imagen model \cite{saharia2022photorealistic}. Yet, they both require the restrictive fine-tuning of the model. Moreover, Imagic requires a new tuning for each editing, while UniTune involves a parameter search for each image. Our method enables us to apply the text-only intuitive editing of Prompt-to-Prompt \cite{hertz2022prompt} on real images. We do not require any fine-tuning and provide highly-quality local and global modifications using the publicly available Stable Diffusion \cite{rombach2021highresolution} model.

% \vspace{-0.1cm}
\section{Method}
%\vspace{-0.1cm}
% \begin{figure}[t]
% \centering 
% \scriptsize{}
% \begin{overpic}[width=\columnwidth]{figures/null_diagram_mini.pdf} 
% \put(83,23.8){Input Image}
% \put(77.7,1){Output Inversion}

% \put(35,40){DDIM Inversion}

% \put(30.5,11){Null-text Optimization}

% \put(76,37){$z_{0}$}
% \put(59,37){$z_{1}$}
% \put(41,37){$z_{2}$}
% \put(19,37){$z_{T-1}$}
% \put(3,37){$z_{T}$}
% {\color{VioletRed}
% \put(25.,30.8){$\varnothing_{T}$}
% \put(37.5,29.3){$\varnothing_{3}$}
% \put(54.5,27.1){$\varnothing_{2}$}
% \put(72.8,25.3){$\varnothing_1$}
% }
% \put(73.9,13){$\bar{z}_{0}$}
% \put(57.2,17.6){$\bar{z}_{1}$}
% \put(39.8,22.){$\bar{z}_{2}$}
% \put(18.5,27.4){$\bar{z}_{T-1}$}

% % \put(46.2,41.3){\rotatebox{50}{Attention}}

% \end{overpic}
% \vspace{-0.3cm}
% \caption{Null-text Inversion overview. Given an input image and a caption, we first apply a DDIM inversion which estimates an initial diffusion trajectory $\{z_t\}_{0}^{T}$ (Top). Afterwards, we use the initial trajectory as a pivot for a Null-text optimization (bottom) which brings the diffusion backwards trajectory $\{\bar{z}_t\}_{1}^{T}$ closer to the initial pivots by optimizing the Null-text embeddings  $\{\varnothing_t\}_{1}^{T}$.} 
% \label{fig:method_diagram} 
% \vspace{-0.3cm}
% \end{figure}

\begin{figure}[t]
\centering 
\scriptsize{}
\begin{overpic}[width=\columnwidth]{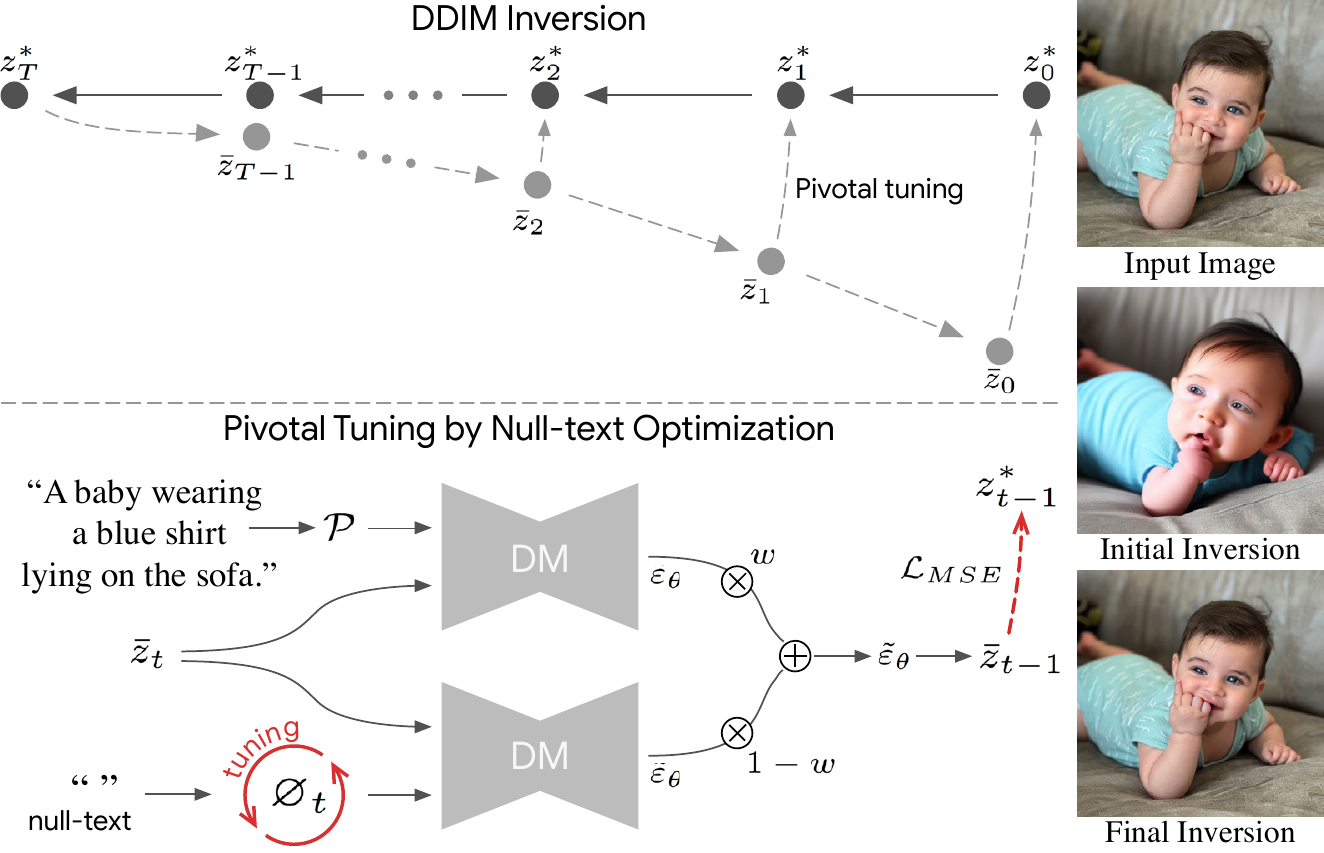} 
% \put(83,23.8){Input Image}
% \put(77.7,1){Output Inversion}

% \put(35,40){DDIM Inversion}

% \put(30.5,11){Null-text Optimization}

% \put(76,37){$z_{0}$}
% \put(59,37){$z_{1}$}
% \put(41,37){$z_{2}$}
% \put(19,37){$z_{T-1}$}
% \put(3,37){$z_{T}$}
% {\color{VioletRed}
% \put(25.,30.8){$\varnothing_{T}$}
% \put(37.5,29.3){$\varnothing_{3}$}
% \put(54.5,27.1){$\varnothing_{2}$}
% \put(72.8,25.3){$\varnothing_1$}
% }
% \put(73.9,13){$\bar{z}_{0}$}
% \put(57.2,17.6){$\bar{z}_{1}$}
% \put(39.8,22.){$\bar{z}_{2}$}
% \put(18.5,27.4){$\bar{z}_{T-1}$}

% \put(46.2,41.3){\rotatebox{50}{Attention}}

\end{overpic}
\vspace{-0.3cm}
\caption{{\bf Null-text Inversion overview.} {\it Top: pivotal inversion. We first apply an initial DDIM inversion on the input image which estimates a diffusion trajectory $\{z_t^*\}_{0}^{T}$. Starting the diffusion process from the last latent $z_T^*$ results in unsatisfying reconstruction as the latent codes become farther away from the original trajectory. We use the initial trajectory as a pivot for our optimization which brings the diffusion backward trajectory $\{\bar{z}_t\}_{1}^{T}$ closer to the original image encoding $z^*_0$.
%The deviation from the original trajectory comes from using classifier-free guidance, where text-conditioned and unconditional predictions are extrapolated.
Bottom: null-text optimization for timestamp $t$. Recall that classifier-free guidance consists of performing the prediction $\epsilon_\theta$ twice -- using text condition embedding and unconditionally using null-text embedding $\varnothing$ (bottom-left). Then, these are extrapolated with guidance scale $w$ (middle). 
We optimize only the unconditional embeddings $\varnothing_t$ by employing a reconstruction MSE loss (in red) between the predicated latent code $z_{t-1}$ to the pivot  $z_{t-1}^*$}.}
% For each timestamp $t$, we optimize only the unconditional embeddings $\varnothing_t$ by replacing the default null-text embeddings with our optimized one. The optimization is guided to reconstruct the pivot latent code $z_{t-1}^*$ using the MSE loss function.}} 
\label{fig:method_diagram} 
\vspace{-0.2cm}
\end{figure}

%%%% OLD CAPTION:
% \bf Null-text Inversion overview.} {\it Given an input image and a caption, we first apply a DDIM inversion which estimates an initial diffusion trajectory $\{z_t^*\}_{0}^{T}$ (Top). Starting the diffusion process from the result of the DDIM inversion $z_T^*$, results in unsatisfying reconstruction as the latent codes become farther away from this trajectory as the diffusion progress. Therefore, we use the initial trajectory as a pivot for our optimization which brings the diffusion backward trajectory $\{\bar{z}_t\}_{1}^{T}$ closer to the initial pivots.
% The deviation from the original trajectory comes from using classifier-free guidance, where text-conditioned and unconditional predictions are extrapolated.
% Therefore, instead of optimizing the textual embedding or the model's weights, we optimize the unconditional embeddings $\{\varnothing_t\}_{1}^{T}$ by replacing the default Null-text embeddings.

Let $\mathcal{I}$ be a real image. Our goal is to edit $\mathcal{I}$, using only text guidance, to get an edited image $\mathcal{I}^*$. We use the setting defined by Prompt-to-Prompt \cite{hertz2022prompt}, where the editing is guided by source prompt $\mathcal{P}$ and edited prompt $\mathcal{P}^*$.
This requires the user to provide a source prompt. Yet, we found that automatically producing the source prompt using an off-the-shelf captioning model \cite{mokady2021clipcap} works well (see \cref{sec:ablation}). 
%For example, see \cref{fig:teaser}, where given a cat image and a source prompt "A cat sitting...", we can replace the cat with a tiger with the edited prompt "A tiger sitting...". 
For example, see \cref{fig:exp_edit_people}, given an image and a source prompt "A baby wearing...", we replace the baby with a robot by providing the edited prompt "A robot wearing...". 

Such editing operations first require inverting $\mathcal{I}$ to the model's output domain. Namely, the main challenge is faithfully reconstructing $\mathcal{I}$ by feeding the source prompt $\mathcal{P}$ to the model, while still retaining the intuitive text-based editing abilities.

Our approach is based on two main observations. 
First, DDIM inversion produces unsatisfying reconstruction when classifier-free guidance is applied, but provides a good starting point for the optimization, enabling us to efficiently achieve high-fidelity inversion. Second, optimizing the unconditional null embedding, which is used in classifier-free guidance, allows an accurate reconstruction while avoiding the tuning of the model and the conditional embedding. Thereby preserving the desired editing capabilities.

Next, we provide a short background, followed by a detailed description of our approach in \cref{sec:pivotal} and \cref{sec:null}. A general overview is provided in \cref{fig:method_diagram}.

\begin{figure*}[t]
\centering 
\vspace{-0.5cm}
\includegraphics[width=\textwidth]{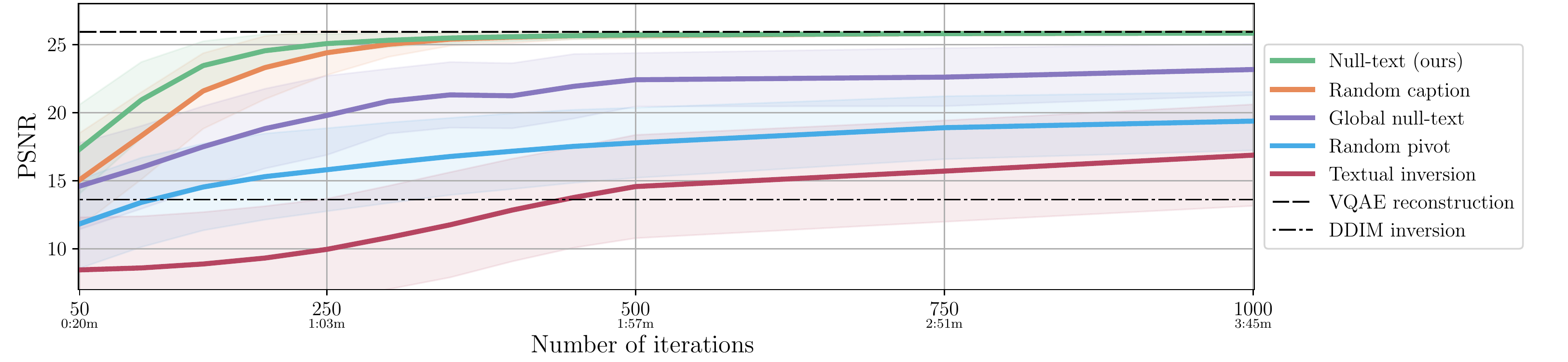} 
% \vspace{0.25cm}
\\[4pt]
\includegraphics[width=\textwidth]{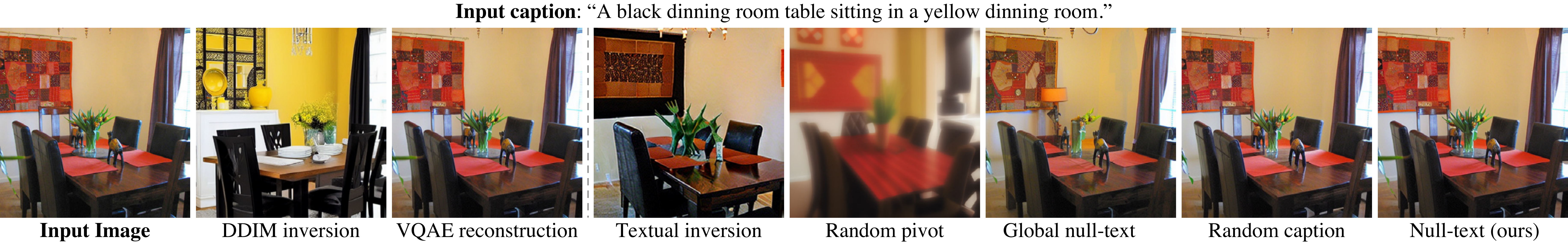} 
\vspace{-0.5cm}
\caption{{ \bf Ablation Study.} { \it Top: we compare the performance of our full algorithm (green line) to different variations, evaluating the reconstruction quality by measuring the PSNR score as a function of number optimization iterations and running time in minutes. Bottom: we visually show the inversion results after $200$ iterations of our full algorithm (on right) compared to other baselines. Results for all iterations are shown in \cref{sec:ablation_supp} (\cref{fig:exp_ablation_qual_sup,fig:exp_ablation_qual_sup2}).}} 
\label{fig:exp_graph} 
\vspace{-0.25cm}
\end{figure*}

%\input{figures/comparison_ablation}

% \vspace{-0.1cm}
\subsection{Background and Preliminaries}
\label{sec:background}
%\vspace{-0.1cm}

Text-guided diffusion models aim to map a random noise vector $z_t$ and textual condition $\mathcal{P}$ to an output image $z_0$, which corresponds to the given conditioning prompt.
In order to perform sequential denoising, the network $\eps_\theta$ is trained to predict artificial noise, following the objective: \\[-8pt]
\begin{equation}
\min_\theta E_{z_0,\eps\sim N(0,I),t\sim \text{Uniform}(1,T)} \norm{\eps-\eps_\theta(z_t,t,\textemb)}^2.
\end{equation}
\\[-12pt]
Note that $\textemb = \psi(\mathcal{P})$ is the embedding of the text condition and $z_t$ is a noised sample, where noise is added to the sampled data $z_0$ according to timestamp $t$.
At inference, given a noise vector $z_T$, The noise is gradually removed by sequentially predicting it using our trained network for $T$ steps. 

Since we aim to accurately reconstruct a given real image, we employ the deterministic DDIM sampling \cite{song2020denoising}: \\[-8pt]
$$
z_{t-1} = \sqrt{\frac{\alpha_{t-1}}{\alpha_t}}z_t + \left(\sqrt{\frac{1}{\alpha_{t-1}}-1}-\sqrt{\frac{1}{\alpha_t}-1}\right) \cdot \eps_\theta(z_t,t,\textemb).
$$
\\[-8pt]
For the definition of $\alpha_t$ and additional details, please refer to \cref{sec:supp_background}. Diffusion models often operate in the image pixel space where $z_0$ is a sample of a real image. In our case, we use the popular and publicly available Stable Diffusion model \cite{rombach2021highresolution} where the diffusion forward process is applied on a latent image encoding $z_0 = E(x_0)$ and an image decoder is employed at the end of the diffusion backward process $x_0 = D(z_0)$.

\vspace{-0.2cm}
\paragraph{Classifier-free guidance.}
One of the key challenges in text-guided generation is the amplification of the effect induced by the conditioned text. To this end, Ho et al. \cite{ho2021classifier} have presented the classifier-free guidance technique, where the prediction is also performed unconditionally, which is then extrapolated with the conditioned prediction. More formally, let $\varnothing = \psi("")$ be the \textit{embedding} of a null text and let $w$ be the guidance scale parameter, then the classifier-free guidance prediction is defined by: \\[-10pt]
$$
\tilde{\eps}_\theta(z_t,t,\textemb,\varnothing) = w \cdot \eps_\theta(z_t,t,\textemb) + (1-w) \ \cdot \eps_\theta(z_t,t,\varnothing).
$$
%\\[-8pt]
E.g., $w=7.5$ is the default parameter for Stable Diffusion.

\vspace{-0.25cm}
\paragraph{DDIM inversion.}

A simple inversion technique was suggested for the DDIM sampling \cite{dhariwal2021diffusion,song2020denoising}, based on the assumption that the ODE process can be reversed in the limit of small steps: \\[-8pt]
$$
z_{t+1} = \sqrt{\frac{\alpha_{t+1}}{\alpha_t}}z_t + \left(\sqrt{\frac{1}{\alpha_{t+1}}-1}-\sqrt{\frac{1}{\alpha_t}-1}\right)\cdot \eps_\theta(z_t,t,\textemb).
$$ \\[-8pt]
In other words, the diffusion process is performed in the reverse direction, that is $z_0 \rightarrow z_T$ instead of $z_T \rightarrow z_0$, where $z_0$ is set to be the encoding of the given real image.

\subsection{Pivotal Inversion}
\label{sec:pivotal}
%\vspace{-0.1cm}

Recent inversion works use random noise vectors for each iteration of their optimization, aiming at mapping every noise vector to a single image. We observe that this is inefficient as inference requires only a single noise vector. Instead, inspired by GAN literature \cite{roich2021pivotal}, we seek to perform a more "local" optimization, ideally using only a single noise vector. In particular, we aim to perform our optimization around a pivotal noise vector which is a good approximation and thus allows a more efficient inversion.

We start by studying the DDIM inversion. In practice, a slight error is incorporated in every step. For unconditional diffusion models, the accumulated error is negligible and the DDIM inversion succeeds. However, recall that meaningful editing using the Stable Diffusion model \cite{rombach2021highresolution} requires applying classifier-free guidance with a large guidance scale $w > 1$. We observe that such a guidance scale amplifies the accumulated error.
Therefore, performing the DDIM inversion procedure with classifier-free guidance results not only in visual artifacts, but the obtained noise vector might be out of the Gaussian distribution. The latter decreases the editability, i.e., the ability to edit using the particular noise vector.

We do recognize that using DDIM inversion with guidance scale $w=1$ provides a rough approximation of the original image which is highly editable but far from accurate. More specifically, the reversed DDIM produces a $T$ steps trajectory between the image encoding $z_0$ to a Gaussian noise vector $z^*_T$. 
Again, a large guidance scale is essential for editing. Hence, we focus on feeding $z^*_T$ to the diffusion process with classifier-free guidance ($w > 1$). This results in high editability but inaccurate reconstruction, since the intermediate latent codes deviate from the trajectory, as illustrated in \cref{fig:method_diagram}. Analysis of different guidance scale values for the DDIM inversion is provided in \cref{sec:ablation_supp} (\cref{fig:ddim_graphs}).

Motivated by the high editability, we refer to this initial DDIM inversion with $w=1$ as our \emph{pivot} trajectory and perform our optimization around it with the standard guidance scale, $w > 1$. That is, our optimization maximizes the similarity to the original image while maintaining our ability to perform meaningful editing.
 In practice, we execute a separate optimization for each timestamp $t$ in the order of the diffusion process $t=T\rightarrow t=1$ with the objective of getting close as possible to the initial trajectory $z^*_T,\ldots,z^*_0$: \\[-8pt]
 \begin{equation}
\min \norm{z^*_{t-1}-z_{t-1}}^2,
\end{equation}
 \\[-8pt]
 where $z_{t-1}$ is the intermediate result of the optimization.
 Since our pivotal DDIM inversion provides a rather good starting point, this optimization is highly efficient compared to using random noise vectors, as demonstrated in \cref{sec:ablation}.

 Note that for every $t<T$, the optimization should start from the endpoint of the previous step ($t+1$) optimization, otherwise our optimized trajectory would not hold at inference.
Therefore, after the optimization of step $t$, we compute the current noisy latent $\bar{z}_{t}$, which is then used in the optimization of the next step to ensure our new trajectory would end near $z_0$ (see \cref{eq} for more details).

% : \\[-7pt]
%  \begin{equation}
% \min \norm{z^*_{t-1}-z_{t-1}(\bar{z_t})}^2. 
% \end{equation}
% \\[-10pt]

\begin{figure}
\centering 
\includegraphics[width=\columnwidth]{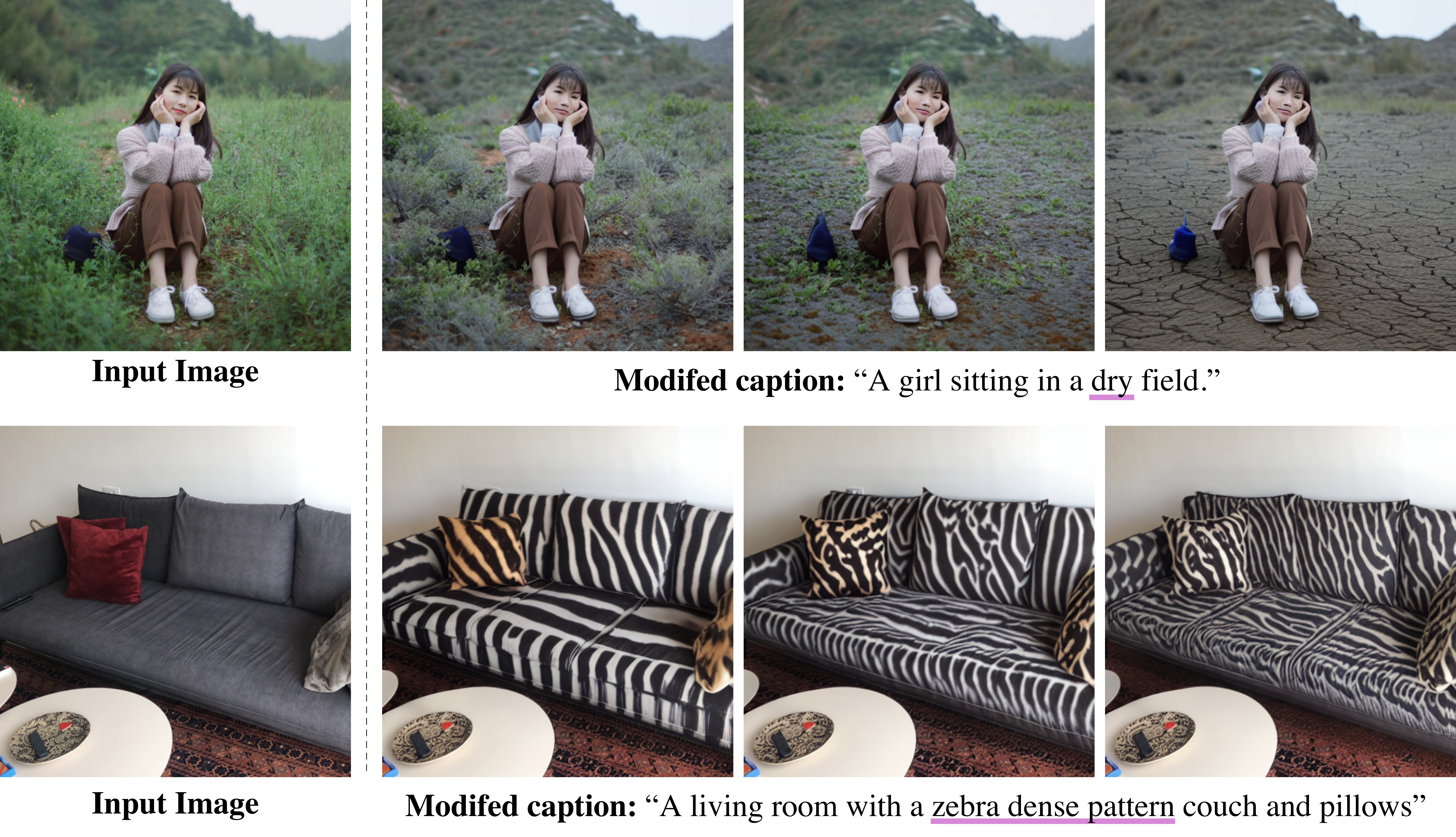} 

\vspace{-0.2cm}
\caption{\bf{Fine control editing using attention re-weighting.} \it{We can use attention re-weighting to further control the level of dryness over the field or create a denser zebra pattern over the couch.} } 
\vspace{-0.2cm}
\label{fig:exp_reweight} 

\end{figure}

%\vspace{-0.1cm}
\subsection{Null-text optimization}
\label{sec:null}
%\vspace{-0.1cm}

To successfully invert real images into the model's domain, recent works optimize the textual encoding \cite{gal2022image}, the network's weights \cite{ruiz2022dreambooth, valevski2022unitune}, or both \cite{Kawar2022ImagicTR}. Fine-tuning the model's weight for each image involves duplicating the entire model which is highly inefficient in terms of memory consumption. Moreover, unless fine-tuning is applied for each and every edit, it necessarily hurts the learned prior of the model and therefore the semantics of the edits. Direct optimization of the textual embedding results in a non-interpretable representation since the optimized tokens does not necessarily match pre-existing words. Therefore, an intuitive prompt-to-prompt edit becomes more challenging.

Instead, we exploit the key feature of the classifier-free guidance --- the result is highly affected by the unconditional prediction. Therefore, we replace the default null-text embedding with an optimized one, referred to as \emph{null-text optimization}. Namely, for each input image, we optimize only the unconditional embedding $\varnothing$, initialized with the null-text embedding. The model and the conditional textual embedding are kept unchanged.

This results in high-quality reconstruction while still allowing intuitive editing with Prompt-to-Prompt \cite{hertz2022prompt} by simply using the optimized unconditional embedding. Moreover, after a single inversion process, the same unconditional embedding can be used for multiple editing operations over the input image. Since null-text optimization is naturally less expressive than fine-tuning the entire model, it requires the more efficient pivotal inversion scheme.

We refer to optimizing a single unconditional embedding $\varnothing$ as a \emph{Global} null-text optimization. During our experiments, as shown in \cref{fig:exp_graph}, we have observed that optimizing a different "null embedding" $\varnothing_t$ for each timestamp $t$ significantly improves the reconstruction quality while this is well suited for our pivotal inversion. And so, we use per-timestamp unconditional embeddings $\{\varnothing_t\}_{t=1}^{T}$, and initialize $\varnothing_t$ with the embedding of the previous step $\varnothing_{t+1}$.
%Additional analysis of the Global Null-text is shown in the supplementary materials.

Putting the two components together, our full algorithm is presented in algorithm \ref{alg:null}.
 The DDIM inversion with $w=1$ outputs a sequence of noisy latent codes $z^*_T,\ldots,z^*_0$ where $z^*_0=z_0$. We initialize $\bar{z_T} = z_t$, and perform the following optimization with the default guidance scale $w = 7.5$ for the timestamps $t=T,\ldots,1$, each for $N$ iterations: \\[-14pt]
\begin{equation}\label{eq}
\min_{\varnothing_t} \norm{z^*_{t-1}-z_{t-1}(\bar{z_t}, \varnothing_t, \textemb)}^2.
\end{equation}
\\[-10pt]
For simplicity, $z_{t-1}(\bar{z_t}, \varnothing_t, \textemb)$ denotes applying DDIM sampling step using $\bar{z_t}$, the unconditional embedding $\varnothing_t$, and the conditional embedding $\textemb$. At the end of each step, we update \\[-16pt] 
$$
\bar{z}_{t-1} = z_{t-1}(\bar{z_t},  \varnothing_t, \textemb).
$$ 
\\[-15pt]
 We find that early stopping reduces time consumption, resulting in $\sim 1$ minute using a single $A100$ GPU.
 
 Finally, we can edit the real input image by using the noise $\bar{z_T} = z^*_T$ and the optimized unconditional embeddings $\{\varnothing_t\}_{t=1}^{T}$. Please refer to \cref{sup_implementation} for additional implementation details.

\begin{algorithm}[h]\label{alg}
\SetAlgoLined
\textbf{Input:} A source prompt embedding $\textemb = \psi(\mathcal{P})$ and input image $\mathcal{I}$.\\ %intermediate results of DDIM inversion $z^*_T,\ldots,z^*_0$ (using $w=1$).\\
\textbf{Output:} Noise vector $z_T$ and optimized embeddings $\{\varnothing_t\}_{t=1}^{T}$ .\\
 %Let $(s_{Normal},s_T,s_{T-1},\ldots,s_1)$ be a sequence of random seeds\; 
 \vspace{1mm} \hrule \vspace{1mm}
 Set guidance scale $w=1$; \\
 Compute the intermediate results  $z^*_T,\ldots,z^*_0$ using DDIM inversion over $\mathcal{I}$; \\
 Set guidance scale $w=7.5$; \\
 Initialize $\bar{z_T} \leftarrow z^*_T$, $\varnothing_T \leftarrow  \psi("")$; \\
 \For{$t=T,T-1,\ldots,1$}{
    \For{$j=0,\ldots,N-1$}{
        $\varnothing_t \leftarrow  \varnothing_t - \eta \nabla_{\varnothing} \norm{z^*_{t-1}-z_{t-1}(\bar{z_t}, \varnothing_t, \textemb)}^2$;
    }
    Set $\bar{z}_{t-1} \leftarrow z_{t-1}(\bar{z_t}, \varnothing_t, \textemb)$, $\varnothing_{t-1} \leftarrow  \varnothing_{t} $;
 }
 \textbf{Return} $\bar{z_T}$, $\{\varnothing_t\}_{t=1}^{T}$
 \caption{Null-text inversion}

\label{alg:null}
\end{algorithm}
%\vspace{-0.8cm}

%\vspace{-0.1cm}
\section{Ablation Study}
\label{sec:ablation}
\vspace{-0.1cm}

In this section, we validate the contribution of our main components, thoroughly analyzing the effectiveness of our design choices by conducting an ablation study. We focus on the fidelity to the input image which is an essential evaluation for image editing. In \cref{sec:results} we demonstrate that our method performs high-quality and meaningful manipulations.

\vspace{-0.4cm}
\paragraph{Experimental setting.} Evaluation is provided in \cref{fig:exp_graph}. We have used a subset of $100$ images and captions pairs, randomly selected from the \textit{COCO} \cite{chen2015microsoft} validation set.
We then applied our approach on each image-caption pair using the default Stable Diffusion hyper-parameters for an increasing number of iterations per diffusion step, $N=1,\ldots,20$ (see algorithm~\ref{alg:null}). The reconstruction quality was measured in terms of mean PSNR. We now turn to analyze different variations of our algorithm. 

\vspace{-0.4cm}
\paragraph{DDIM inversion.} We mark the DDIM inversion as a lower bound for our algorithm, as it is the starting point of our optimization, producing unsatisfying reconstruction when classifier-free guidance is applied (see \cref{sec:pivotal}).

\vspace{-0.4cm}
\paragraph{VQAE.} For an upper bound, we consider the reconstruction using the VQ auto-encoder \cite{esser2020taming}, denoted \textit{VQAE}, which is used by the Stable Diffusion model. Although the latent code or the VQ-decoder can be further optimized according to the input image, this is out of our scope, since it would be only applicable to this specific model \cite{rombach2021highresolution} while we aim for a general algorithm. Therefore, our optimization treats its encoding $z_0$ as ground truth, as the obtained reconstruction is quite accurate in most cases.

\begin{figure}
\setlength{\tabcolsep}{0.1pt}
    %\centering
    { \scriptsize %\footnotesize

\begin{tabular}{p{0.2\columnwidth}  p{0.2\columnwidth} p{0.2\columnwidth} p{0.2\columnwidth} p{0.2\columnwidth}}

\multicolumn{1}{c}{{ \bf Input}}&
\multicolumn{1}{c}{{ \bf Text2LIVE}}&
\multicolumn{1}{c}{{ \bf VQGAN+CLIP}}&
\multicolumn{1}{c}{{ \bf SDEdit}}&
\multicolumn{1}{c}{{ \bf Ours}} \\

{\includegraphics[width=0.19\columnwidth]{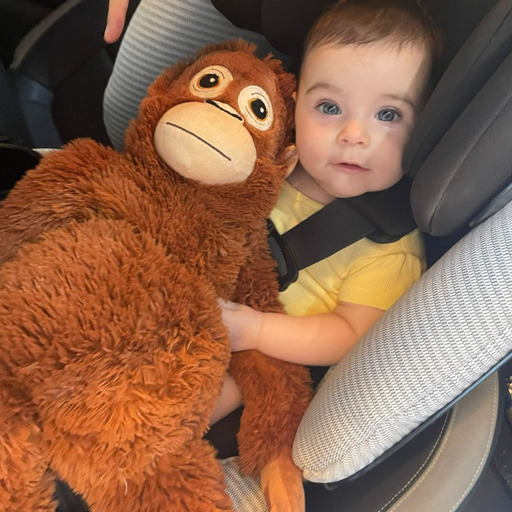}}&
{\includegraphics[width=0.19\columnwidth]{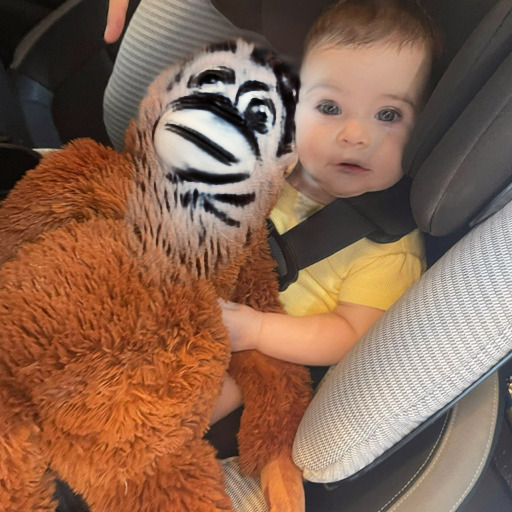}} &
{\includegraphics[width=0.19\columnwidth]{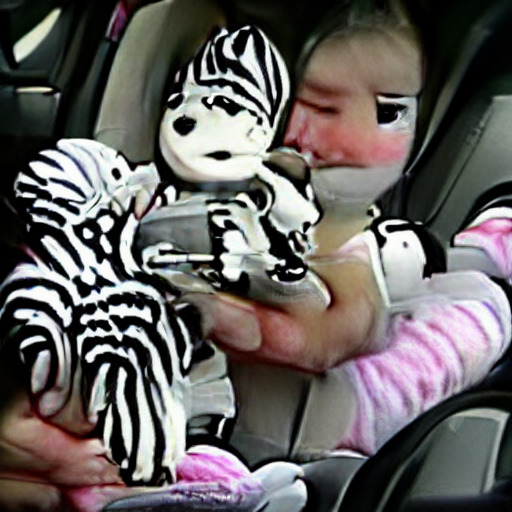}} &
{\includegraphics[width=0.19\columnwidth]{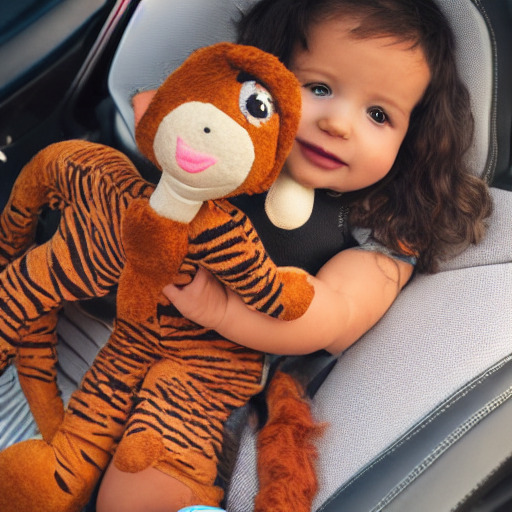}} &
{\includegraphics[width=0.19\columnwidth]{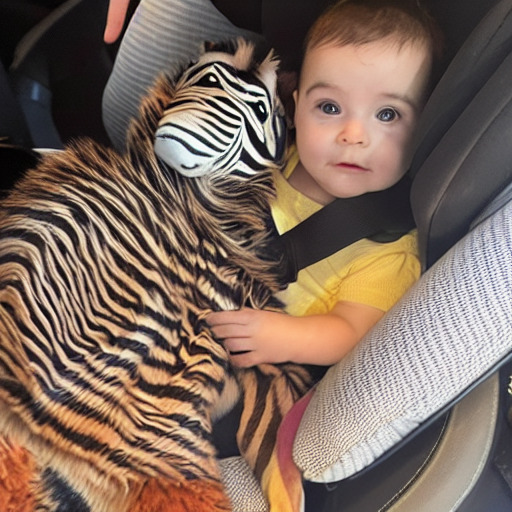}} \\

\multicolumn{5}{c}{"A baby holding her \st{monkey} {\color{RoyalPurple} \bf zebra} doll."} \\

{\includegraphics[width=0.19\columnwidth]{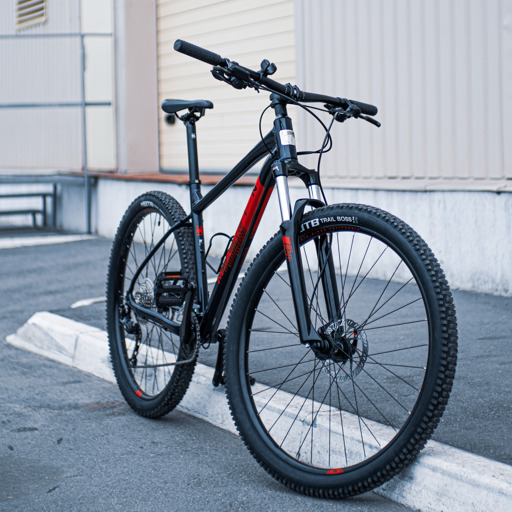}}&
{\includegraphics[width=0.19\columnwidth]{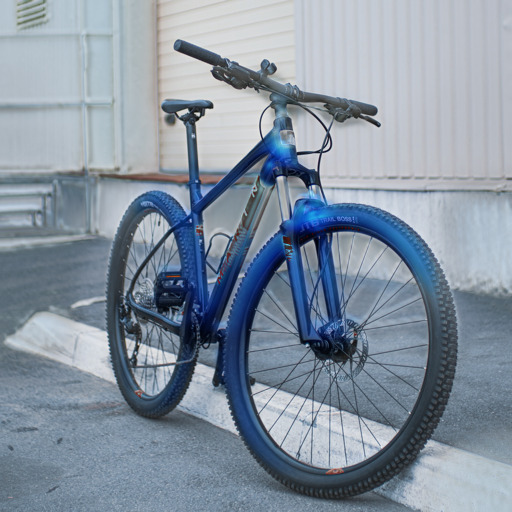}} &
{\includegraphics[width=0.19\columnwidth]{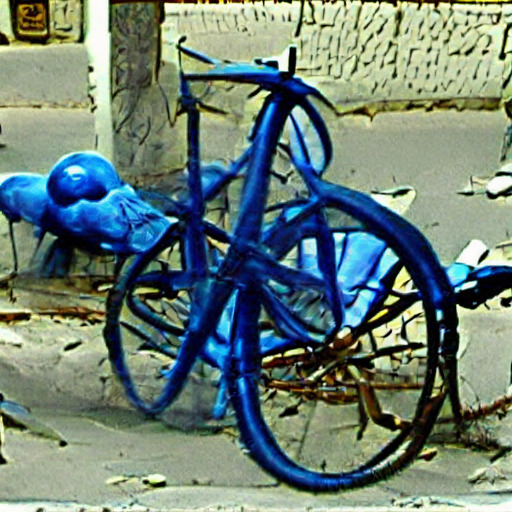}} &
{\includegraphics[width=0.19\columnwidth]{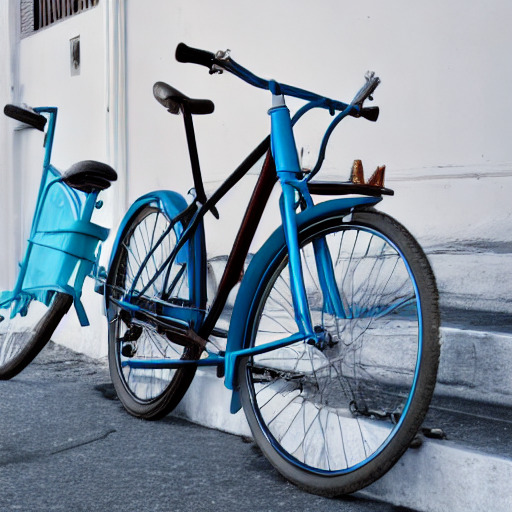}} &
{\includegraphics[width=0.19\columnwidth]{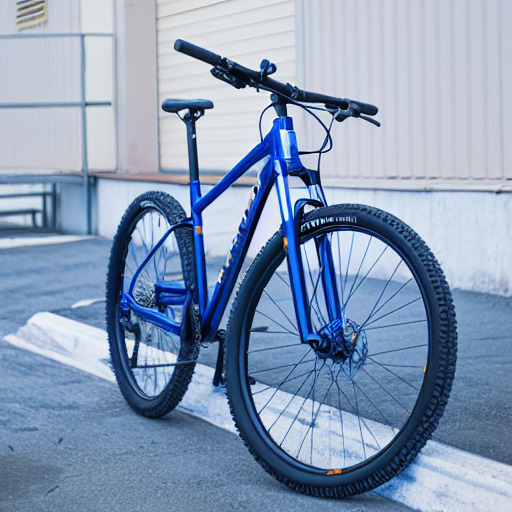}} \\

\multicolumn{5}{c}{"A {\color{RoyalPurple} \bf blue } bicycle is parking on the side of the street"} \\

{\includegraphics[width=0.19\columnwidth]{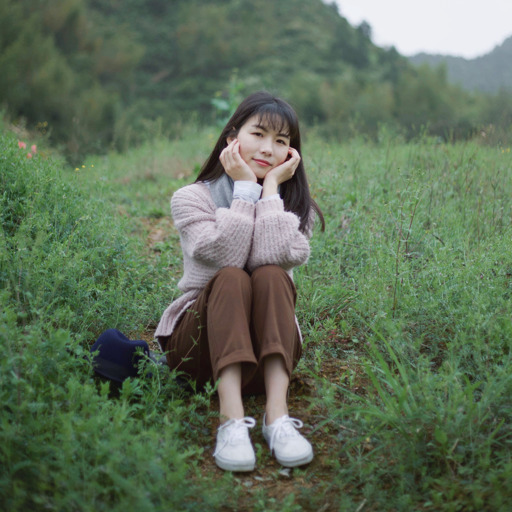}}&
{\includegraphics[width=0.19\columnwidth]{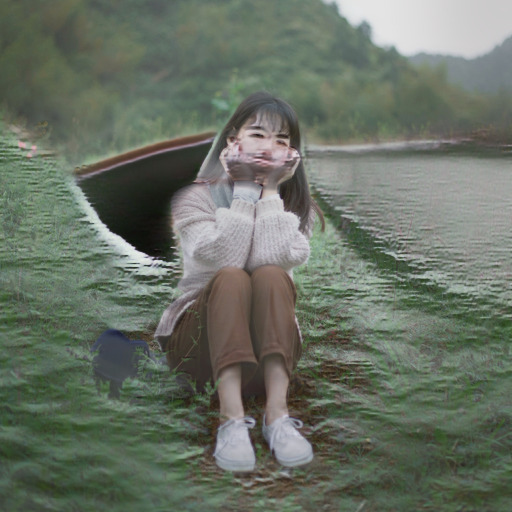}} &
{\includegraphics[width=0.19\columnwidth]{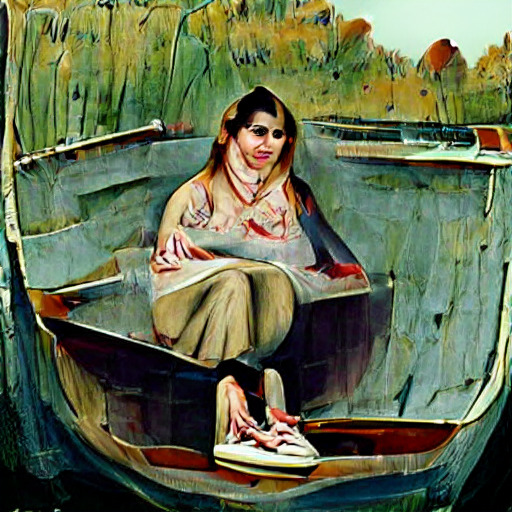}} &
{\includegraphics[width=0.19\columnwidth]{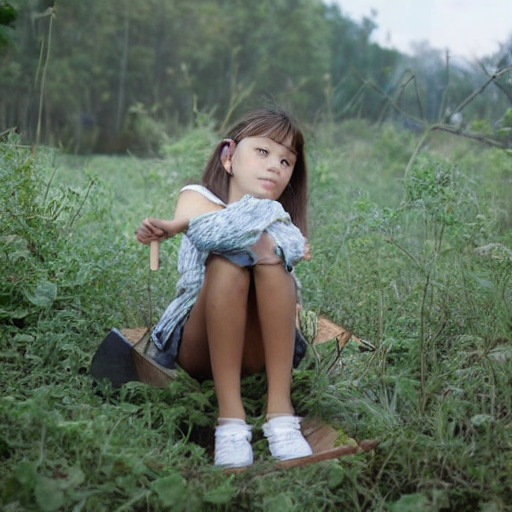}} &
{\includegraphics[width=0.19\columnwidth]{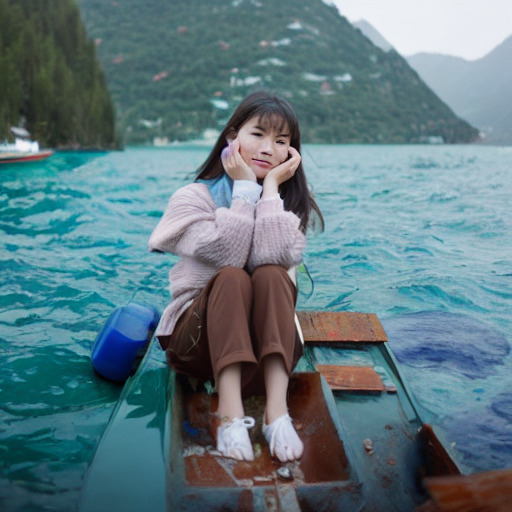}} \\

\multicolumn{5}{c}{"A girl sitting in a \st{field} {\color{RoyalPurple} \bf boat}"} \\

% {\includegraphics[width=0.19\columnwidth]{images/gt/kid_tree.jpg}}&
% {\includegraphics[width=0.19\columnwidth]{images/text2live/kid_tree_099.jpg}} &
% {\includegraphics[width=0.19\columnwidth]{images/sde/kid_tree_099.jpg}} &
% {\includegraphics[width=0.19\columnwidth]{images/ours/kid_tree_099.jpg}} \\

% \multicolumn{5}{c}{"A \st{child} {\color{RoyalPurple} \bf monkey} is climbing on a tree""} \\

{\includegraphics[width=0.19\columnwidth]{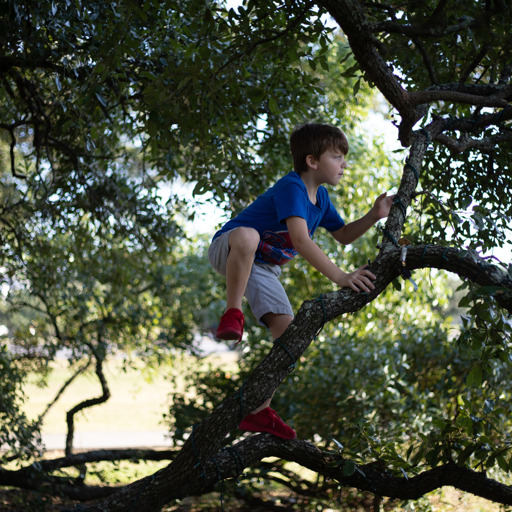}}&
{\includegraphics[width=0.19\columnwidth]{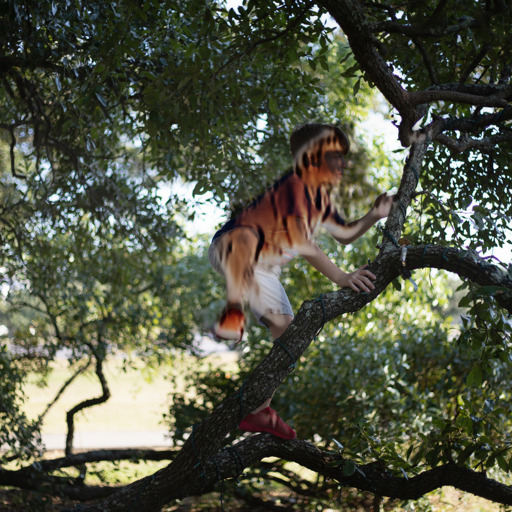}} &
{\includegraphics[width=0.19\columnwidth]{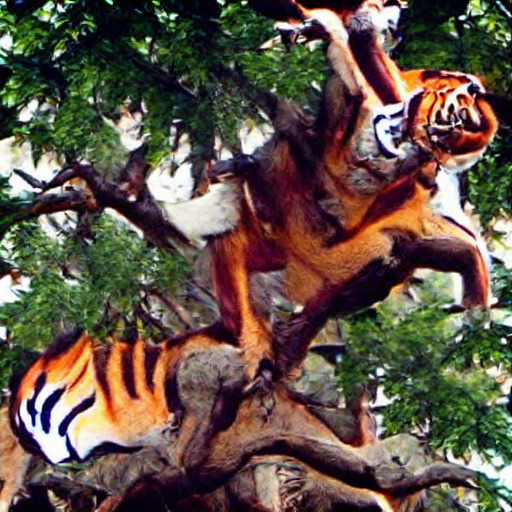}} &
{\includegraphics[width=0.19\columnwidth]{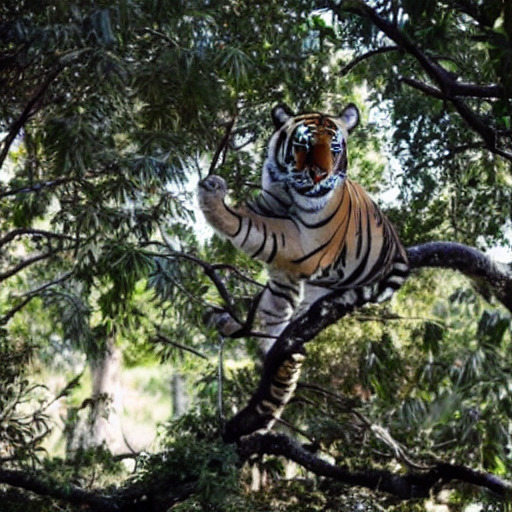}} &
{\includegraphics[width=0.19\columnwidth]{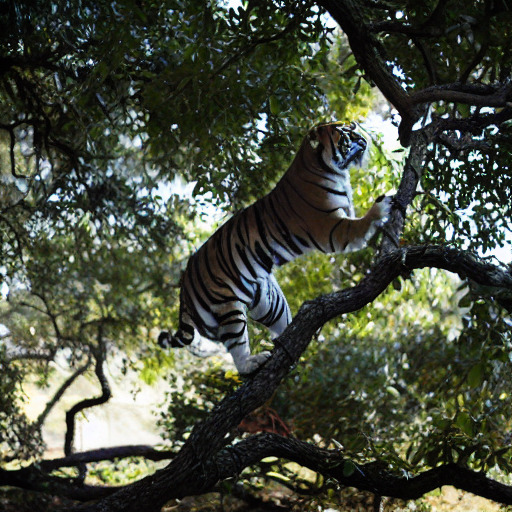}} \\

\multicolumn{5}{c}{"A \st{child} {\color{RoyalPurple} \bf tiger} is climbing on a tree""} \\

\end{tabular}
}

\vspace{-0.2cm}
\caption{{\bf Comparison.} {\it Text2LIVE \cite{bar2022text2live} excels at replacing textures locally but struggles to perform more structured editing, such as replacing a kid with a tiger. VQGAN+CLIP \cite{katherine2021vqganclip} obtains inferior realism. SDEdit \cite{meng2021sdedit} fails to faithfully reconstruct the original image, resulting in identity drift when humans are involved. Our method achieves realistic editing of both textures and structured objects while retaining high fidelity to the original image. Additional examples provided in \cref{sec:supp_results} (\cref{fig:supp_comparison}). }}
\vspace{-0.25cm}
% \vspace{-0.15cm}
% \ron{The gap is best viewed in the supplementary videos.}}
\label{fig:comparison} %\vspace{-7pt}
\end{figure}

\vspace{-0.4cm}
\paragraph{Our method.} As can be seen in \cref{fig:exp_graph}, our method converges to a near-optimal reconstruction with respect to the VQAE upper bound after a total number of $500$ iterations ($N=10$) and even after $250$ iterations (\(\sim\) 1 minute on an A100 GPU) we achieve high-quality inversion. 

\vspace{-0.4cm}
\paragraph{Random Pivot.}
We validate the importance of the DDIM initialization by replacing the DDIM-based trajectory with a \textit{single random trajectory} of latent codes, sharing the same starting point $z_0$ --- the input image encoding. In other words, we randomly sample a single Gaussian noise $\sim N(0, I)$ for each image and use it to noise the corresponding encoding $z_0$ from $t=1$ to $t=T$ using the diffusion scheduler. As presented in \cref{fig:exp_graph}, the DDIM initialization is crucial for fast convergence, since the initial error becomes significantly larger when the pivot is random.

\vspace{-0.4cm}
\paragraph{Robustness to different input captions.}
Since we require an input caption, it is only natural to ask whether our method is highly sensitive to the chosen caption. We take this to the extreme by sampling a \textit{random caption} from the dataset for each image. Even with unaligned captions, the optimization converges to an optimal reconstitution with respect to the VQAE. Therefore, we conclude that our inversion is robust to the input caption. Clearly, choosing a random caption is undesired for text-based editing. But, providing any reasonable and editable prompt would work, including using an off-the-shelve captioning model \cite{mokady2021clipcap, stefanini2022show}. This is illustrated in \cref{sec:ablation_supp} (\cref{fig:exp_multi_cap}). We invert an image using multiple captions, demonstrating that the edited parts should be included in the source caption in order to produce semantic attention maps for editing. For example, to edit the print on the shirt, the source caption should include a "shirt with a drawing" or a similar phrase.

\vspace{-0.4cm}
\paragraph{Global null-text embedding.}
We refer to optimizing a single embedding $\varnothing$ for all timestamps as a \textit{Global} embedding. As can be seen, such optimization struggles to converge, since it is less expressive than our final approach, which uses embedding per-timestamp $\{\varnothing_t\}_{t=1}^{T}$. See additional implementation details in \cref{sup_implementation}.

\vspace{-0.4cm}
\paragraph{Textual inversion.}
We compare our method to \textit{textual inversion}, similar to the proposed method by Gal et al.~\cite{gal2022image}. We optimize the textual embedding $\textemb = \psi(\mathcal{P})$ using random noise samples instead of pivotal inversion. 
That is, we randomly sample a different Gaussian noise for each optimization step and obtain $z_t$ by adding it to $z_0$ according to the diffusion scheduler. Intuitively, this objective aims to map all noise vectors to a single image, in contrast to our pivotal tuning inversion which focuses on a single trajectory of noisy vectors. The optimization objective is then defined: \\[-8pt]
\begin{equation}
\min_\textemb E_{z_0,\eps\sim N(0,I),t} \norm{\eps-\eps_\theta(z_t,t,\textemb)}^2.
\end{equation}
\\[-10pt]
Note that Gal et al.~\cite{gal2022image} have attempted to regenerate a specific object rather than achieve an accurate inversion. As presented in \cref{fig:exp_graph}, the convergence is much slower than ours and results in poor reconstruction quality.

\vspace{-0.4cm}
\paragraph{Textual inversion with a pivot.}
We observe that employing our pivotal inversion with the mentioned textual inversion improves the reconstruction quality significantly, results in a comparable reconstruction to ours. This further demonstrates the power of performing the optimization using a pivot. %In other words, optimizing the conditioned embedding around a pivot results in a comparable reconstruction to ours. 
However, we do observe that editability is reduced compared to the null-text optimization. In particular, as demonstrated in \cref{sec:ablation_supp} (\cref{fig:albation_text_pivot}), the attention maps are less accurate which decreases the performance of Prompt-to-prompt editing.

\vspace{-0.4cm}
\paragraph{Null-text optimization without pivotal inversion.}
We observe that optimizing the unconditional null-text embedding using random noise vectors, instead of pivotal inversion as described in previous paragraphs, completely breaks the null-text optimization. The results are inferior even to the DDIM inversion baseline as presented in \cref{sec:ablation_supp} (\cref{fig:exp_ablation_qual_sup,fig:exp_ablation_qual_sup2}). We hypothesize that null-text optimization is less expressive than model-tuning and thus depends on the efficient pivotal inversion, as it struggles to map all noise vectors to a single image.

\begin{figure}
\centering 
\vspace{-0.5cm}
\includegraphics[width=\columnwidth]{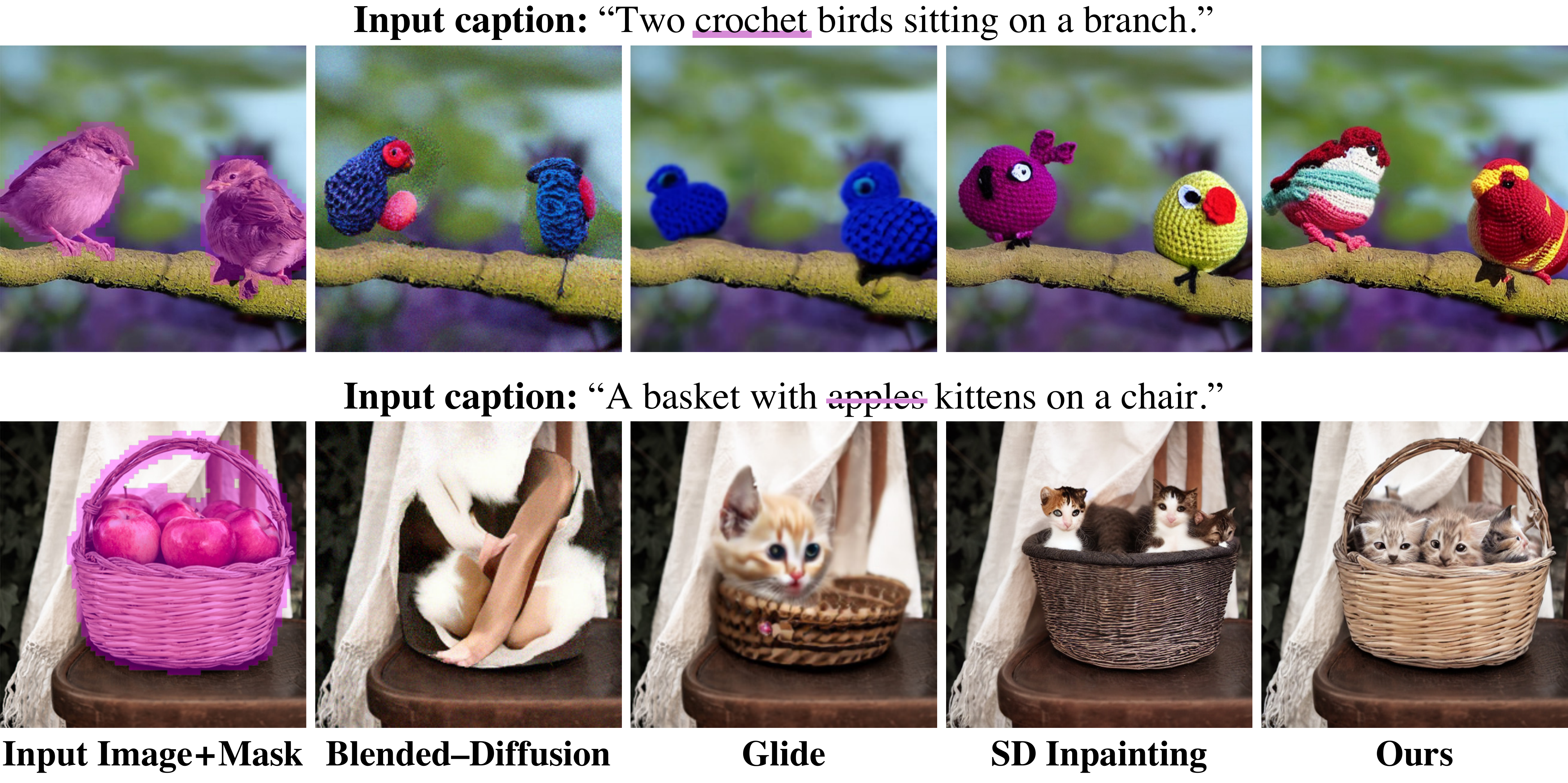} 

\vspace{-0.2cm}
\caption{{\bf Comparison to mask-based methods .} {\it As can be seen, mask-based methods do not require inversion as the region outside the mask is kept. However, unlike our approach, such methods often struggle to preserve details that are found inside the masked region. For example, basket size is not preserved. } } 
\vspace{-0.4cm}
\label{fig:exp_inaint} 

\end{figure}

%\vspace{-0.1cm}
\section{Results}
\label{sec:results}
%\vspace{-0.1cm}

Real image editing is presented in \cref{fig:exp_edit_people,fig:exp_reweight,fig:teaser}, showing our method not only reaches remarkable reconstruction quality but also retains high editability. In particular, we use the intuitive approach of Prompt-to-Prompt \cite{hertz2022prompt} and demonstrate that the editing capabilities which previously were constrained to synthesized images are now applied to real images using our inversion technique.

As can be seen in \cref{fig:exp_edit_people}, our method effectively modifies both textures ("floral shirt") and structured objects ("baby" to "robot"). Since we support the local editing of Prompt-to-Prompt and achieve high-fidelity reconstruction, the original identity is well preserved, even in the challenging case of a baby face.
\cref{fig:exp_edit_people} also illustrates that our method requires only a single inversion process to perform multiple editing operations. Using a single inversion procedure, we can modify hair color, glasses, expression, background, and lighting and even replace objects or put on a joker make-up (bottom rows). % \ypc{mask may be confusing with editing mask discussed elsewhere. maybe use "apply makeup"}.
Using Prompt-to-Prompt, we can also attenuate or amplify the effect of a specific word over real images, as appeared in \cref{fig:exp_reweight}.
For additional examples, please refer to \cref{sec:supp_results}.

Visual results for our high-fidelity reconstruction are presented in \cref{fig:teaser,fig:sdedit_null}, and \cref{sec:supp_results} (\cref{fig:supp_comparison}), supporting our quantitative measures in \cref{sec:ablation}.

 \begin{table}
%\vspace{-0.5cm}
 \caption{{\bf User study results.} {\it The participants were asked to select the best editing result in terms of fidelity to both the input image and the textual edit instruction.}}
 \footnotesize
 \centering
\vspace{-0.25cm}
\begin{tabular}{cccc}

\toprule

VQGAN+CLIP & Text2Live & SDEDIT & Ours \\
% 2.1\% & 9.9\% & 5.2\%  & {\bf 82.8\%} \\
3.8\% & 16.6\% & 14.5\%  & {\bf 65.1\%} \\

\bottomrule
\end{tabular}
\vspace{-0.2cm}
\label{tab:user-study}
 \end{table}

%\vspace{-0.1cm}
\subsection{Comparisons}
\label{sec:comparisons}
%\vspace{-0.1cm}

Our method aims attention at intuitive editing using only text, and so we compare our results to other text-only editing methods: (1) \textit{VQGAN+CLIP} \cite{katherine2021vqganclip}, (2) \textit{Text2Live} \cite{bar2022text2live}, and (3) \textit{SDEedit} \cite{meng2021sdedit}. We evaluated these on the images used by Bar-Tal et al.~\cite{bar2022text2live} and photos that include more structured objects, such as humans and animals, which we gathered from the internet. In total, we use $100$ samples of images, input captions, and edited captions.

We also compare our method to the mask-based methods of (4) Glide \cite{nichol2021glide}, (5) Blended-Diffusion \cite{avrahami2022blended}, and (6) Stable Diffusion Inpaint \cite{rombach2021highresolution}. The latter fine-tunes the diffusion model using an inpainting objective, allowing simple editing by inpainting a masked region using a target prompt.

Lastly, we consider the concurrent work of (7) Imagic \cite{Kawar2022ImagicTR}, which employs model-tuning per editing operation and has been designed for the Imagen model \cite{saharia2022photorealistic}. 
We refrain from comparing to the concurrent works of Unitune \cite{valevski2022unitune} and DiffEdit \cite{couairon2022diffedit} as there are no available implementations.

\vspace{-0.3cm}
\paragraph{Qualitative Comparison.}
As presented in \cref{fig:comparison}, VQGAN+CLIP \cite{katherine2021vqganclip} mostly produces unrealistic results. Text2LIVE \cite{bar2022text2live} handles texture modification well but fails to manipulate more structured objects, e.g., placing a boat (3rd row). 
Both struggle due to the use of CLIP \cite{radford2021learning} which lacks a generative ability.
In SDEdit \cite{meng2021sdedit}, the noisy image is fed to an intermediate step in the diffusion process, and therefore, it struggles to faithfully reconstruct the original details. This results in severe artifacts when fine details are involved, such as human faces. For instance, identity drifts in the top row, and the background is not well preserved in the 2nd row.
Contrarily, our method successfully preserves the original details, while allowing a wide range of realistic and meaningful editing, from simple textures to replacing well-structured objects.

\cref{fig:exp_inaint} presents a comparison to mask-based methods, showing these struggles to preserve details that are found inside the masked region. This is due to the masking procedure that removes important structural information, and therefore, some capabilities are out of the inpainting reach.

A comparison to Imagic \cite{Kawar2022ImagicTR}, which operates in a different setting -- requiring model-tuning for each editing operation, is provided in \cref{sec:ablation_supp} (\cref{fig:imagic}). We first employ the unofficial Imagic implementation for Stable Diffusion and present the results for different values of the interpolation parameter $\alpha = 0.6,0.7,0.8,0.9$. This parameter is used to interpolate between the target text embedding and the optimized one \cite{Kawar2022ImagicTR}. %, where a larger value of $\alpha$ increases the fidelity to the target text. 
In addition, the Imagic authors applied their method using the Imagen model over the same images, using the following parameters $\alpha=0.93, 0.86, 1.08$.
As can be seen, Imagic produces highly meaningful editing, especially when the Imagen model is involved.
However, Imagic struggles to preserve the original details, such as the identity of the baby ($1$st row) or cups in the background ($2$nd row). Furthermore, we observe that Imagic is quite sensitive to the interpolation parameter $\alpha$, as a high value reduces the fidelity to the image and a low value reduces the fidelity to the text guidance, while a single value cannot be applied to all examples. Lastly, Imagic takes a longer inference time, as shown in \cref{sec:supp_results} (\cref{tab:times}).

\vspace{-0.3cm}
\paragraph{Quantitative Comparison.}

Since ground truth is not available for text-based editing of real images, quantitative evaluation remains an open challenge. Similar to \cite{bar2022text2live, hertz2022prompt}, we present a user study in \cref{tab:user-study}. $50$ participants have rated a total of $48$ images for each baseline. The participants were recruited using \emph{Prolific} (prolific.co). We presented side-by-side images produced by: VQGAN+CLIP, Text2LIVE, SDEdit, and our method (in random order). We focus on methods that share a similar setting to ours -- no model tuning and mask requirement. The participants were asked to choose the method that better applies the requested edit while preserving most of the original details. A print screen is provided in \cref{sup:user_study} (\cref{fig:user_study}).
As shown in \cref{tab:user-study}, most participants favored our method.

Quantitative comparison to Imagic is presented in \cref{sec:ablation_supp} (\cref{fig:exp_graph_imagic}), using the unofficial Stable Diffusion implementation. According to these measures, our method achieves better scores for LPIPS perceptual distance, indicating a better fidelity to the input image.

\label{tab:times}

%\vspace{-0.1cm}
\subsection{Evaluating Additional Editing Technique}
%\vspace{-0.1cm}

Most of the presented results consist of applying our method with the editing technique of Prompt-to-Prompt \cite{hertz2022prompt}. However, we demonstrate that our method is not confined to a specific editing approach, by showing it improves the results of the SDEdit \cite{meng2021sdedit} editing technique.

\begin{figure}
\setlength{\tabcolsep}{0.2pt}
    %\centering
    { \scriptsize %\footnotesize
\vspace{-0.4cm}
\includegraphics[width=\columnwidth]{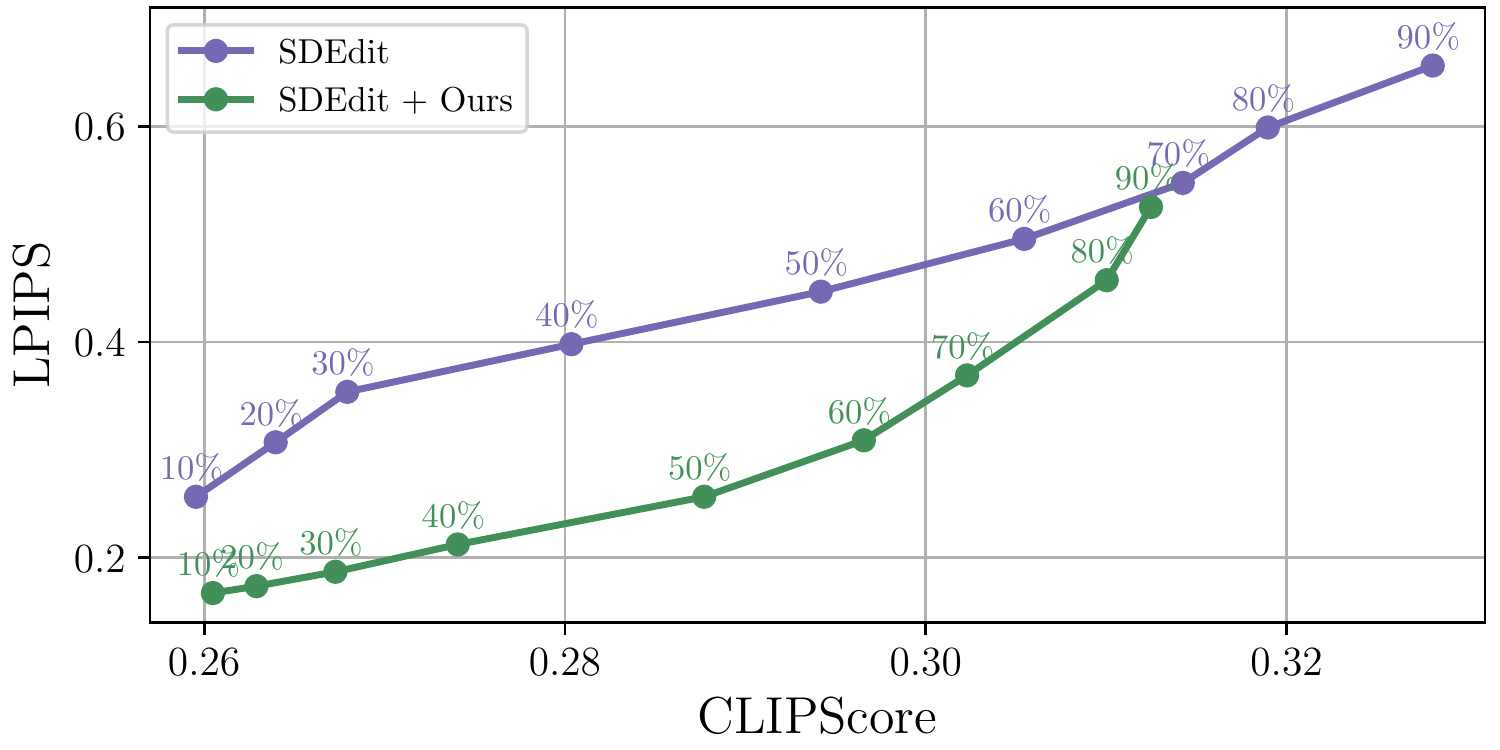} 
\\[4pt]
\begin{tabular}{>{\centering\arraybackslash}p{0.2\columnwidth}  >{\centering\arraybackslash}p{0.2\columnwidth} >{\centering\arraybackslash}p{0.2\columnwidth} >{\centering\arraybackslash}p{0.2\columnwidth} >{\centering\arraybackslash}p{0.2\columnwidth}}

\multicolumn{1}{c}{{ \bf Input Image}}&
\multicolumn{1}{c}{{ \bf Our Inversion}}&
\multicolumn{1}{c}{{ \bf SDEdit}}&
\multicolumn{1}{c}{{ \bf Ours + SDEdit}}&
\multicolumn{1}{c}{{ \bf Ours + P2P}} \\

% \multicolumn{5}{c}{"A living room with a {\color{RoyalPurple} \bf leather} couch and pillows"} \\

{\includegraphics[width=0.19\columnwidth]{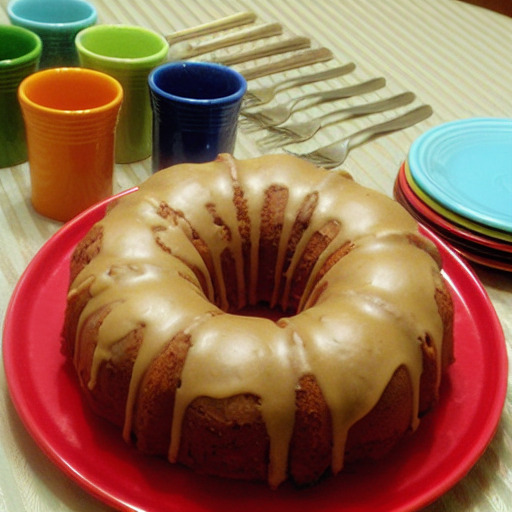}}&
{\includegraphics[width=0.19\columnwidth]{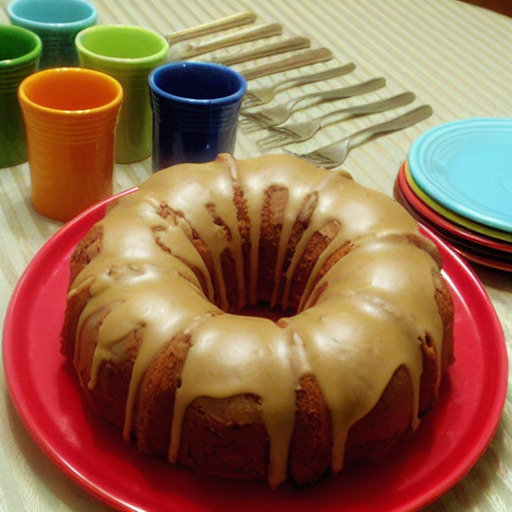}} &
{\includegraphics[width=0.19\columnwidth]{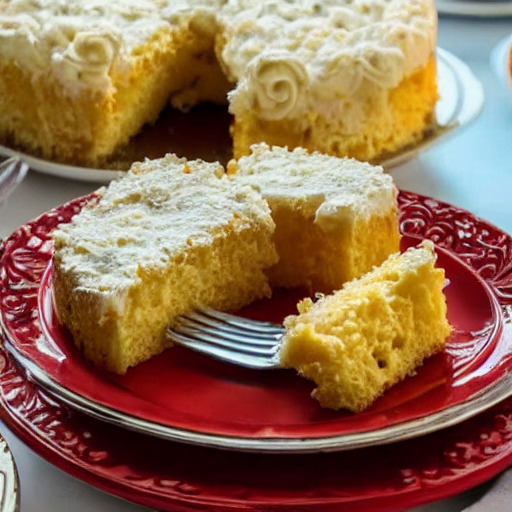}} &
{\includegraphics[width=0.19\columnwidth]{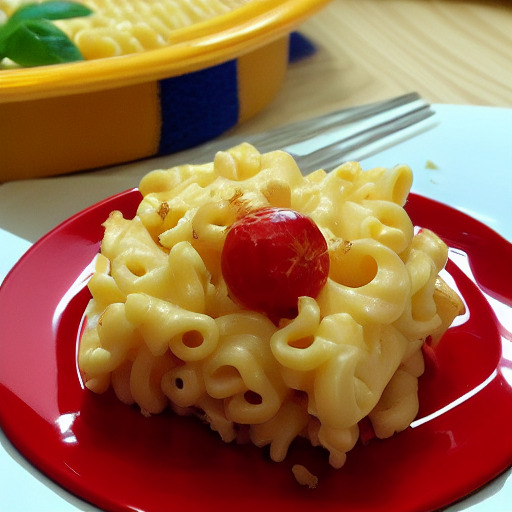}} &
{\includegraphics[width=0.19\columnwidth]{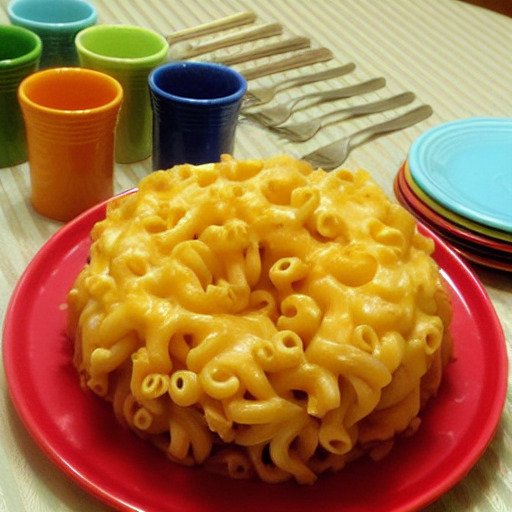}} \\

\multicolumn{5}{c}{"{\color{RoyalPurple} \bf Macaroni} cake on a table."} \\

{\includegraphics[width=0.19\columnwidth]{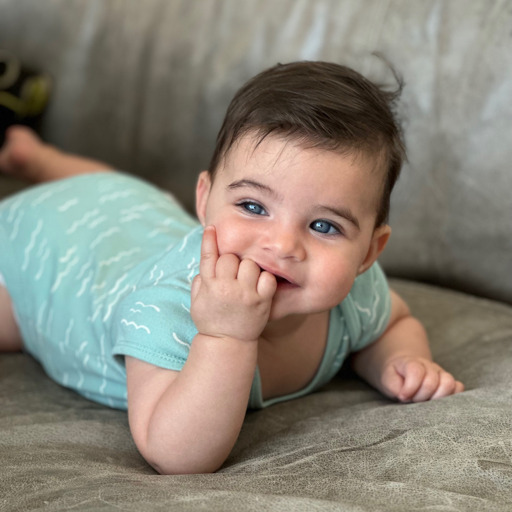}}&
{\includegraphics[width=0.19\columnwidth]{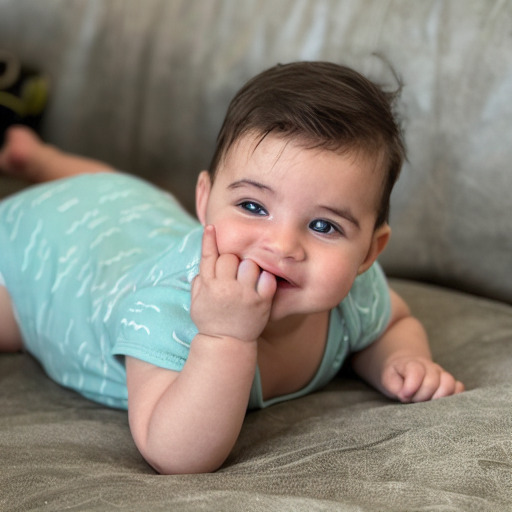}} &
{\includegraphics[width=0.19\columnwidth]{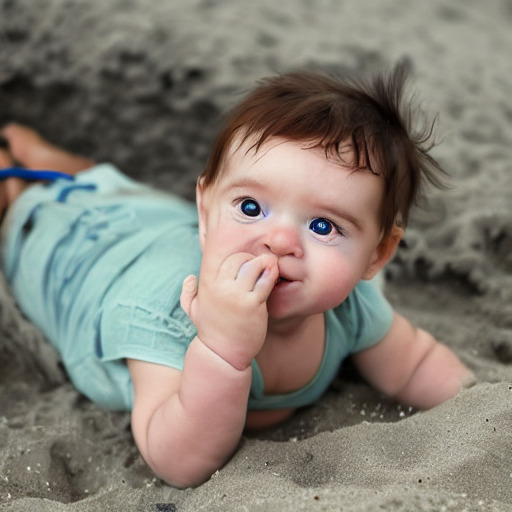}} &
{\includegraphics[width=0.19\columnwidth]{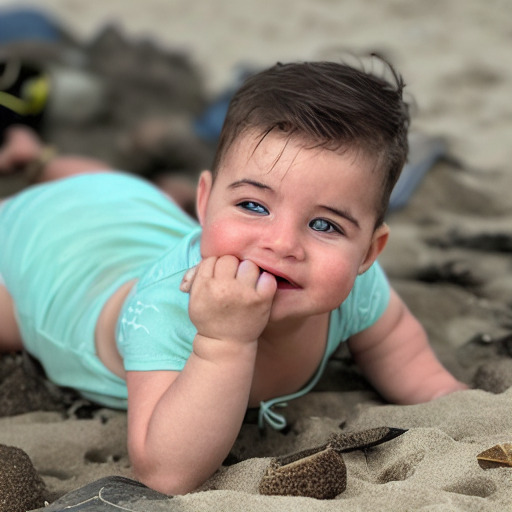}} &
{\includegraphics[width=0.19\columnwidth]{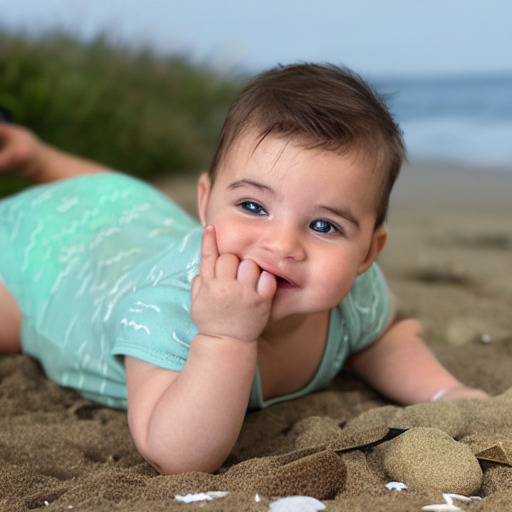}} \\

\multicolumn{5}{c}{"A baby wearing a blue shirt lying on the \st{sofa} {\color{RoyalPurple} \bf beach}"} \\

% {\includegraphics[width=0.19\columnwidth]{images/gt/basket_052.jpg}}&
% {\includegraphics[width=0.19\columnwidth]{images/inversion/basket.jpg}} &
% {\includegraphics[width=0.19\columnwidth]{images/sde/049.jpg}} &
% {\includegraphics[width=0.19\columnwidth]{images/sdedit_null/049.jpg}} &
% {\includegraphics[width=0.19\columnwidth]{images/ours/049.jpg}} \\

% \multicolumn{5}{c}{"A basket with \st{apples} {\color{RoyalPurple} \bf pears} on a chair"} \\

\end{tabular}
}

\vspace{-0.25cm}
\caption{{\bf Our method improves SDEdit results.} {\it 
Top: we evaluate SDEdit with and without applying null-text inversion. 
In each measure, a different SDEdit parameter is used, i.e., different percent of diffusion steps are applied over the noisy image (marked on the curve). 
We measure both fidelity to the original image (via LPIPS, low is better) and fidelity to the target text (via CLIP, high is better).
Bottom, from left to right: input image, null-text inversion, SDEdit, applying SDEdit after null-text inversion, and applying Prompt-to-Prompt after null-text inversion. 
As can be seen, our inversion significantly improves the fidelity to the original image when applied before SDEdit. }}
\vspace{-0.4cm}
% \vspace{-0.15cm}
% \ron{The gap is best viewed in the supplementary videos.}}
\label{fig:sdedit_null} %\vspace{-7pt}
\end{figure}

In \cref{fig:sdedit_null} (top), we measure the fidelity to the original image using LPIPS perceptual distance \cite{Zhang2018TheUE} (lower is better), and the fidelity to the target text using CLIP similarity \cite{radford2021learning} (higher is better) over $100$ examples. We use different values of the SDEdit parameter $t_0$ (marked on the curve), i.e., we start the diffusion process from different $t = t_0 \cdot T$ using a correspondingly noised input image. This parameter controls the trade-off between fidelity to the input image (low $t_0$) and alignment to the text (high $t_0$). We compare the standard SDEdit to first applying our inversion and then performing SDEdit while replacing the null-text embedding with our optimized embeddings. As shown, our inversion significantly improves the fidelity to the input image.

This is visually demonstrated in \cref{fig:sdedit_null} (bottom). Since the parameter $t_0$ controls a reconstruction-editability trade-off, we have used a different parameter for each method (SDEdit with and without our inversion) such that both achieve the same CLIP score. As can be seen, when using our method, the true identity of the baby is well preserved.

%\input{figures/experiments_graph_sdedit}

%\vspace{-0.15cm}
\section{Limitations}
%\vspace{-0.15cm}

While our method works well in most scenarios, it still faces some limitations. The most notable one is inference time. Our approach requires approximately one minute on GPU for inverting a single image. Then, infinite editing operations can be made, each takes only ten seconds. %While this is faster than most previous methods, it is still not enough for real-time applications. 
This is not enough for real-time applications.
Other limitations come from using Stable Diffusion \cite{rombach2021highresolution} and Prompt-to-Prompt editing \cite{hertz2022prompt}. First, the VQ auto-encoder produces artifacts in some cases, especially when human faces are involved. We consider the optimization of the VQ decoder as out of scope here, since this is specific to Stable Diffusion and we aim for a general framework. Second, we observe that the generated attentions maps of Stable Diffusion are less accurate compared to the attention maps of Imagen \cite{saharia2022photorealistic}, i.e., words might not relate to the correct region, indicating inferior text-based editing capabilities. 
Lastly, complicated structure modifications are out of reach for Prompt-to-Prompt, such as changing a seating dog to a standing one as in \cite{Kawar2022ImagicTR}.
Our inversion approach is orthogonal to the specific model and editing techniques, and we believe that these will be improved in the near future.

%\vspace{-0.15cm}
\section{Conclusions}
%\vspace{-0.15cm}
We have presented an approach to invert real images with corresponding captions into the latent space of a text-guided diffusion model while maintaining its powerful editing capabilities. 
Our two-step approach first uses DDIM inversion to compute a sequence of noisy codes, which roughly approximate the original image (with the given caption), then uses this sequence as a fixed pivot to optimize the input null-text embedding. Its fine optimization compensates for the inevitable reconstruction error caused by the classifier-free guidance component. 
Once the image-caption pair is accurately embedded in the output domain of the model, prompt-to-prompt editing can be instantly applied at inference time. 
By introducing two new technical concepts to text-guided diffusion models -- pivotal inversion and null-text optimization, we were able to bridge the gap between reconstruction and editability.
Our approach offers a surprisingly simple and compact means to reconstruct an arbitrary image, avoiding the computationally intensive model-tuning. We believe that null-text inversion paves the way for real-world use case scenarios for intuitive, text-based, image editing.

%\vspace{-0.2cm}

\section{Acknowledgments}
We thank Yuval Alaluf, Rinon Gal, Aleksander Holynski, Bryan Eric Feldman, Shlomi Fruchter and David Salesin
for their valuable inputs that helped improve this work, and to Bahjat Kawar, Shiran Zada and Oran Lang for providing us with their support for the Imagic \cite{Kawar2022ImagicTR} comparison.
We also thank Jay Tenenbaum for the help with writing the background.

\vspace{-.15cm}
{\small
\bibliographystyle{ieee_fullname}
\bibliography{egbib}

\begin{thebibliography}{10}\itemsep=-1pt

\bibitem{abdal2019image2stylegan}
Rameen Abdal, Yipeng Qin, and Peter Wonka.
\newblock Image2stylegan: How to embed images into the stylegan latent space?
\newblock In {\em Proceedings of the IEEE/CVF International Conference on
  Computer Vision}, pages 4432--4441, 2019.

\bibitem{abdal2020image2stylegan++}
Rameen Abdal, Yipeng Qin, and Peter Wonka.
\newblock Image2stylegan++: How to edit the embedded images?
\newblock In {\em Proceedings of the IEEE/CVF conference on computer vision and
  pattern recognition}, pages 8296--8305, 2020.

\bibitem{alaluf2021hyperstyle}
Yuval Alaluf, Omer Tov, Ron Mokady, Rinon Gal, and Amit~H. Bermano.
\newblock Hyperstyle: Stylegan inversion with hypernetworks for real image
  editing, 2021.

\bibitem{avrahami2022blendedlatent}
Omri Avrahami, Ohad Fried, and Dani Lischinski.
\newblock Blended latent diffusion.
\newblock {\em arXiv preprint arXiv:2206.02779}, 2022.

\bibitem{avrahami2022blended}
Omri Avrahami, Dani Lischinski, and Ohad Fried.
\newblock Blended diffusion for text-driven editing of natural images.
\newblock In {\em Proceedings of the IEEE/CVF Conference on Computer Vision and
  Pattern Recognition}, pages 18208--18218, 2022.

\bibitem{bar2022text2live}
Omer Bar-Tal, Dolev Ofri-Amar, Rafail Fridman, Yoni Kasten, and Tali Dekel.
\newblock Text2live: Text-driven layered image and video editing.
\newblock {\em arXiv preprint arXiv:2204.02491}, 2022.

\bibitem{bermano2022state}
Amit~H Bermano, Rinon Gal, Yuval Alaluf, Ron Mokady, Yotam Nitzan, Omer Tov,
  Oren Patashnik, and Daniel Cohen-Or.
\newblock State-of-the-art in the architecture, methods and applications of
  stylegan.
\newblock In {\em Computer Graphics Forum}, volume~41, pages 591--611. Wiley
  Online Library, 2022.

\bibitem{chen2015microsoft}
Xinlei Chen, Hao Fang, Tsung-Yi Lin, Ramakrishna Vedantam, Saurabh Gupta, Piotr
  Doll{\'a}r, and C~Lawrence Zitnick.
\newblock Microsoft coco captions: Data collection and evaluation server.
\newblock {\em arXiv preprint arXiv:1504.00325}, 2015.

\bibitem{couairon2022diffedit}
Guillaume Couairon, Jakob Verbeek, Holger Schwenk, and Matthieu Cord.
\newblock Diffedit: Diffusion-based semantic image editing with mask guidance.
\newblock {\em arXiv preprint arXiv:2210.11427}, 2022.

\bibitem{creswell2018inverting}
Antonia Creswell and Anil~Anthony Bharath.
\newblock Inverting the generator of a generative adversarial network.
\newblock {\em IEEE transactions on neural networks and learning systems},
  30(7):1967--1974, 2018.

\bibitem{katherine2021vqganclip}
Katherine Crowson.
\newblock Vqgan + clip, 2021.
\newblock
  \url{https://colab.research.google.com/drive/1L8oL-vLJXVcRzCFbPwOoMkPKJ8-aYdPN}.

\bibitem{crowson2022vqgan}
Katherine Crowson, Stella Biderman, Daniel Kornis, Dashiell Stander, Eric
  Hallahan, Louis Castricato, and Edward Raff.
\newblock Vqgan-clip: Open domain image generation and editing with natural
  language guidance.
\newblock {\em arXiv preprint arXiv:2204.08583}, 2022.

\bibitem{dhariwal2021diffusion}
Prafulla Dhariwal and Alexander Nichol.
\newblock Diffusion models beat gans on image synthesis.
\newblock {\em Advances in Neural Information Processing Systems},
  34:8780--8794, 2021.

\bibitem{esser2020taming}
Patrick Esser, Robin Rombach, and Björn Ommer.
\newblock Taming transformers for high-resolution image synthesis, 2020.

\bibitem{gal2022image}
Rinon Gal, Yuval Alaluf, Yuval Atzmon, Or Patashnik, Amit~H Bermano, Gal
  Chechik, and Daniel Cohen-Or.
\newblock An image is worth one word: Personalizing text-to-image generation
  using textual inversion.
\newblock {\em arXiv preprint arXiv:2208.01618}, 2022.

\bibitem{hertz2022prompt}
Amir Hertz, Ron Mokady, Jay Tenenbaum, Kfir Aberman, Yael Pritch, and Daniel
  Cohen-Or.
\newblock Prompt-to-prompt image editing with cross attention control.
\newblock {\em arXiv preprint arXiv:2208.01626}, 2022.

\bibitem{ho2020denoising}
Jonathan Ho, Ajay Jain, and Pieter Abbeel.
\newblock Denoising diffusion probabilistic models.
\newblock {\em Advances in Neural Information Processing Systems},
  33:6840--6851, 2020.

\bibitem{ho2021classifier}
Jonathan Ho and Tim Salimans.
\newblock Classifier-free diffusion guidance.
\newblock In {\em NeurIPS 2021 Workshop on Deep Generative Models and
  Downstream Applications}, 2021.

\bibitem{Kawar2022ImagicTR}
Bahjat Kawar, Shiran Zada, Oran Lang, Omer Tov, Hui-Tang Chang, Tali Dekel,
  Inbar Mosseri, and Michal Irani.
\newblock Imagic: Text-based real image editing with diffusion models.
\newblock {\em ArXiv}, abs/2210.09276, 2022.

\bibitem{kim2022diffusionclip}
Gwanghyun Kim, Taesung Kwon, and Jong~Chul Ye.
\newblock Diffusionclip: Text-guided diffusion models for robust image
  manipulation.
\newblock In {\em Proceedings of the IEEE/CVF Conference on Computer Vision and
  Pattern Recognition}, pages 2426--2435, 2022.

\bibitem{kwon2021clipstyler}
Gihyun Kwon and Jong~Chul Ye.
\newblock Clipstyler: Image style transfer with a single text condition.
\newblock {\em arXiv preprint arXiv:2112.00374}, 2021.

\bibitem{lipton2017precise}
Zachary~C Lipton and Subarna Tripathi.
\newblock Precise recovery of latent vectors from generative adversarial
  networks.
\newblock {\em arXiv preprint arXiv:1702.04782}, 2017.

\bibitem{meng2021sdedit}
Chenlin Meng, Yang Song, Jiaming Song, Jiajun Wu, Jun-Yan Zhu, and Stefano
  Ermon.
\newblock Sdedit: Image synthesis and editing with stochastic differential
  equations.
\newblock {\em arXiv preprint arXiv:2108.01073}, 2021.

\bibitem{mokady2021clipcap}
Ron Mokady, Amir Hertz, and Amit~H Bermano.
\newblock Clipcap: Clip prefix for image captioning.
\newblock {\em arXiv preprint arXiv:2111.09734}, 2021.

\bibitem{nichol2021glide}
Alex Nichol, Prafulla Dhariwal, Aditya Ramesh, Pranav Shyam, Pamela Mishkin,
  Bob McGrew, Ilya Sutskever, and Mark Chen.
\newblock Glide: Towards photorealistic image generation and editing with
  text-guided diffusion models.
\newblock {\em arXiv preprint arXiv:2112.10741}, 2021.

\bibitem{radford2021learning}
Alec Radford, Jong~Wook Kim, Chris Hallacy, Aditya Ramesh, Gabriel Goh,
  Sandhini Agarwal, Girish Sastry, Amanda Askell, Pamela Mishkin, Jack Clark,
  et~al.
\newblock Learning transferable visual models from natural language
  supervision.
\newblock {\em arXiv preprint arXiv:2103.00020}, 2021.

\bibitem{ramesh2022hierarchical}
Aditya Ramesh, Prafulla Dhariwal, Alex Nichol, Casey Chu, and Mark Chen.
\newblock Hierarchical text-conditional image generation with clip latents.
\newblock {\em arXiv preprint arXiv:2204.06125}, 2022.

\bibitem{richardson2020encoding}
Elad Richardson, Yuval Alaluf, Or Patashnik, Yotam Nitzan, Yaniv Azar, Stav
  Shapiro, and Daniel Cohen-Or.
\newblock Encoding in style: a stylegan encoder for image-to-image translation.
\newblock {\em arXiv preprint arXiv:2008.00951}, 2020.

\bibitem{roich2021pivotal}
Daniel Roich, Ron Mokady, Amit~H. Bermano, and Daniel Cohen-Or.
\newblock Pivotal tuning for latent-based editing of real images.
\newblock {\em ACM Transactions on Graphics (TOG)}, 2022.

\bibitem{rombach2021highresolution}
Robin Rombach, Andreas Blattmann, Dominik Lorenz, Patrick Esser, and Björn
  Ommer.
\newblock High-resolution image synthesis with latent diffusion models, 2021.

\bibitem{ruiz2022dreambooth}
Nataniel Ruiz, Yuanzhen Li, Varun Jampani, Yael Pritch, Michael Rubinstein, and
  Kfir Aberman.
\newblock Dreambooth: Fine tuning text-to-image diffusion models for
  subject-driven generation.
\newblock {\em arXiv preprint arXiv:2208.12242}, 2022.

\bibitem{saharia2022photorealistic}
Chitwan Saharia, William Chan, Saurabh Saxena, Lala Li, Jay Whang, Emily
  Denton, Seyed Kamyar~Seyed Ghasemipour, Burcu~Karagol Ayan, S~Sara Mahdavi,
  Rapha~Gontijo Lopes, Tim Salimans, Tim Salimans, Jonathan Ho, David~J Fleet,
  and Mohammad Norouzi.
\newblock Photorealistic text-to-image diffusion models with deep language
  understanding.
\newblock {\em arXiv preprint arXiv:2205.11487}, 2022.

\bibitem{sheynin2022knn}
Shelly Sheynin, Oron Ashual, Adam Polyak, Uriel Singer, Oran Gafni, Eliya
  Nachmani, and Yaniv Taigman.
\newblock Knn-diffusion: Image generation via large-scale retrieval.
\newblock {\em arXiv preprint arXiv:2204.02849}, 2022.

\bibitem{sohl2015deep}
Jascha Sohl-Dickstein, Eric Weiss, Niru Maheswaranathan, and Surya Ganguli.
\newblock Deep unsupervised learning using nonequilibrium thermodynamics.
\newblock In {\em International Conference on Machine Learning}, pages
  2256--2265. PMLR, 2015.

\bibitem{song2020denoising}
Jiaming Song, Chenlin Meng, and Stefano Ermon.
\newblock Denoising diffusion implicit models.
\newblock In {\em International Conference on Learning Representations}, 2020.

\bibitem{song2019generative}
Yang Song and Stefano Ermon.
\newblock Generative modeling by estimating gradients of the data distribution.
\newblock {\em Advances in Neural Information Processing Systems}, 32, 2019.

\bibitem{stefanini2022show}
Matteo Stefanini, Marcella Cornia, Lorenzo Baraldi, Silvia Cascianelli,
  Giuseppe Fiameni, and Rita Cucchiara.
\newblock From show to tell: a survey on deep learning-based image captioning.
\newblock {\em IEEE Transactions on Pattern Analysis and Machine Intelligence},
  2022.

\bibitem{tov2021designing}
Omer Tov, Yuval Alaluf, Yotam Nitzan, Or Patashnik, and Daniel Cohen-Or.
\newblock Designing an encoder for stylegan image manipulation.
\newblock {\em arXiv preprint arXiv:2102.02766}, 2021.

\bibitem{valevski2022unitune}
Dani Valevski, Matan Kalman, Yossi Matias, and Yaniv Leviathan.
\newblock Unitune: Text-driven image editing by fine tuning an image generation
  model on a single image.
\newblock {\em arXiv preprint arXiv:2210.09477}, 2022.

\bibitem{Wang2021HighFidelityGI}
Tengfei Wang, Yong Zhang, Yanbo Fan, Jue Wang, and Qifeng Chen.
\newblock High-fidelity gan inversion for image attribute editing.
\newblock {\em ArXiv}, abs/2109.06590, 2021.

\bibitem{xia2021gan}
Weihao Xia, Yulun Zhang, Yujiu Yang, Jing-Hao Xue, Bolei Zhou, and Ming-Hsuan
  Yang.
\newblock Gan inversion: A survey, 2021.

\bibitem{yeh2017semantic}
Raymond~A. Yeh, Chen Chen, Teck~Yian Lim, Alexander~G. Schwing, Mark
  Hasegawa-Johnson, and Minh~N. Do.
\newblock Semantic image inpainting with deep generative models, 2017.

\bibitem{Zhang2018TheUE}
Richard Zhang, Phillip Isola, Alexei~A. Efros, Eli Shechtman, and Oliver Wang.
\newblock The unreasonable effectiveness of deep features as a perceptual
  metric.
\newblock {\em 2018 IEEE/CVF Conference on Computer Vision and Pattern
  Recognition}, pages 586--595, 2018.

\bibitem{zhu2016generative}
Jun-Yan Zhu, Philipp Kr{\"a}henb{\"u}hl, Eli Shechtman, and Alexei~A Efros.
\newblock Generative visual manipulation on the natural image manifold.
\newblock In {\em European conference on computer vision}, pages 597--613.
  Springer, 2016.

\end{thebibliography}
}

\newpage

\appendix

\begin{figure}
     \centering
     \vspace{-0.4cm}
       \begin{subfigure}[b]{0.48\columnwidth}
         \centering
         \includegraphics[width=\columnwidth]{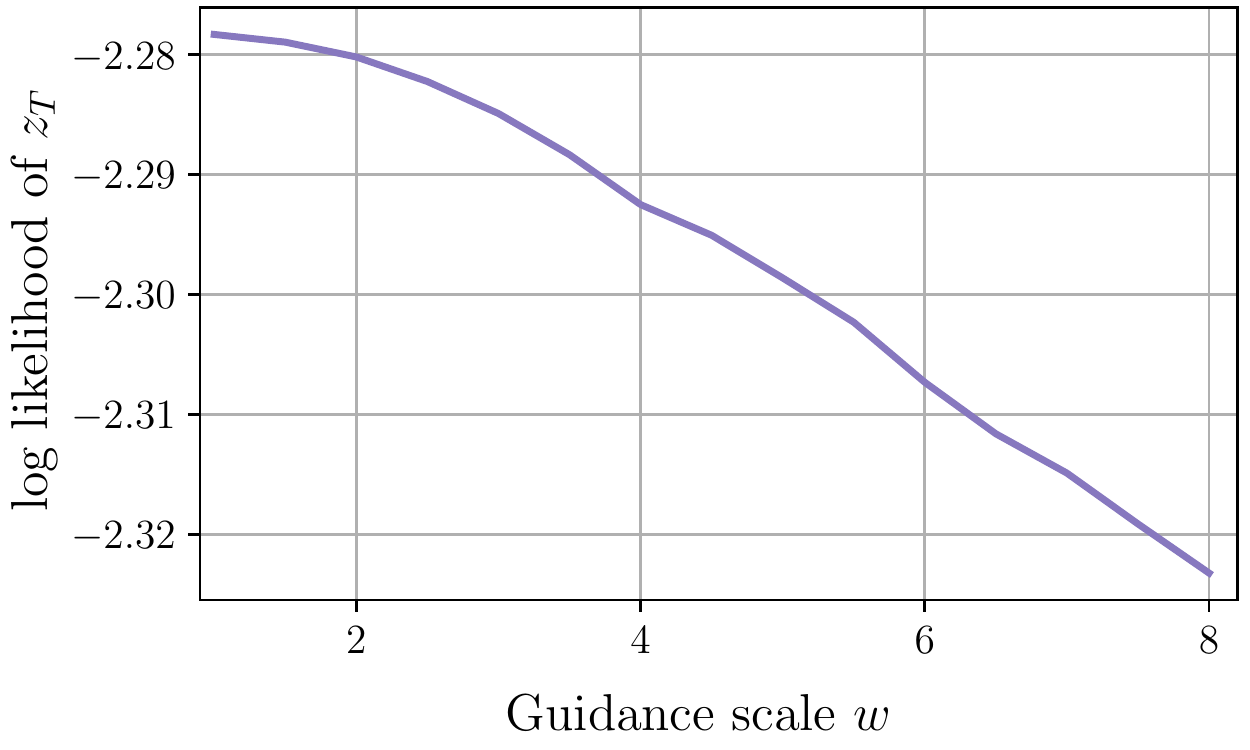}
         \caption{Edibility}
         \label{fig:ddim log-likelihood}
     \end{subfigure}
      \hfill
     \begin{subfigure}[b]{0.48\columnwidth}
         \centering
         \includegraphics[width=\columnwidth]{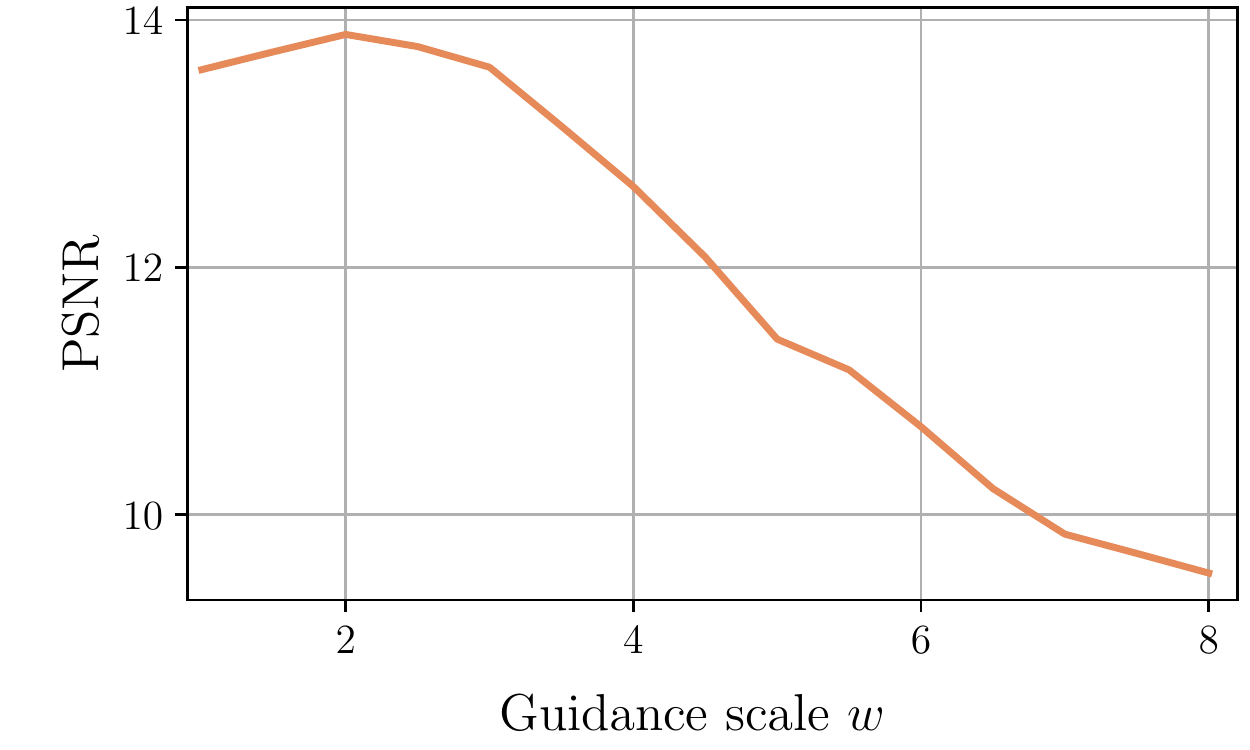}
         \caption{Reconstruction}
         \label{fig:ddim reconstcution}
     \end{subfigure}
        \caption{{\bf Setting the guidance scale for DDIM.} {\it  We evaluate the DDIM inversion with different values of the guidance scale. 
        On left, we measure the log-likelihood of the latent vector $z_T$ with respect to multivariate normal distribution. This estimates the editability as $z_T$ should ideally distribute normally and deviation from this distribution reduces our ability to edit the image. On right, we measure the reconstruction quality using PSNR. As can be seen, using a small guidance scale, such as $w=1$, results in better editability and reconstruction.}}
        \vspace{-0.2cm}
        \label{fig:ddim_graphs}
\end{figure}

\noindent{\Large \textbf{{Appendix}}}

\section{Societal Impact}

Our work suggests a new editing technique for manipulating real images using state-of-the-art text-to-image diffusion models.
This modification of real photos might be exploited by malicious parties to produce fake content in order to spread disinformation.
This is a known problem, common to all image editing techniques.
However, research in identifying and preventing malicious editing is already making significant progress. We believe our work would contribute to this line of work, since we provide an analysis of the inversion and editing procedures using text-to-image diffusion models.

\section{Ablation Study}
\label{sec:ablation_supp}

\paragraph{DDIM Inversion.}
To validate our selection of the guidance scale parameter of $w=1$ during the DDIM Inversion (see Algorithm 1, line 3, in the main text), we conduct the DDIM inversion with different values of $w$ from $1$ to $8$ using the same data as in Section 4.
For each inversion, we measure the log-likelihood of the result latent image $z^*_T \in \mathcal{R}^{64\times 64 \times 4}$ under the standard multivariate normal distribution. Intuitively, to achieve high edibility we would like to maximize this term since during training $z^*_T$ distributes normally. The mean log-likelihood as a function of $w$ is plotted in \cref{fig:ddim log-likelihood}. In addition, we measure the reconstruction with respect to the ground truth input image using the PSNR metric.
%The results are shown in~\cref{fig:ddim_graphs} 
As can be seen in \cref{fig:ddim reconstcution}, increasing the value of $w$ results in less editable latent vector $z^*_T$ and poorer initial reconstruction for our optimization, and therefore we use $w=1$.

\begin{figure*}[t]
\centering 
\includegraphics[width=\textwidth]{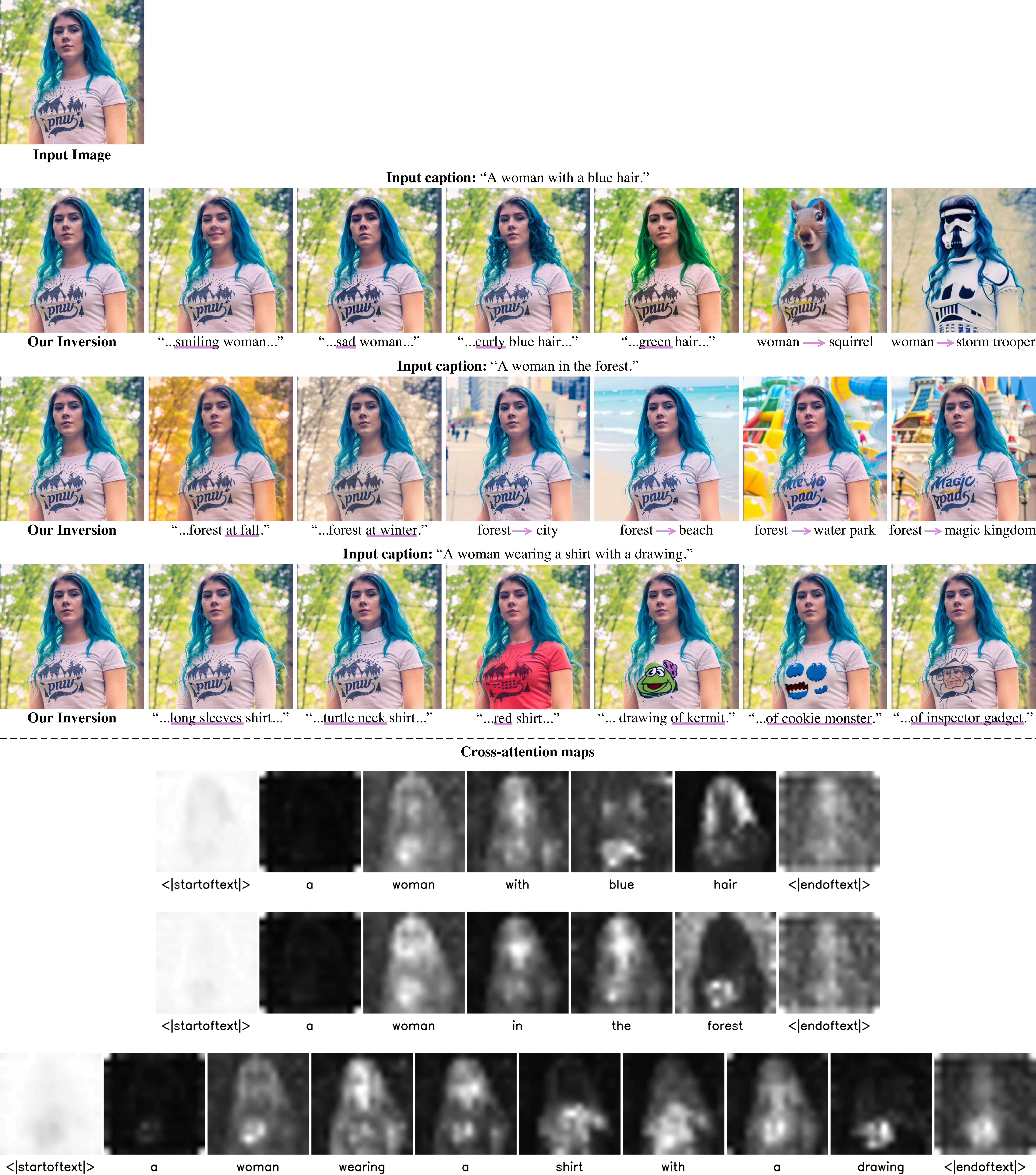} 

\caption{{\bf Robustness to the input caption}. {\it  We can invert an input image (top) using different input captions (first column). Naturally, the selection of the caption effects the editing abilities with Prompt-to-Prompt, as can be seen in the visualization of the cross-attention map (bottom). Yet, our method is not particularly sensitive to the exact wording of the prompt.} }
\label{fig:exp_multi_cap} 
\end{figure*}

\paragraph{Robustness to different input captions.}
In \cref{fig:exp_multi_cap} (top), we demonstrate our robustness to different input captions by successfully inverting an image using multiple captions. Yet, the edited parts should be included in the source caption in order to produce semantic attention maps for these (\cref{fig:exp_multi_cap} bottom). For example, to edit the print on the shirt, the source caption should include a "shirt with a drawing" term or a similar one.

% \paragraph{Additional visual results.}  \cref{fig:exp_ablation_qual_sup,fig:exp_ablation_qual_sup2} show how our method converges to high-quality reconstruction more efficiently.

\paragraph{Null-text optimization without pivotal inversion.}

\begin{figure*}
\centering 
\includegraphics[width=\textwidth]{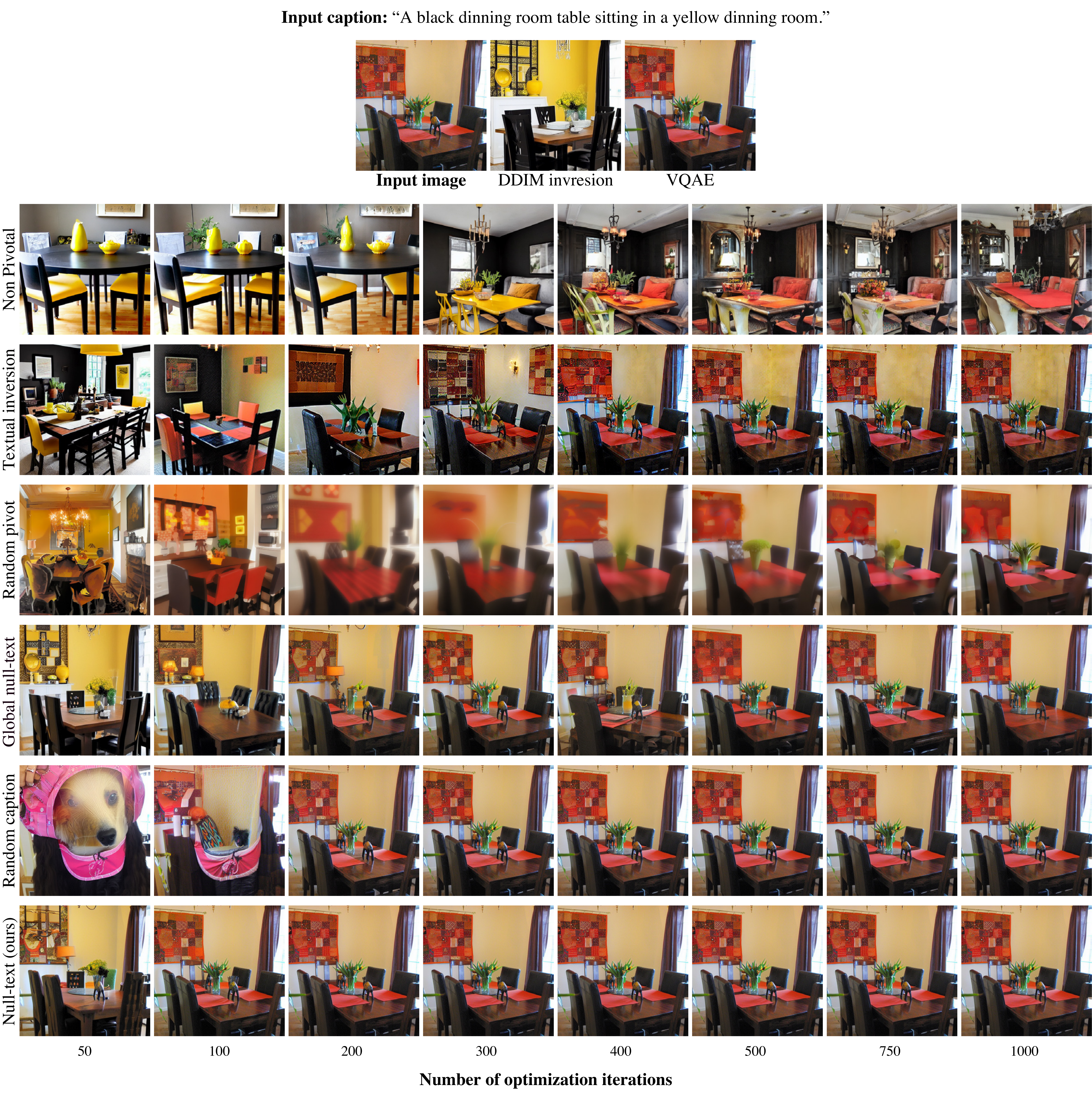} 

\caption{{\bf Ablation study.} {\it We show the inversion results for an increasing number of optimization iterations. Our method achieves high-quality reconstruction with fewer optimization steps.}} 
\label{fig:exp_ablation_qual_sup} 
\end{figure*}

\begin{figure*}
\centering 
\includegraphics[width=\textwidth]{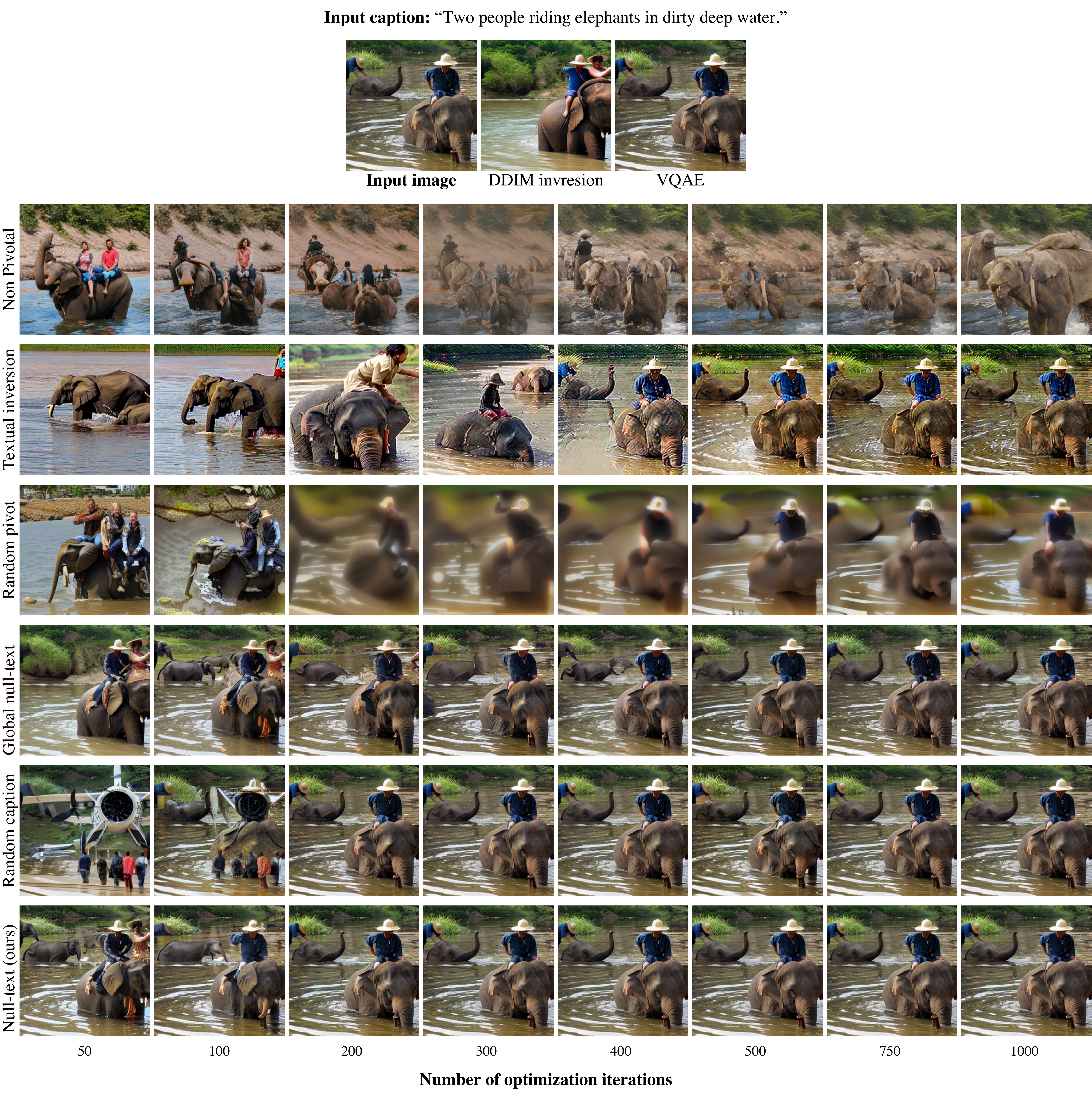} 

\caption{{\bf Ablation study.} {\it We show the inversion results for an increasing number of optimization iterations. Our method achieves high-quality reconstruction with fewer optimization steps.}} 
\label{fig:exp_ablation_qual_sup2} 
\end{figure*}

Optimizing the null-text embedding fails without the efficient pivotal inversion. This is demonstrated in \cref{fig:exp_ablation_qual_sup} and \ref{fig:exp_ablation_qual_sup2}, where the non-pivotal null-text optimization produces low-quality reconstruction ($2$nd row).

\begin{figure*}
\setlength{\tabcolsep}{2.5pt}
    \centering
    { %\scriptsize %\footnotesize

{ \bf Attention maps of Text embedding optimization + Pivotal Inversion} 
{\includegraphics[width=\textwidth]{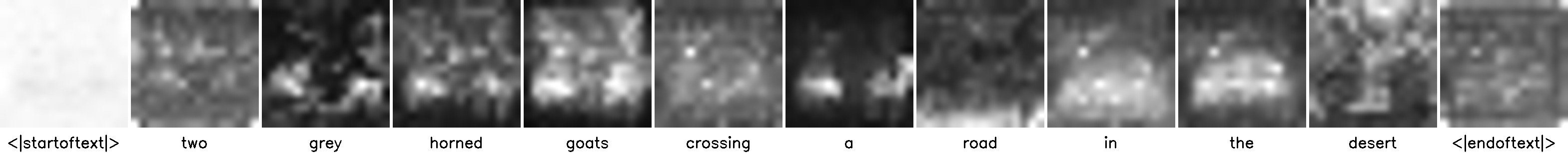}}

{ \bf Attention maps of null-text optimization} 
{\includegraphics[width=\textwidth]{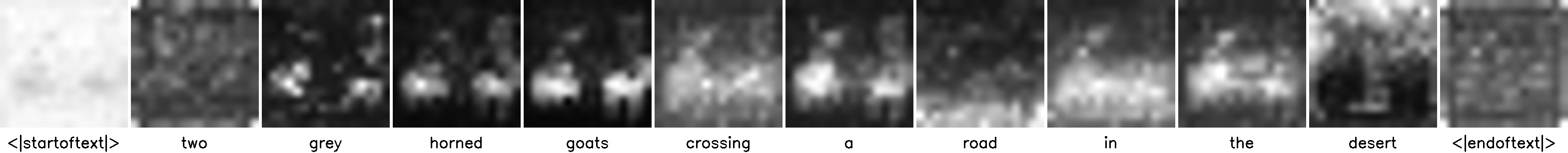}}

\begin{tabular}{c c c c c c}

{ \bf Input}&
{ \bf Inversion (T+P)}&
{ \bf Text + Pivot}&
{ \bf Ours}&
{ \bf Text + Pivot}&
{ \bf Ours} \\

{\includegraphics[width=0.16\textwidth]{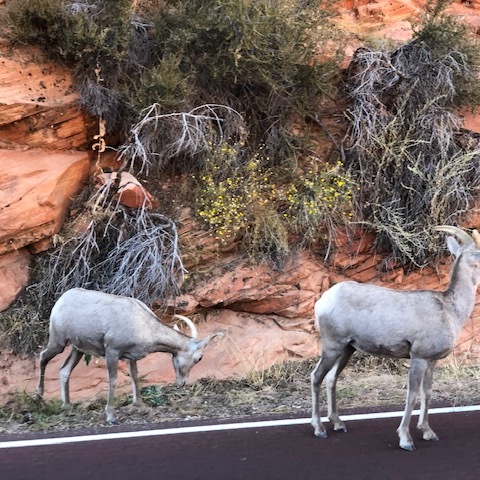}}&
{\includegraphics[width=0.16\textwidth]{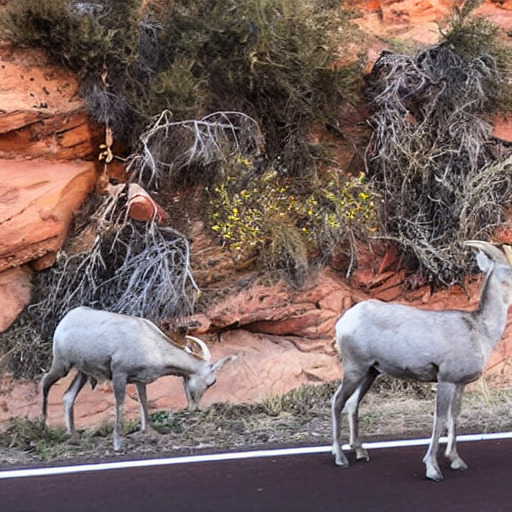}}&
{\includegraphics[width=0.16\textwidth]{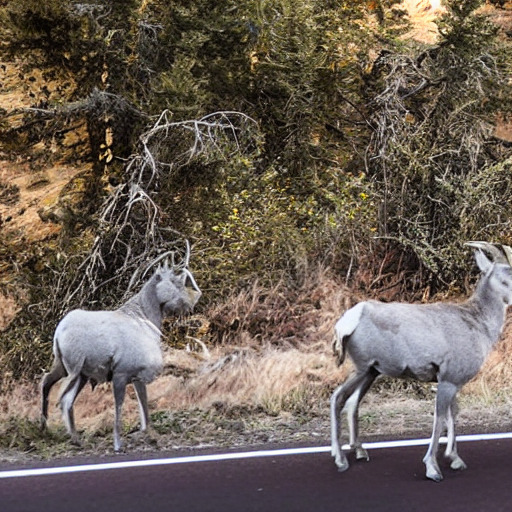}}&
{\includegraphics[width=0.16\textwidth]{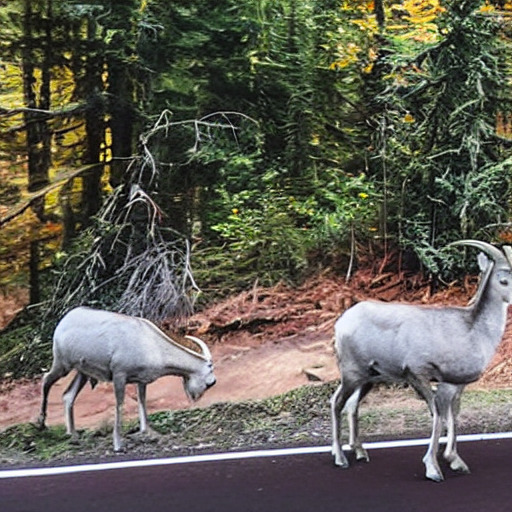}}&
{\includegraphics[width=0.16\textwidth]{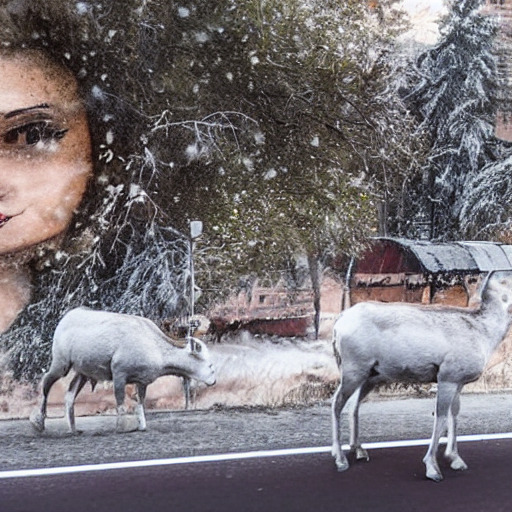}} &
{\includegraphics[width=0.16\textwidth]{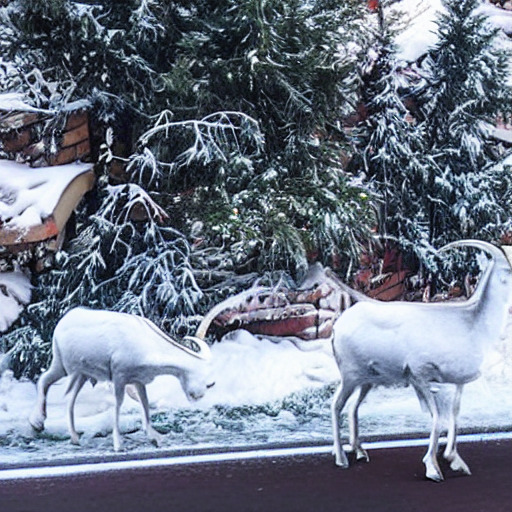}} \\

%  & \multicolumn{2}{c}{"Two grey horned goats crossing a road in the \st{desert} {\color{RoyalPurple} \bf forest} "} &
%  \multicolumn{2}{c}{"Two grey horned goats crossing a road in the \st{desert} {\color{RoyalPurple} \bf snow} "} \\

 & & \multicolumn{2}{c}{ desert $\longrightarrow$ {\color{RoyalPurple} \bf forest} "} &
\multicolumn{2}{c}{desert $\longrightarrow$ {\color{RoyalPurple} \bf snow} "} \\

% \midrule

\end{tabular}
}

\vspace{-0.2cm}
\caption{{\bf Ablation study - Textual inversion with a pivot.} {\it We compare our method to replacing the null-text optimization with optimizing the conditional (textual) embedding while still applying pivotal inversion. As can be seen (top), this results in less accurate attention maps, and thus, in less accurate editing capabilities. In particular, textual inversion with a pivot achieves high-fidelity reconstruction ("Inversion (T+P)"), but goat heads distort (bottom) when editing is applied due to the inaccurate attention maps.}}
\vspace{-0.5cm}
% \vspace{-0.15cm}
% \ron{The gap is best viewed in the supplementary videos.}}
\label{fig:albation_text_pivot} %\vspace{-7pt}
\end{figure*}

\paragraph{Textual inversion with a pivot.}
\cref{fig:albation_text_pivot} illustrate performing textual inversion around a pivot, i.e., similar to our pivotal inversion but optimizing the conditioned embedding. This results in a comparable reconstruction to ours, as demonstrated in \cref{fig:albation_text_pivot} (bottom), but editability is reduced. By analyzing the attention maps (\cref{fig:albation_text_pivot}, top), observing that these are less accurate than ours. For example, using our null-text optimization, the attention referring to "goats" is much more local, and attention referring to "desert" is more accurate. Consequently, editing the "desert" results in artifacts over the goats (\cref{fig:albation_text_pivot}, bottom).

\begin{table}
\label{tab:times}
\vspace{-0.4cm}
\caption{{\bf Inference time comparison.} {\it We measure both inversion and editing time for different methods. SDEdit is faster than ours, as an inversion is not employed by default, but fails to preserve the unedited parts. Our method is more efficient than the rest of the baselines, as it provides accurate reconstruction with faster inversion time, while also allowing multiple editing operations after a single inversion. }}
\vspace{-0.2cm}
\centering
\begin{tabular}{l c c c}
\toprule
Method & Inversion  &  Editing  & Multiple edits \\
\midrule
VQGAN + CLIP & --- & $\sim 1$m & No \\
Text2Live & --- & $\sim 9$m & No \\
SDEdit & --- & 10s & \textbf{Yes} \\
Imagic & $\sim 5$m & 10s & No \\
Ours & $\sim 1$m & 10s & \textbf{Yes} \\
\bottomrule
\end{tabular}
\label{tab:times}
\end{table}

% \begin{table}
% \label{tab:times}
% \caption{Optimization time for text based image editing methods.}
% \centering
% \begin{tabular}{l c c}
% \toprule
% Method & Optimization time & Multiple edits \\
% \midrule
% VQGAN + CLIP & 1m & No \\
% Text2Live & 9m & No \\
% SDEdit & 0m & \textbf{Yes} \\
% Imagic & 5m & No \\
% Ours & 2m & \textbf{Yes} \\
% \bottomrule
% \end{tabular}
% \end{table}

\section{Additional results}
\label{sec:supp_results}

\begin{figure}[t]
\setlength{\tabcolsep}{0.5pt}
    %\centering
    { \scriptsize %\footnotesize

% \begin{tabular}{p{0.25\columnwidth} @{\hskip 0.04in} p{0.25\columnwidth} p{0.25\columnwidth} p{0.25\columnwidth}}
\vspace{-0.4cm}
\begin{tabular}{c c c c}

\multicolumn{4}{c}{"A living room with a couch and pillows"} \\

{\includegraphics[width=0.25\columnwidth]{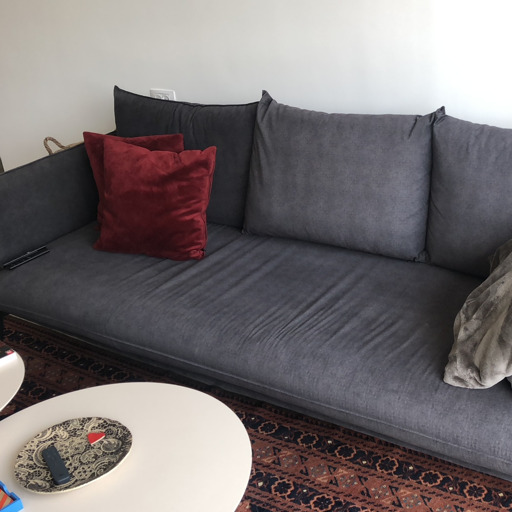}}&
{\includegraphics[width=0.25\columnwidth]{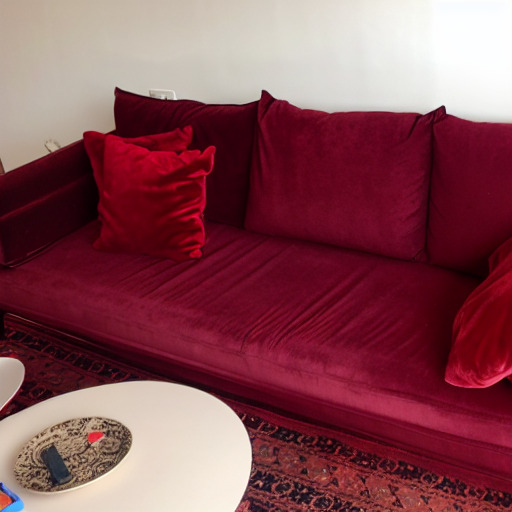}} &
{\includegraphics[width=0.25\columnwidth]{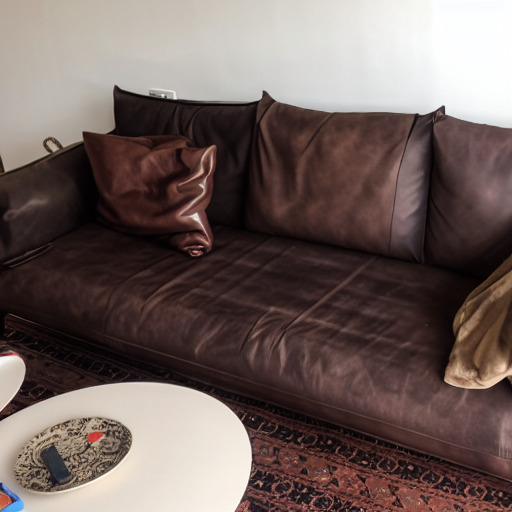}} &
{\includegraphics[width=0.25\columnwidth]{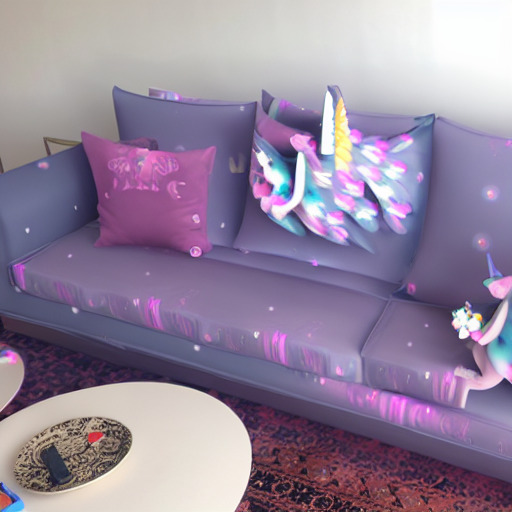}} \\

\multicolumn{1}{c}{Input} &
{\color{RoyalPurple} \bf red velvet} couch &
{\color{RoyalPurple} \bf leather} couch &
{\color{RoyalPurple} \bf unicorn} couch \\

\midrule
\multicolumn{4}{c}{"close up of a giraffe eating a bucket"} \\

{\includegraphics[width=0.25\columnwidth]{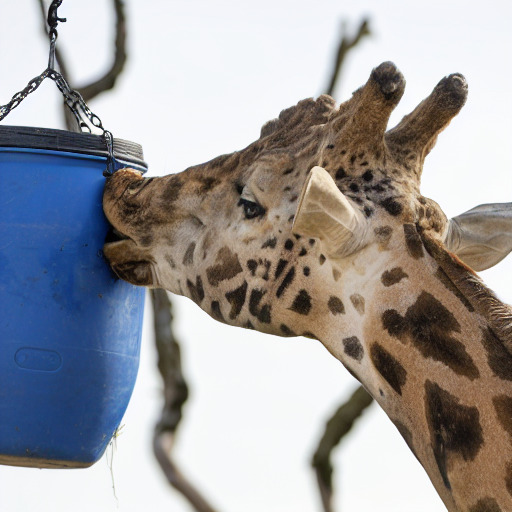}} &
{\includegraphics[width=0.25\columnwidth]{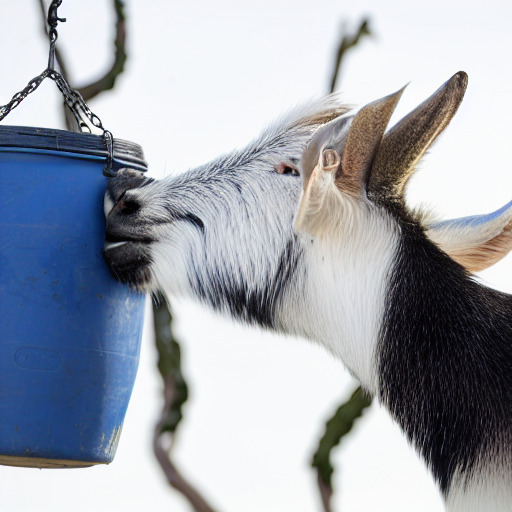}} &
{\includegraphics[width=0.25\columnwidth]{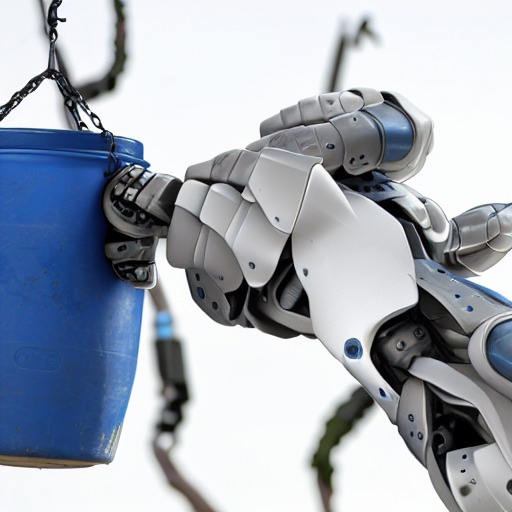}} &
{\includegraphics[width=0.25\columnwidth]{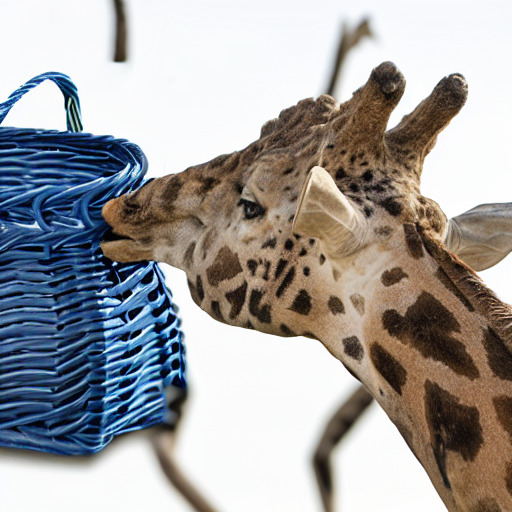}} \\

\multicolumn{1}{c}{Input}&
giraffe $\rightarrow$ "{\color{RoyalPurple} \bf goat} &
giraffe $\rightarrow$ "{\color{RoyalPurple} \bf robot}&
bucket $\rightarrow$ "{\color{RoyalPurple} \bf basket} \\

\midrule
\multicolumn{4}{c}{"A piece of cake"} \\

{\includegraphics[width=0.25\columnwidth]{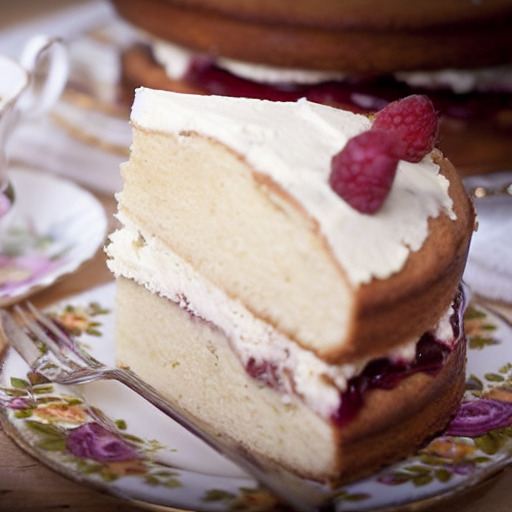}} &
{\includegraphics[width=0.25\columnwidth]{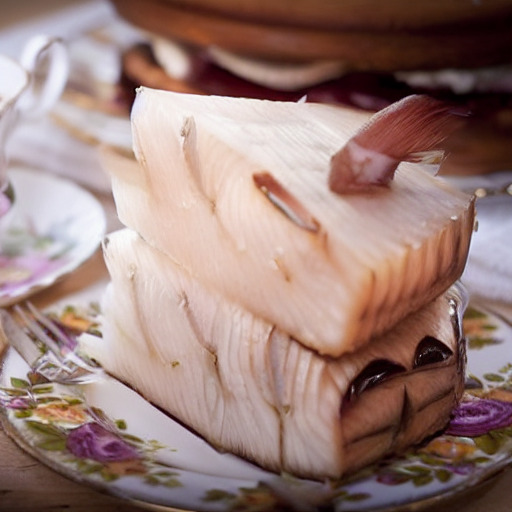}} &
{\includegraphics[width=0.25\columnwidth]{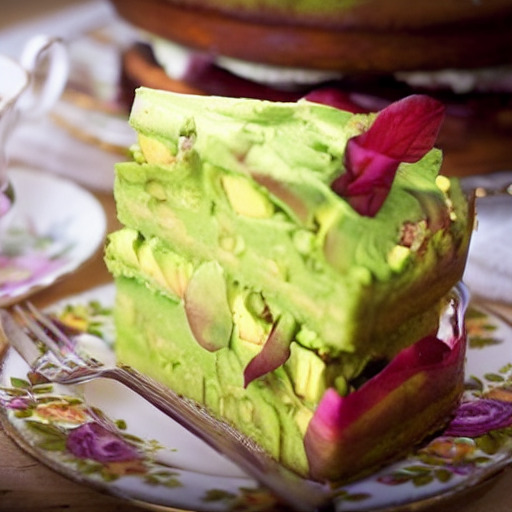}} &
{\includegraphics[width=0.25\columnwidth]{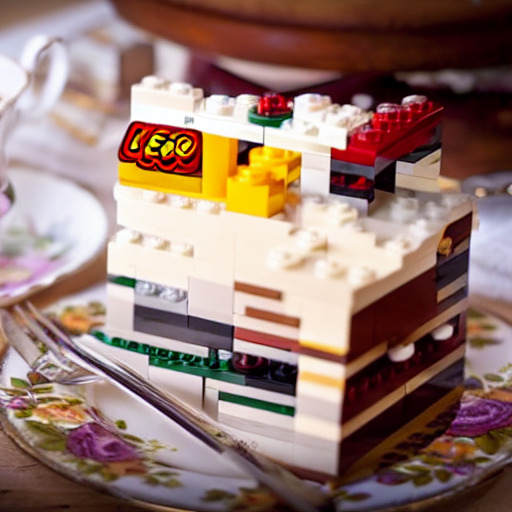}} \\

\multicolumn{1}{c}{Input}&
{\color{RoyalPurple} \bf fish} cake &
{\color{RoyalPurple} \bf avocado} cake &
{\color{RoyalPurple} \bf Lego} cake  \\

\midrule
\multicolumn{4}{c}{"A basket with apples on a chair"} \\

{\includegraphics[width=0.25\columnwidth]{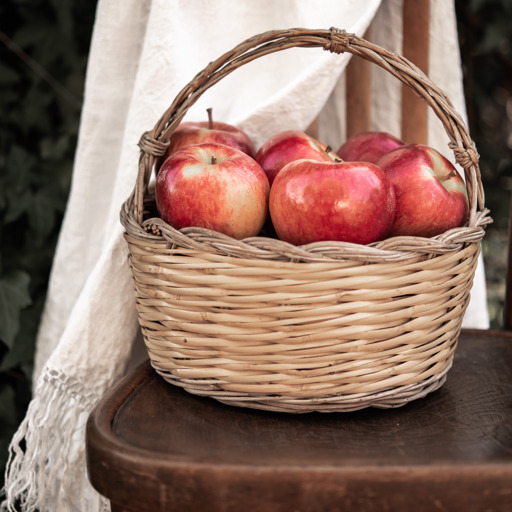}} &
{\includegraphics[width=0.25\columnwidth]{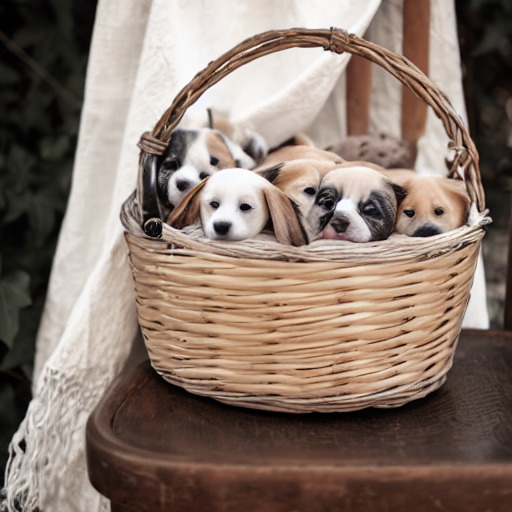}} &
{\includegraphics[width=0.25\columnwidth]{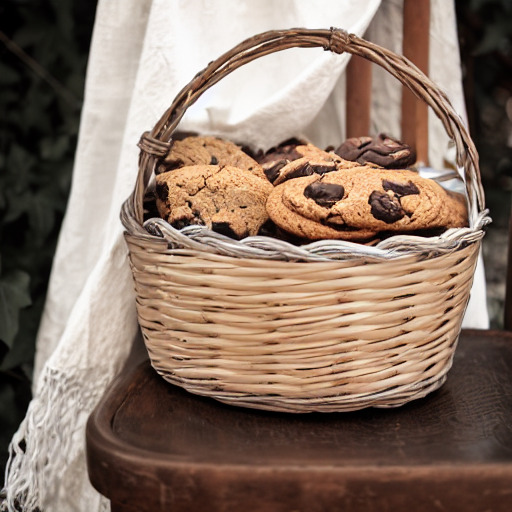}} &
{\includegraphics[width=0.25\columnwidth]{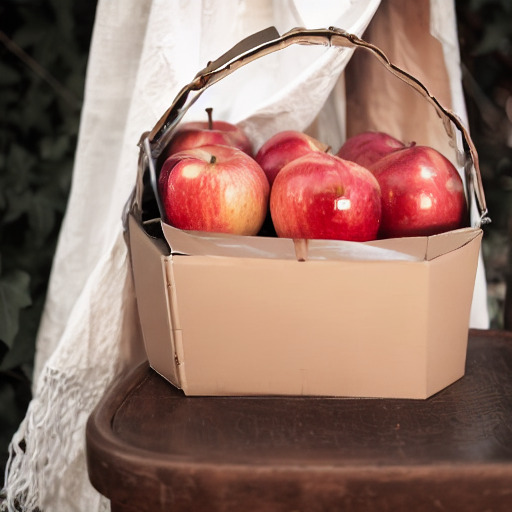}} \\

\multicolumn{1}{c}{Input}&
apples$\rightarrow$ {\color{RoyalPurple}\bf puppies} &
apples$\rightarrow$ {\color{RoyalPurple} \bf cookies} &
{\color{RoyalPurple} \bf cardboard} basket \\

% \midrule
% \multicolumn{4}{c}{"A child is climbing on a tree"} \\

% {\includegraphics[width=0.25\columnwidth]{images/gt/kid_tree.jpg}} &
% {\includegraphics[width=0.25\columnwidth]{images/ours/kid_tree_097.jpg}} &
% {\includegraphics[width=0.25\columnwidth]{images/ours/kid_tree_098.jpg}} &
% {\includegraphics[width=0.25\columnwidth]{images/ours/kid_tree_099.jpg}} \\

% \multicolumn{1}{c}{Input}&
% "child"$\rightarrow$ "{\color{RoyalPurple} \bf tiger}" &
% "child"$\rightarrow$ "{\color{RoyalPurple} \bf zebra}" &
% "child"$\rightarrow$ "{\color{RoyalPurple} \bf monkey}" \\

% \midrule
% \multicolumn{4}{c}{"A girl sitting on the grass and holding a ball"} \\

% {\includegraphics[width=0.25\columnwidth]{images/gt/girl_ball_058.jpg}} &
% {\includegraphics[width=0.25\columnwidth]{images/ours/ball_057.jpg}} &
% {\includegraphics[width=0.25\columnwidth]{images/ours/ball_064.jpg}} &
% {\includegraphics[width=0.25\columnwidth]{images/ours/ball_065.jpg}} \\

% \multicolumn{1}{c}{Input}&
% "{\color{RoyalPurple} \bf golden} ball..." &
% "girl"$\rightarrow$ "{\color{RoyalPurple} \bf tiger}" &
% "girl"$\rightarrow$ "{\color{RoyalPurple} \bf dog}" \\

\midrule
\multicolumn{4}{c}{"A bicycle is parking on the side of the street"} \\

{\includegraphics[width=0.25\columnwidth]{images/gt/bicycle_039.jpg}} &
{\includegraphics[width=0.25\columnwidth]{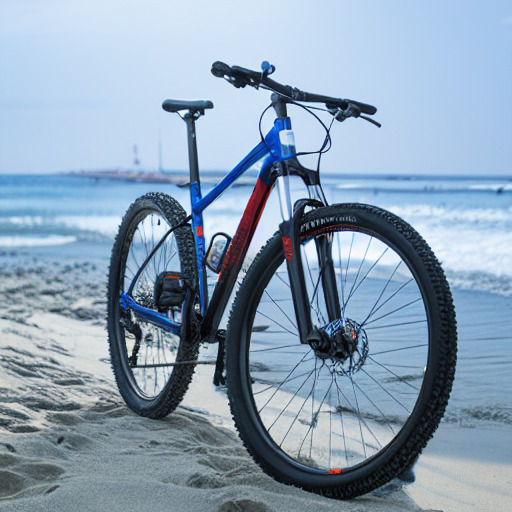}} &
{\includegraphics[width=0.25\columnwidth]{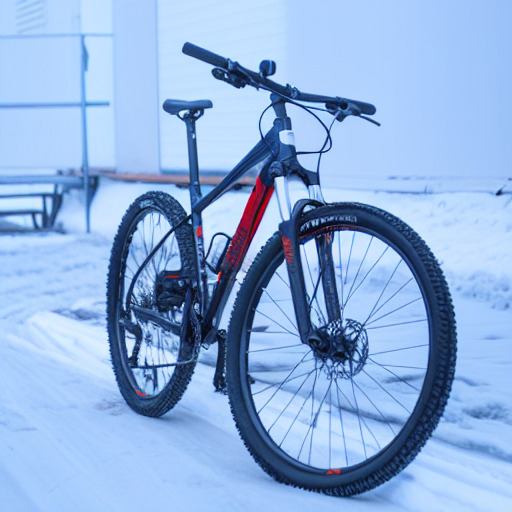}} &
{\includegraphics[width=0.25\columnwidth]{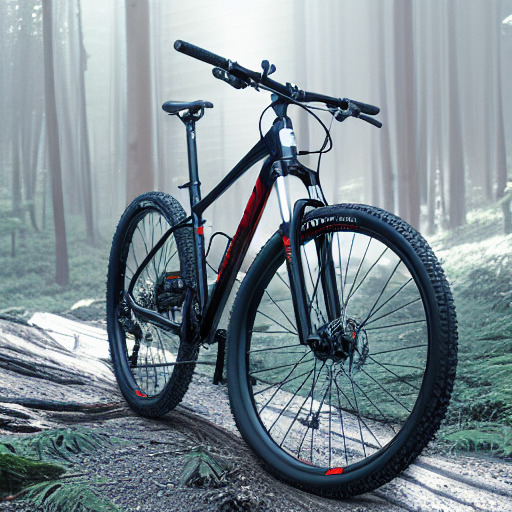}} \\

\multicolumn{1}{c}{Input}&
street$\rightarrow$ {\color{RoyalPurple} \bf beach} &
{\color{RoyalPurple} \bf snowy} street &
street$\rightarrow$ {\color{RoyalPurple} \bf forest}\\

\midrule
\multicolumn{4}{c}{"two birds sitting on a branch"} \\

{\includegraphics[width=0.25\columnwidth]{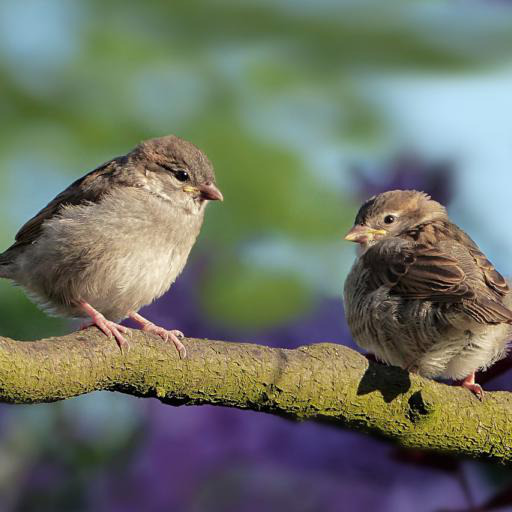}} &
{\includegraphics[width=0.25\columnwidth]{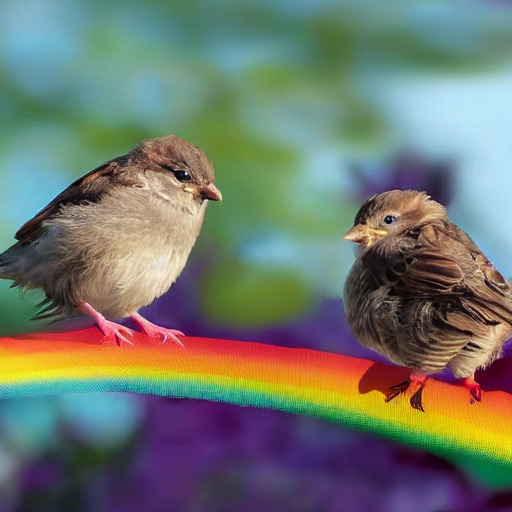}} &
{\includegraphics[width=0.25\columnwidth]{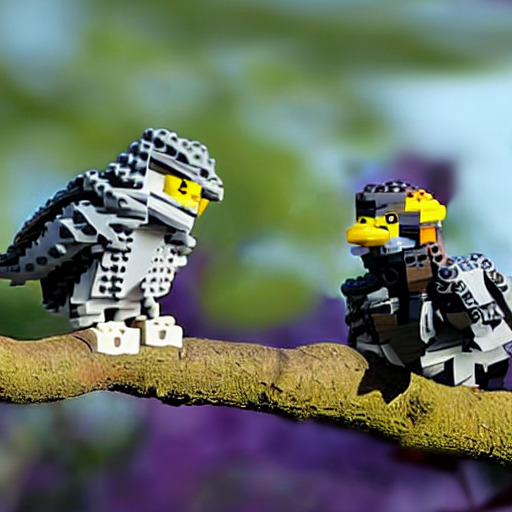}} &
{\includegraphics[width=0.25\columnwidth]{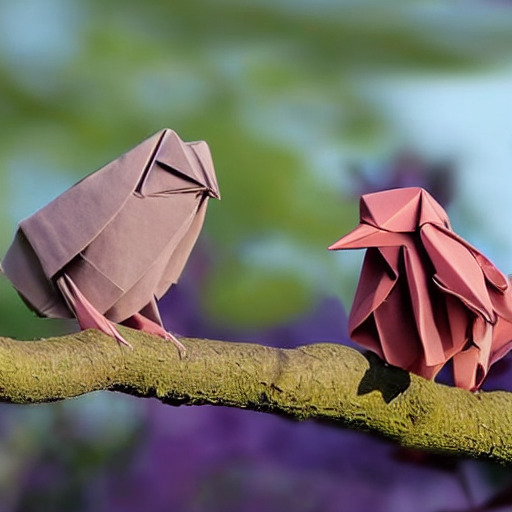}} \\

\multicolumn{1}{c}{Input}&
branch$\rightarrow$ {\color{RoyalPurple} \bf rainbow} &
{\color{RoyalPurple} \bf Lego} birds &
{\color{RoyalPurple} \bf origami} birds \\

\end{tabular}
}

\vspace{-0.1cm}
\caption{{\bf Additional editing results for our method.}} 
\vspace{-0.4cm}%\ahc{It will look better to add another 6X4 examples on right and make it wide figure.}} }
% \vspace{-0.15cm}
% \ron{The gap is best viewed in the supplementary videos.}}
\label{fig:supp_ours} %\vspace{-7pt}
\end{figure}

\begin{figure*}
\setlength{\tabcolsep}{1.5pt}
    \centering
    { %\scriptsize %\footnotesize

\begin{tabular}{c | c c c c c}

\multicolumn{1}{c}{{ \bf Input}}&
\multicolumn{1}{c}{{ \bf Our Inversion}}&
\multicolumn{1}{c}{{ \bf Text2LIVE}}&
\multicolumn{1}{c}{{ \bf VQGAN+CLIP}}&
\multicolumn{1}{c}{{ \bf SDEdit}}&
\multicolumn{1}{c}{{ \bf Our Editing}} \\

{\includegraphics[width=0.16\textwidth]{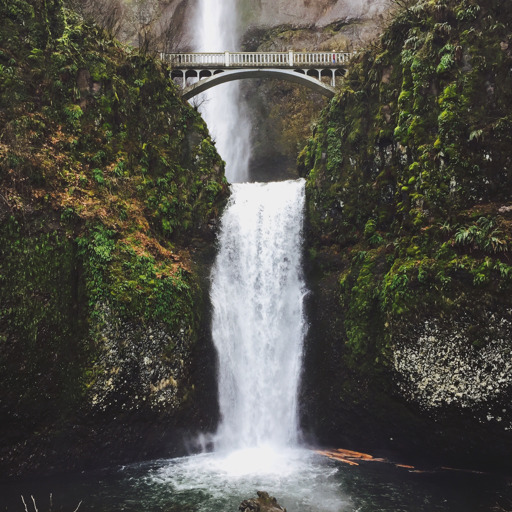}}&
{\includegraphics[width=0.16\textwidth]{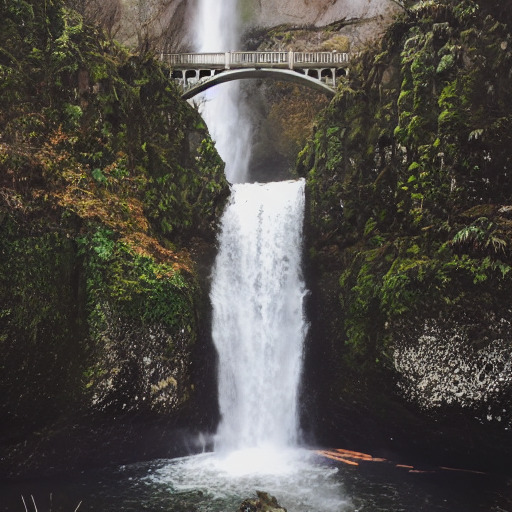}}&
{\includegraphics[width=0.16\textwidth]{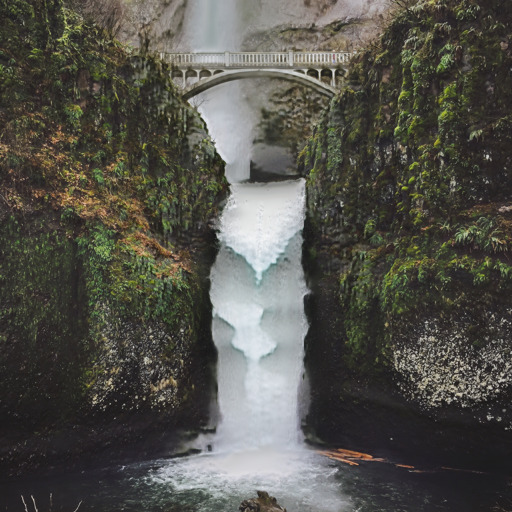}} &
{\includegraphics[width=0.16\textwidth]{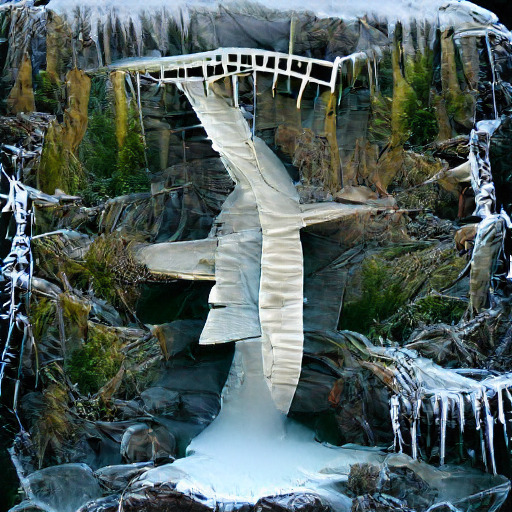}} &
{\includegraphics[width=0.16\textwidth]{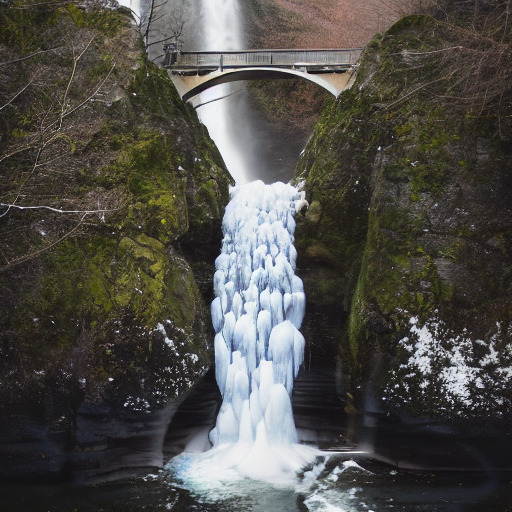}} &
{\includegraphics[width=0.16\textwidth]{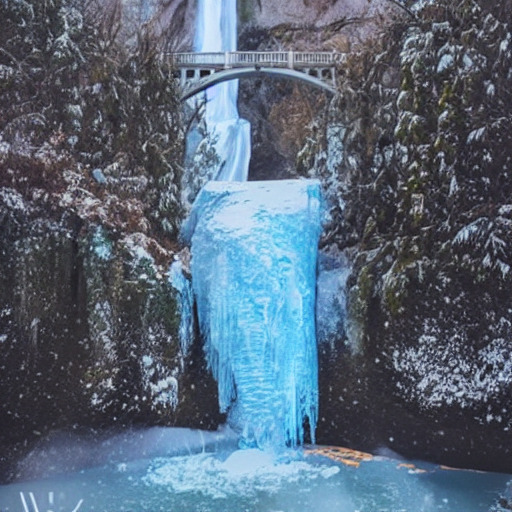}} \\

 & \multicolumn{5}{c}{"a bridge over a {\color{RoyalPurple} \bf frozen} waterfall"} \\
 &&& \\

{\includegraphics[width=0.16\textwidth]{images/gt/034.jpg}}&
{\includegraphics[width=0.16\textwidth]{images/inversion/waterfall.jpg}}&
{\includegraphics[width=0.16\textwidth]{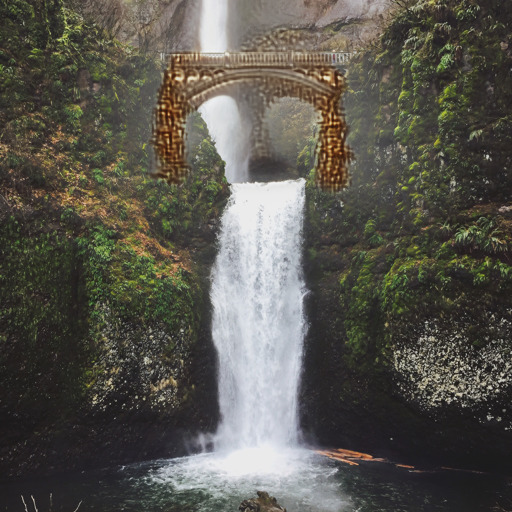}} &
{\includegraphics[width=0.16\textwidth]{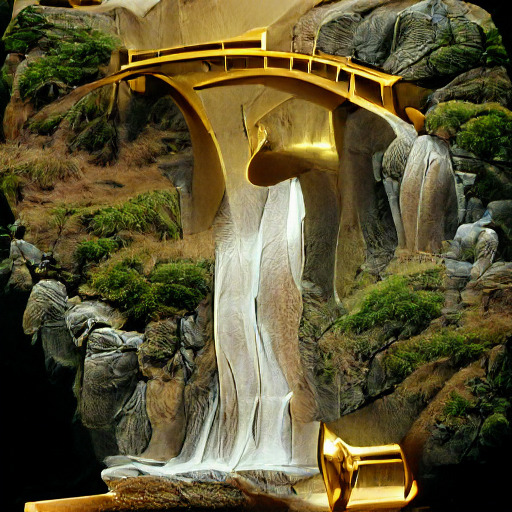}} &
{\includegraphics[width=0.16\textwidth]{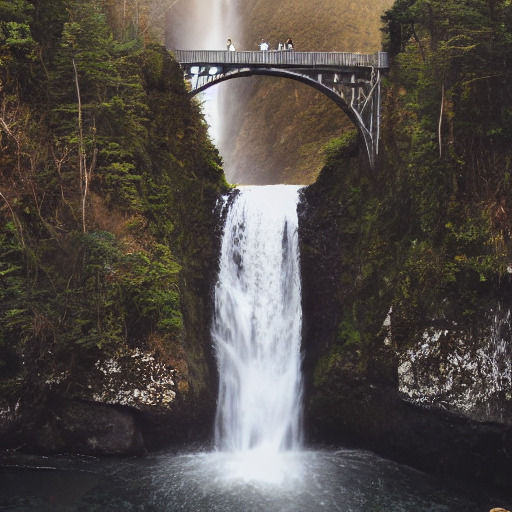}} &
{\includegraphics[width=0.16\textwidth]{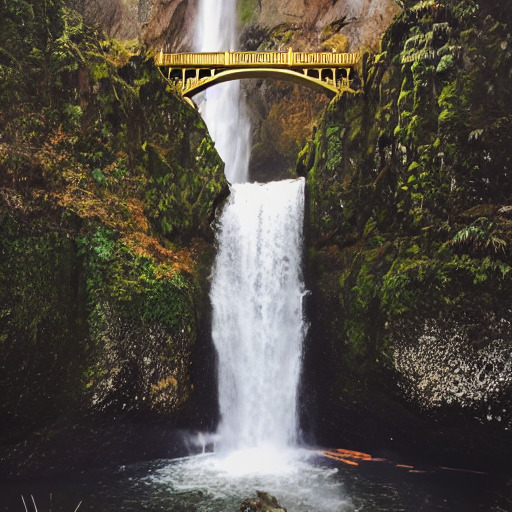}} \\

 & \multicolumn{5}{c}{"A {\color{RoyalPurple} \bf golden} bridge over a waterfall"} \\
 &&& \\

% {\includegraphics[width=0.16\textwidth]{images/gt/girl_ball_058.jpg}}&
% {\includegraphics[width=0.16\textwidth]{images/inversion/girl_t2l.jpg}}&
% {\includegraphics[width=0.16\textwidth]{images/text2live/ball_065.jpg}} &
% {\includegraphics[width=0.16\textwidth]{images/vqclip/065.jpg}} &
% {\includegraphics[width=0.16\textwidth]{images/sde/ball_065.jpg}} &
% {\includegraphics[width=0.16\textwidth]{images/ours/ball_065.jpg}} \\

%  & \multicolumn{5}{c}{"A \st{girl} {\color{RoyalPurple} \bf dog} sitting on the grass and holding a ball"} \\
% \\

% {\includegraphics[width=0.16\textwidth]{images/gt/bicycle_039.jpg}}&
% {\includegraphics[width=0.16\textwidth]{images/inversion/bicycle.jpg}}&
% {\includegraphics[width=0.16\textwidth]{images/text2live/bicycle_039.jpg}} &
% {\includegraphics[width=0.16\textwidth]{images/vqclip/039.jpg}} &
% {\includegraphics[width=0.16\textwidth]{images/sde/bicycle_039.jpg}} &
% {\includegraphics[width=0.16\textwidth]{images/ours/bicycle_039.jpg}} \\

%  & \multicolumn{5}{c}{"A {\color{RoyalPurple} \bf blue } bicycle is parking on the side of the street"} \\
% \\

{\includegraphics[width=0.16\textwidth]{images/gt/kid_tree.jpg}}&
{\includegraphics[width=0.16\textwidth]{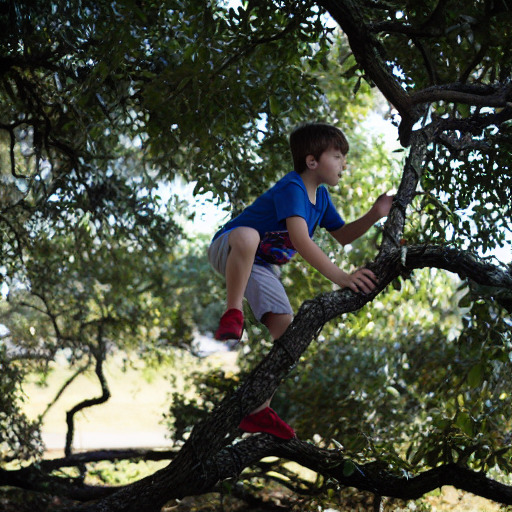}}&
{\includegraphics[width=0.16\textwidth]{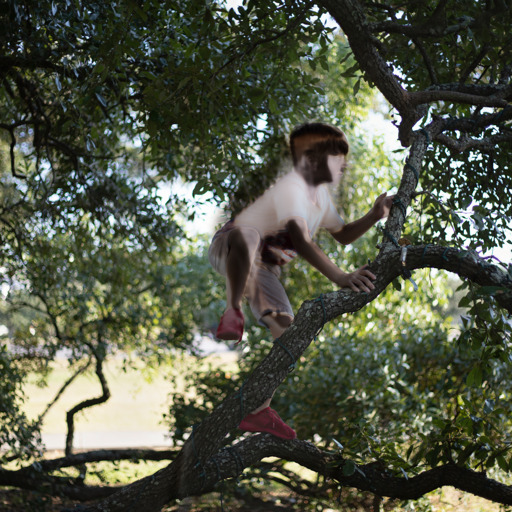}} &
{\includegraphics[width=0.16\textwidth]{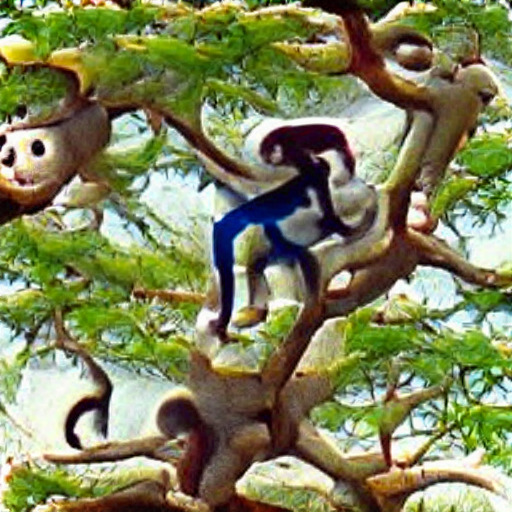}} &
{\includegraphics[width=0.16\textwidth]{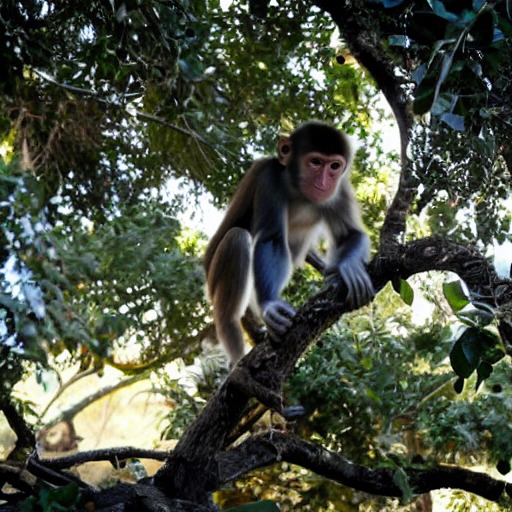}} &
{\includegraphics[width=0.16\textwidth]{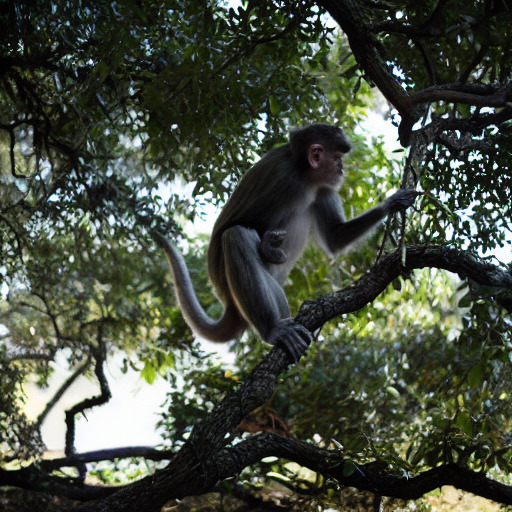}} \\

 & \multicolumn{5}{c}{"A \st{child} {\color{RoyalPurple} \bf monkey} is climbing on a tree"} \\
\\

% {\includegraphics[width=0.16\textwidth]{images/gt/cake_014.jpg}}&
% {\includegraphics[width=0.16\textwidth]{images/text2live/cake_009.jpg}} &
% {\includegraphics[width=0.16\textwidth]{images/vqclip/009.jpg}} &
% {\includegraphics[width=0.16\textwidth]{images/sde/cake_009.jpg}} &
% {\includegraphics[width=0.16\textwidth]{images/ours/cake_009.jpg}} \\

%  & \multicolumn{5}{c}{"A {\color{RoyalPurple} \bf brioche} cake on a table"} \\

% {\includegraphics[width=0.16\textwidth]{images/gt/cake_014.jpg}}&
% {\includegraphics[width=0.16\textwidth]{images/inversion/cake_t2l.jpg}}&
% {\includegraphics[width=0.16\textwidth]{images/text2live/cake_10.jpg}} &
% {\includegraphics[width=0.16\textwidth]{images/vqclip/010.jpg}} &
% {\includegraphics[width=0.16\textwidth]{images/sde/cake_010.jpg}} &
% {\includegraphics[width=0.16\textwidth]{images/ours/cake_010.jpg}} \\

%  & \multicolumn{5}{c}{"A {\color{RoyalPurple} \bf spinach moss} cake on a table"} \\

% {\includegraphics[width=0.16\textwidth]{images/gt/cake_014.jpg}}&
% {\includegraphics[width=0.16\textwidth]{images/inversion/cake_t2l.jpg}}&
% {\includegraphics[width=0.16\textwidth]{images/text2live/cake_15.jpg}} &
% {\includegraphics[width=0.16\textwidth]{images/vqclip/015.jpg}} &
% {\includegraphics[width=0.16\textwidth]{images/sde/cake_015.jpg}} &
% {\includegraphics[width=0.16\textwidth]{images/ours/cake_015.jpg}} \\

%  & \multicolumn{5}{c}{"A {\color{RoyalPurple} \bf birthday} cake on a table"} \\

{\includegraphics[width=0.16\textwidth]{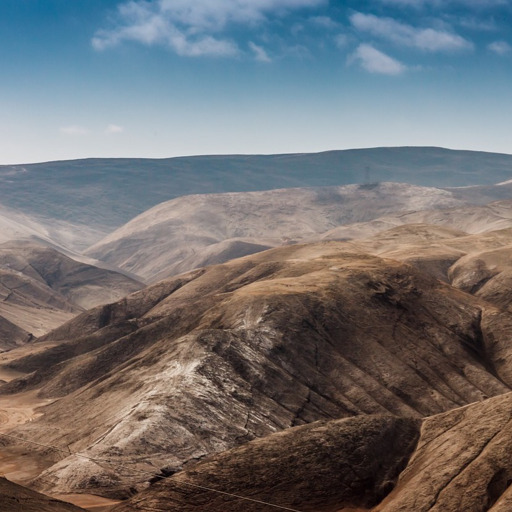}}&
{\includegraphics[width=0.16\textwidth]{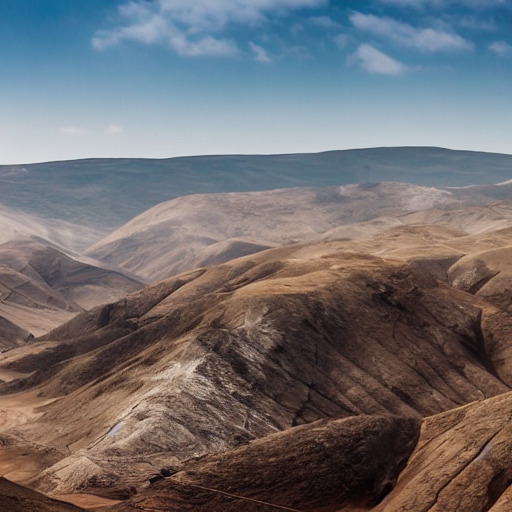}}&
{\includegraphics[width=0.16\textwidth]{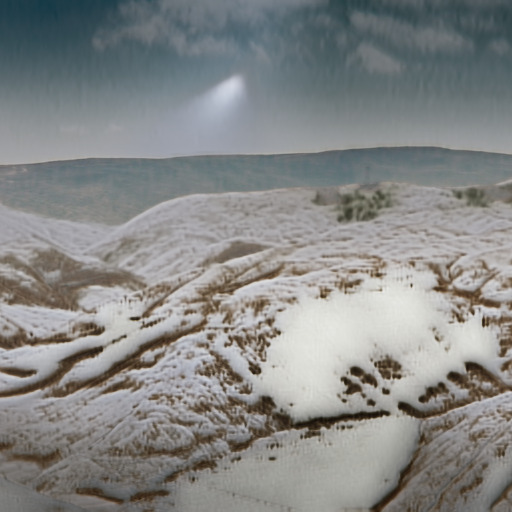}} &
{\includegraphics[width=0.16\textwidth]{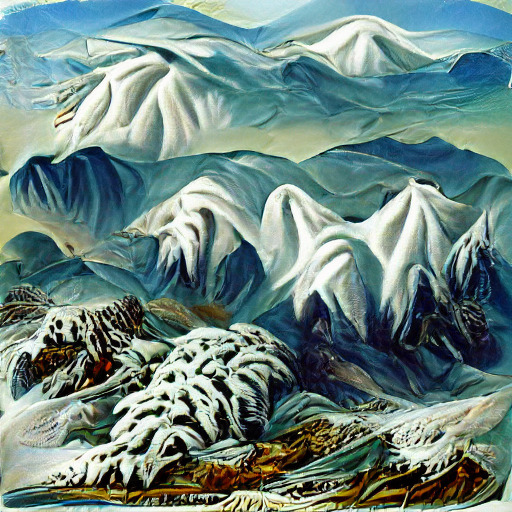}} &
{\includegraphics[width=0.16\textwidth]{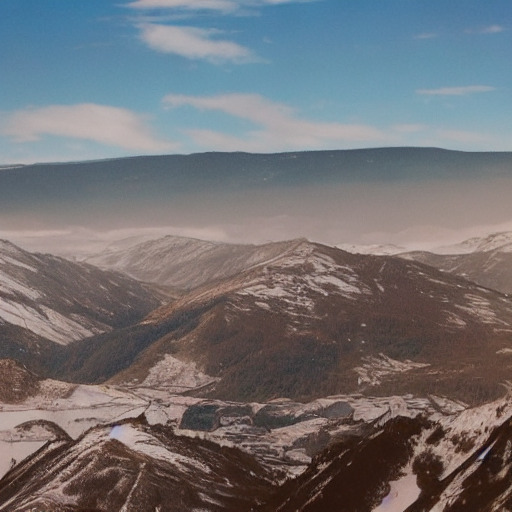}} &
{\includegraphics[width=0.16\textwidth]{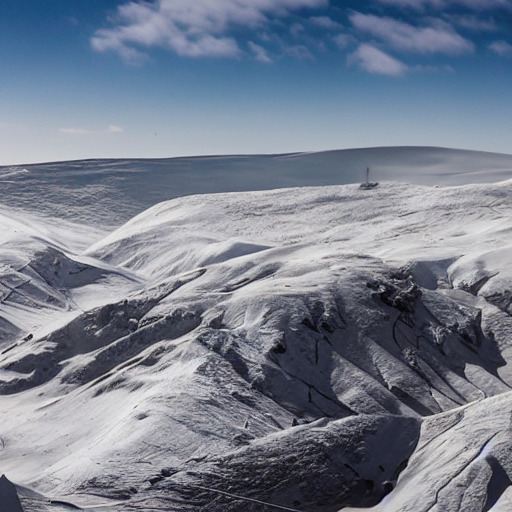}} \\

 & \multicolumn{5}{c}{"A {Landscape of \color{RoyalPurple} \bf Snowy} mountains"} \\
\\

{\includegraphics[width=0.16\textwidth]{images/gt/022.jpg}}&
{\includegraphics[width=0.16\textwidth]{images/inversion/mountain_t2l.jpg}}&
{\includegraphics[width=0.16\textwidth]{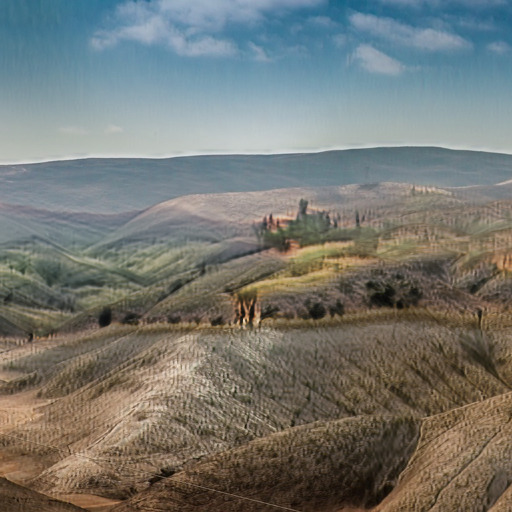}} &
{\includegraphics[width=0.16\textwidth]{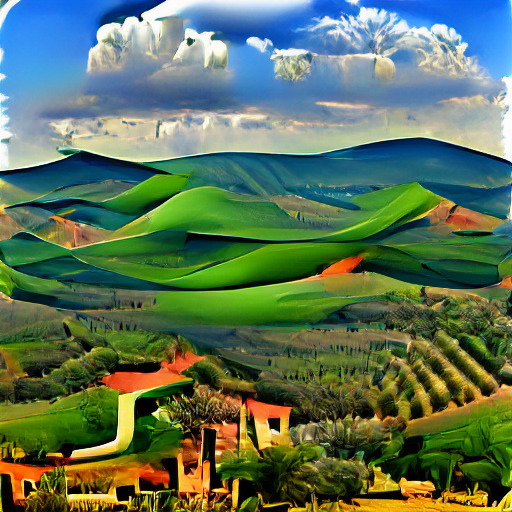}} &
{\includegraphics[width=0.16\textwidth]{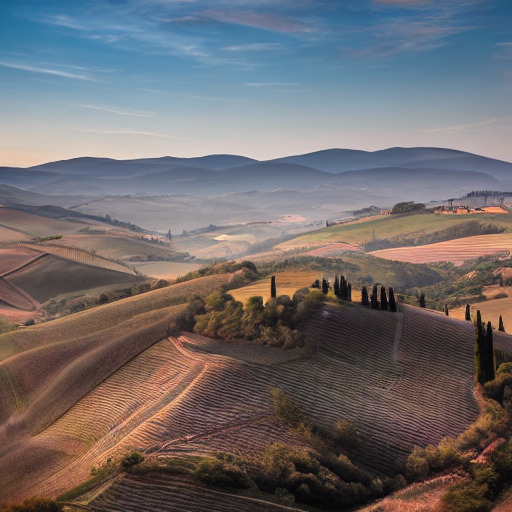}} &
{\includegraphics[width=0.16\textwidth]{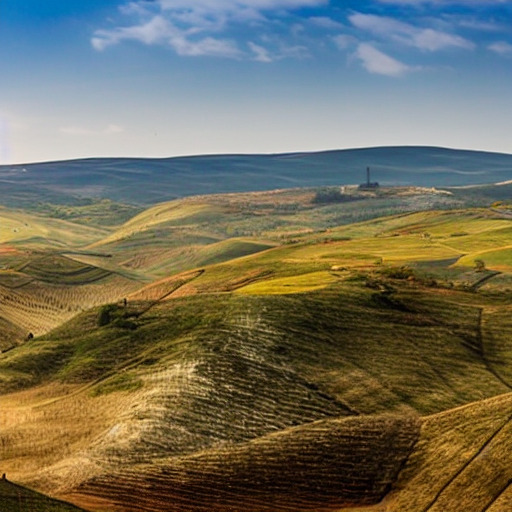}} \\

 & \multicolumn{5}{c}{"A {Landscape of \st{mountains} \color{RoyalPurple} \bf Tuscany}"} \\

{\includegraphics[width=0.16\textwidth]{images/gt/153.jpg}}&
{\includegraphics[width=0.16\textwidth]{images/inversion/6336601581_f53b8b1e86_z.jpg}}&
{\includegraphics[width=0.16\textwidth]{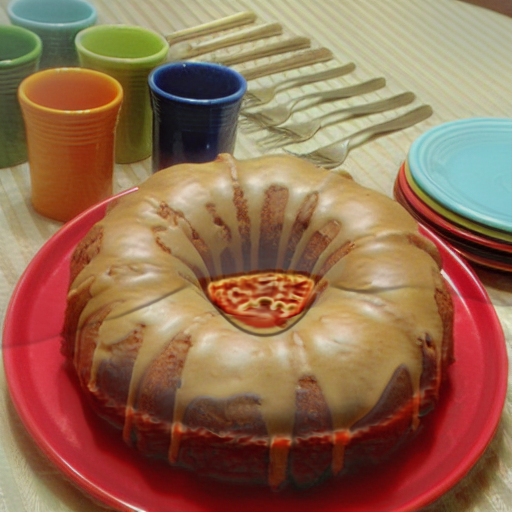}} &
{\includegraphics[width=0.16\textwidth]{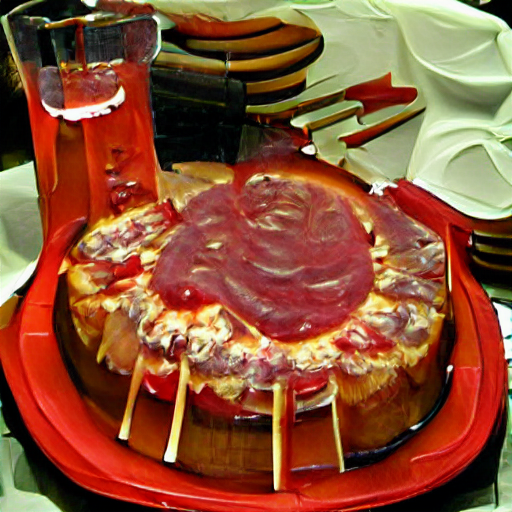}} &
{\includegraphics[width=0.16\textwidth]{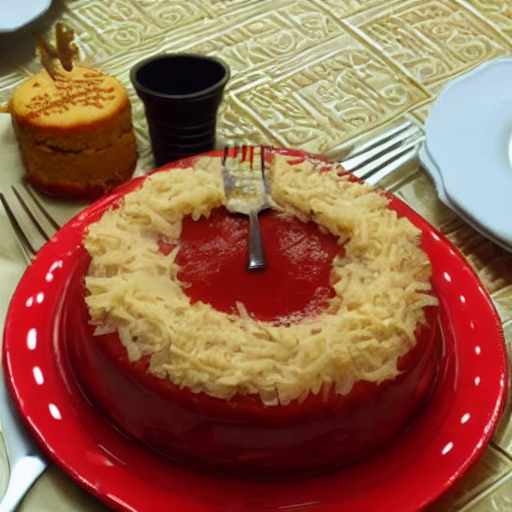}} &
{\includegraphics[width=0.16\textwidth]{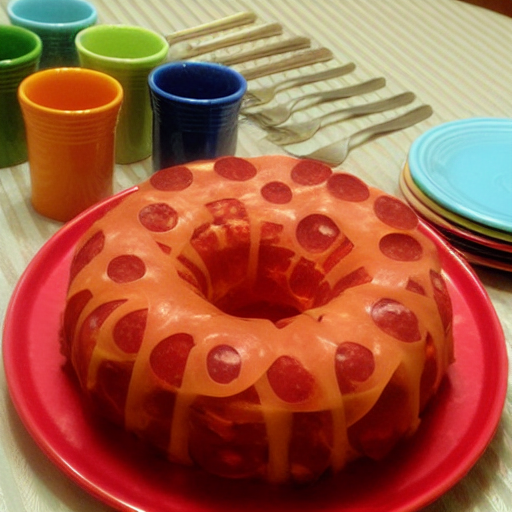}} \\

 & \multicolumn{5}{c}{""A {\color{RoyalPurple} \bf pepperoni} cake on a table""}

\end{tabular}
}

\vspace{-0.2cm}
\caption{{\bf Additional comparison results.}}
\vspace{-0.5cm}
% \vspace{-0.15cm}
% \ron{The gap is best viewed in the supplementary videos.}}
\label{fig:supp_comparison} %\vspace{-7pt}
\end{figure*}

Additional editing results of our method are provided in \cref{fig:supp_ours} and additional comparisons are provided in \cref{fig:supp_comparison}.

\paragraph{Inference time comparison.} As can be seen in \cref{tab:times}, SDEdit is the fastest since an inversion is not employed, but as a result, it fails to preserve the details of the original image. Our method is more efficient than Text2Live \cite{bar2022text2live}, VQGAN+CLIP \cite{crowson2022vqgan} and Imagic \cite{Kawar2022ImagicTR}, as it provides an accurate reconstruction in $\sim 1$ minute, while also allowing multiple editing operations after a single inversion.

\paragraph{Comparison to Imagic}

\begin{figure}[t!]
\centering 
\vspace{-0.4cm}
\includegraphics[width=\columnwidth]{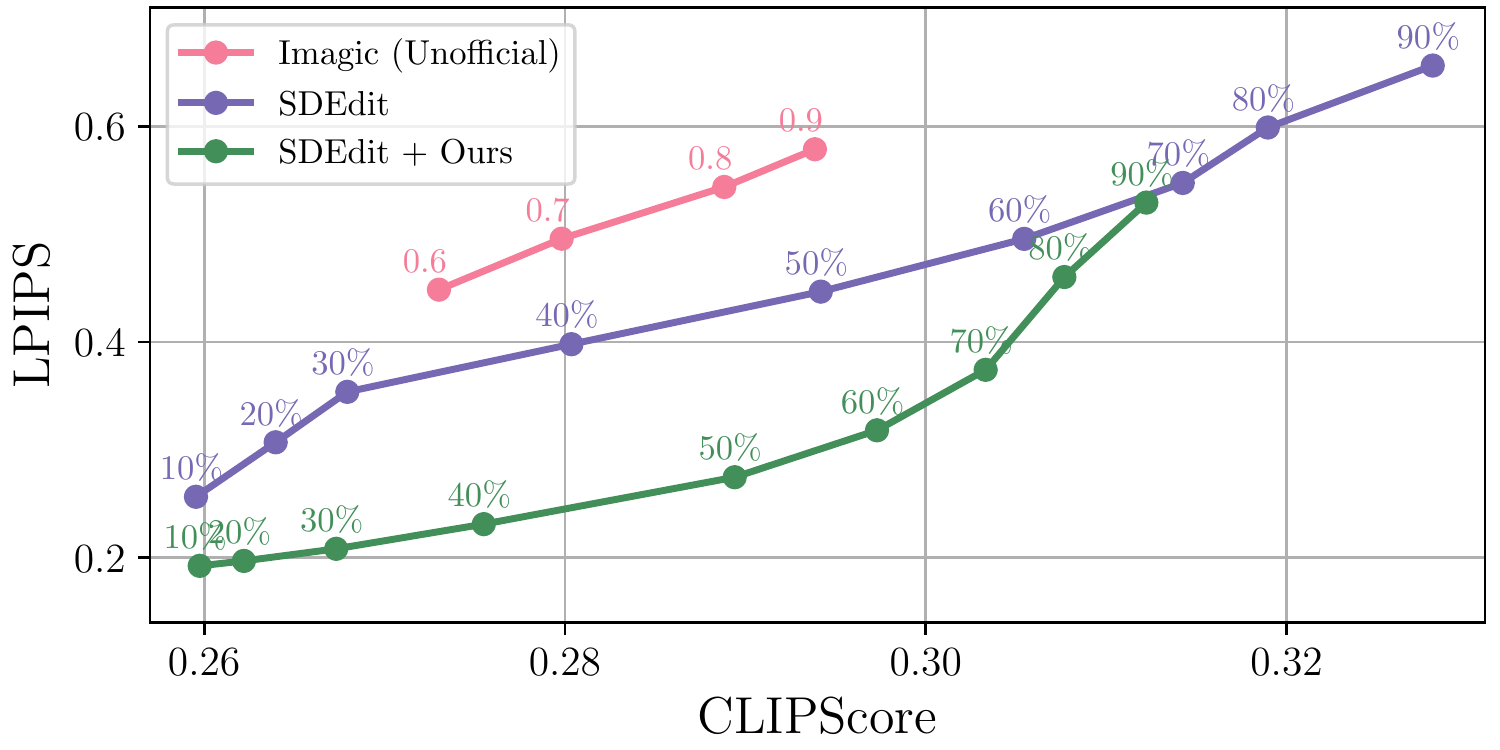} 
% \vspace{-0.3cm}
\caption{{\bf Comparison to Imagic} { \it We quantitatively evaluate Imagic using the unofficial implementation for Stable Diffusion. We measure both fidelity to the original image (via LPIPS, low is better) and fidelity to the target text (via CLIP, high is better). We use different values of the text embedding interpolation parameter $\alpha$, marked on the curve. The high LPIPS perceptual distance indicates that Imagic fails to retain high fidelity to the original image. }} 
\label{fig:exp_graph_imagic} 
\vspace{-0.3cm}
\end{figure}

\begin{figure*}
\setlength{\tabcolsep}{2.0pt}
    \centering
    { %\scriptsize %\footnotesize

\begin{tabular}{c}
% \begin{tabular}{p{0.2\columnwidth}  p{0.2\columnwidth} p{0.2\columnwidth} p{0.2\columnwidth} p{0.2\columnwidth}}
\vspace{0.15cm}
\begin{tabular}{>{\centering\arraybackslash}p{0.14\textwidth} >{\centering\arraybackslash}p{0.55\textwidth} >{\centering\arraybackslash}p{0.145\textwidth} >{\centering\arraybackslash}p{0.13\textwidth}}
{ \bf Input}&
{{ \bf Imagic - Stable Diffusion with $\alpha=0.6,0.7,0.8,0.9$}} &
{ \bf Imagic - Imagen}&
{ \bf Ours} 
\end{tabular}\\

 \begin{tabular}{c|c c c c|c| c}
 {\includegraphics[width=0.134\textwidth]{images/gt/135.jpg}}&
 {\includegraphics[ width=0.134\textwidth]{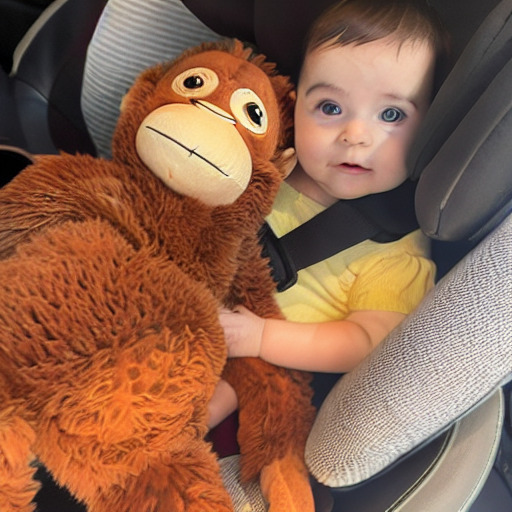}}&
 {\includegraphics[width=0.134\textwidth]{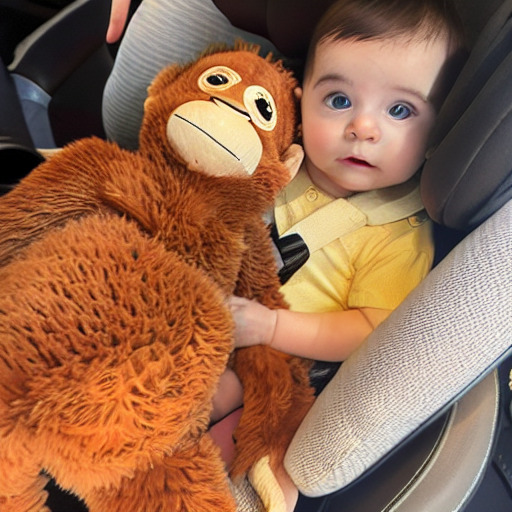}} &
 {\includegraphics[width=0.134\textwidth]{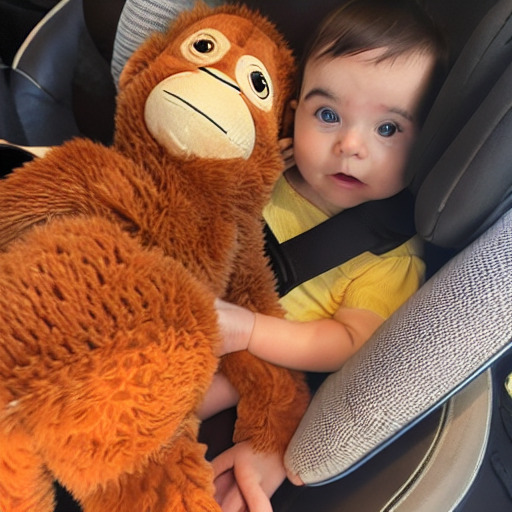}} &
 {\includegraphics[width=0.134\textwidth]{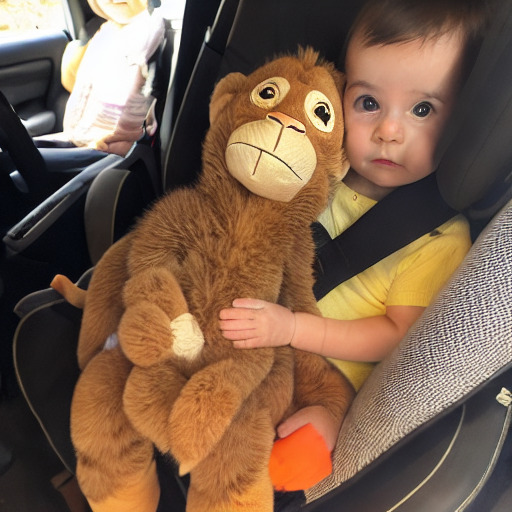}} &
 {\includegraphics[width=0.134\textwidth]{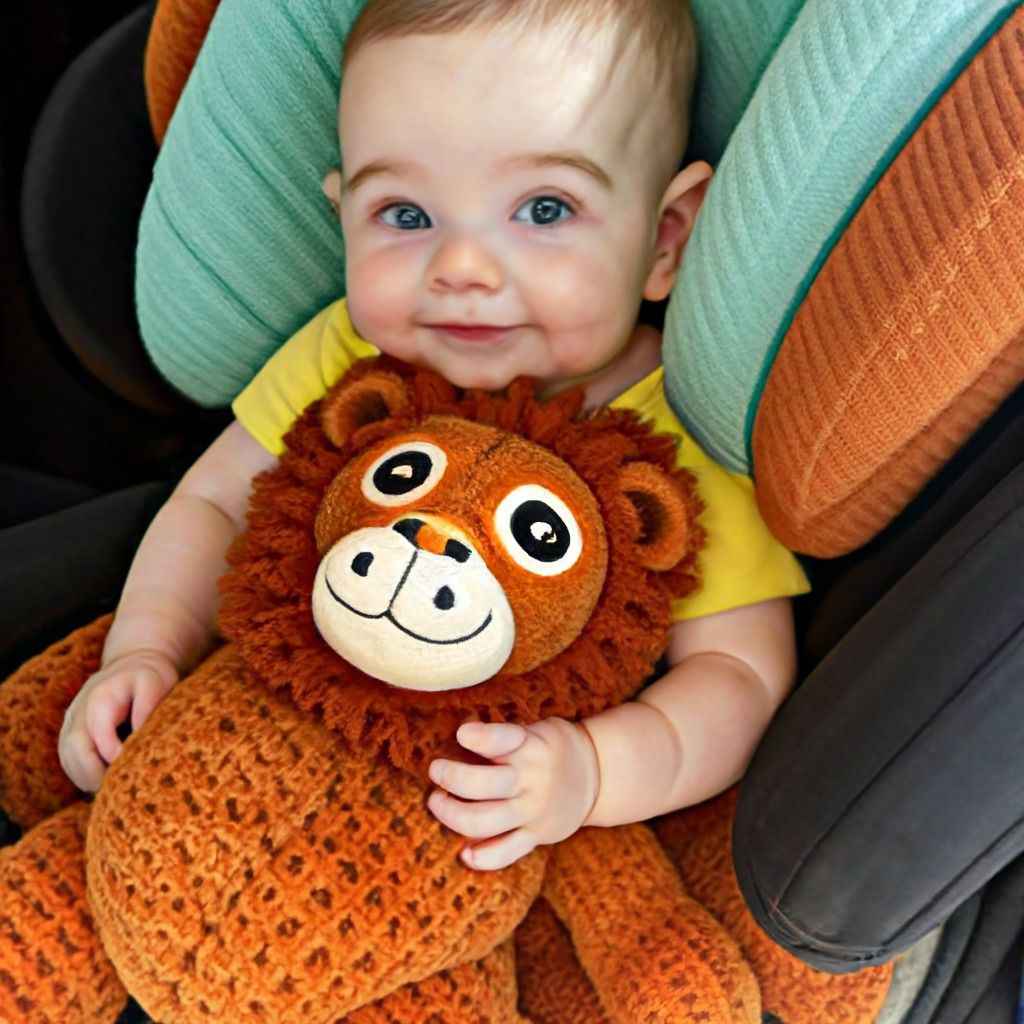}} &
{\includegraphics[width=0.134\textwidth]{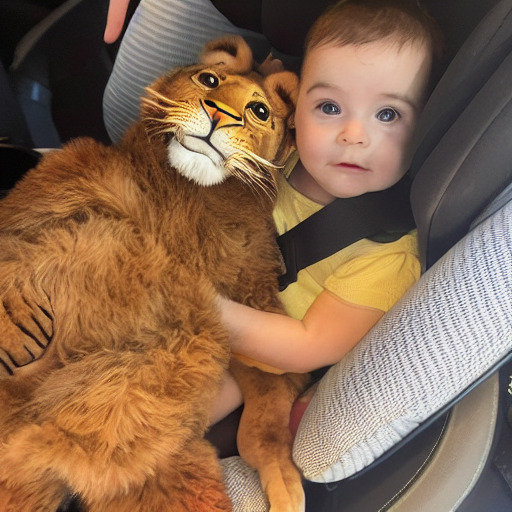}}
 \end{tabular}
  \\
  \vspace{0.3cm}
{"A baby holding her \st{monkey} {\color{RoyalPurple} \bf lion} doll"}\\
 \begin{tabular}{c|c c c c|c|c}
{\includegraphics[width=0.134\textwidth]{images/gt/153.jpg}}&
{\includegraphics[width=0.134\textwidth]{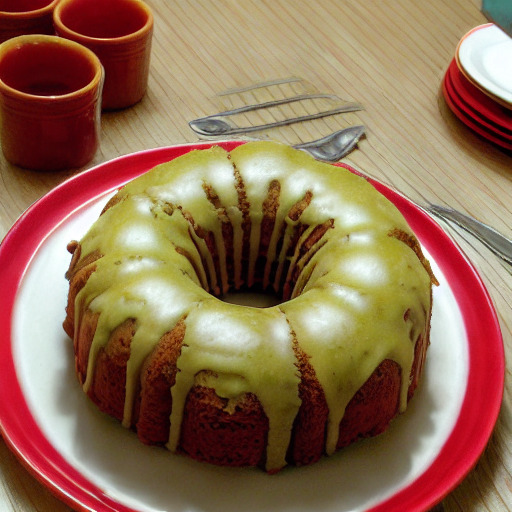}}&
{\includegraphics[width=0.134\textwidth]{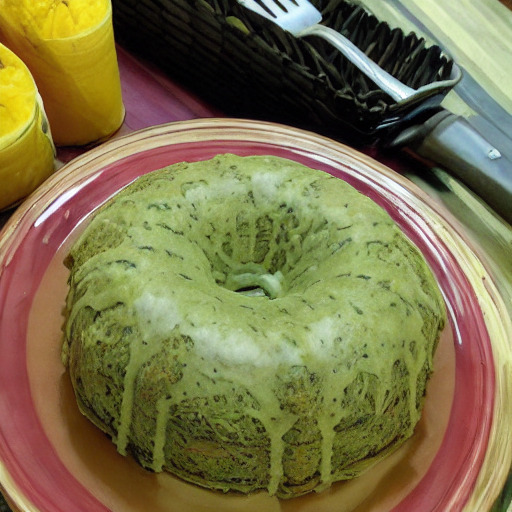}} &
{\includegraphics[width=0.134\textwidth]{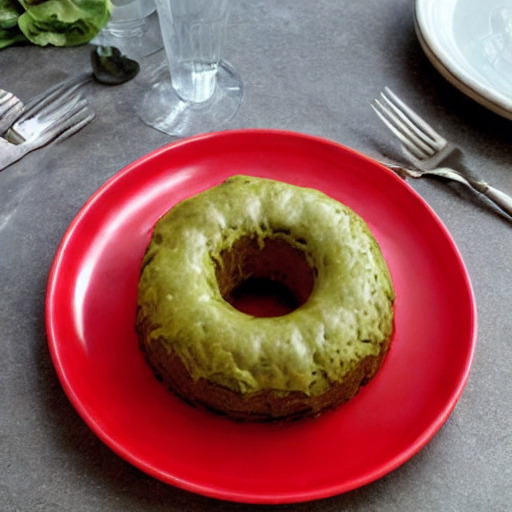}} &
{\includegraphics[width=0.134\textwidth]{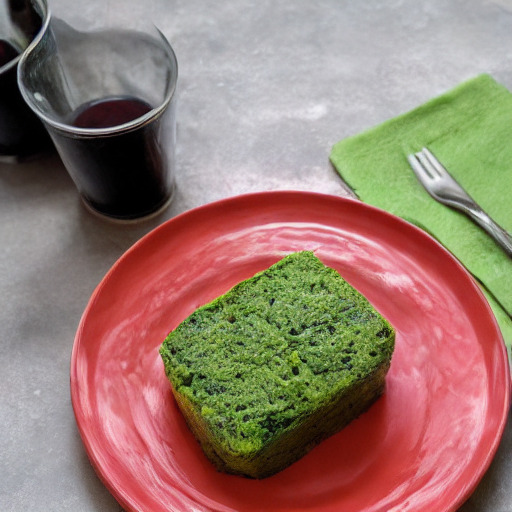}} &
{\includegraphics[width=0.134\textwidth]{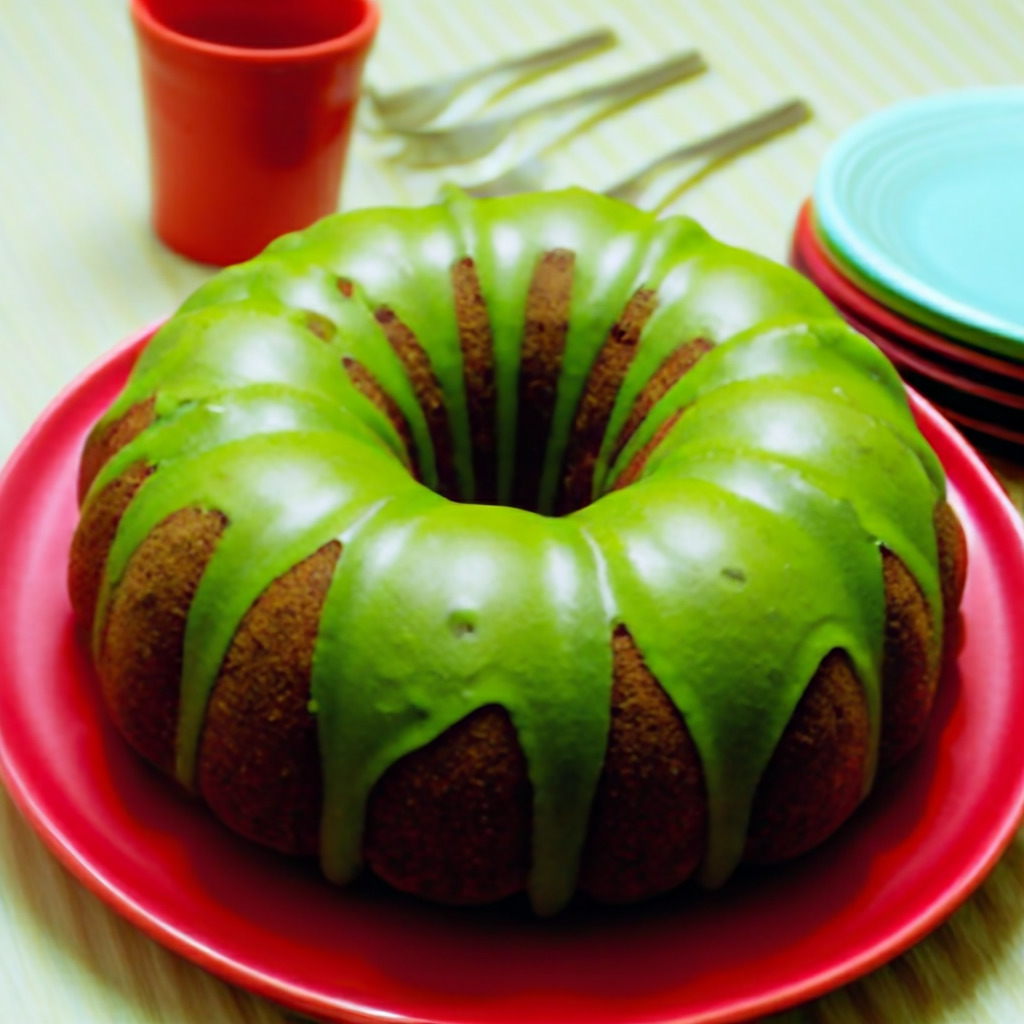}} &
{\includegraphics[width=0.134\textwidth]{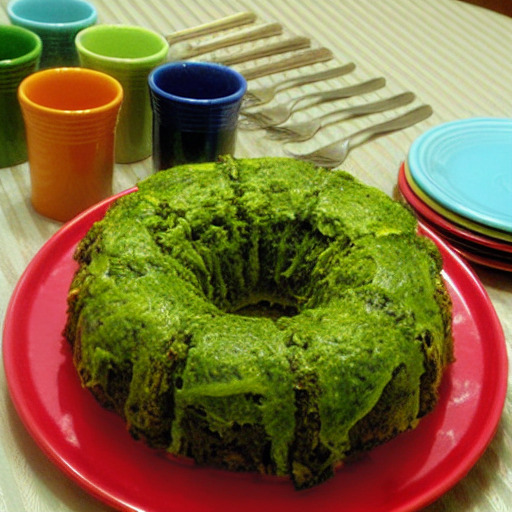}} 
\end{tabular}
\\

\vspace{0.3cm}
 {"A {\color{RoyalPurple} \bf spinach moss} cake on a table"} \\

\begin{tabular}{c|c c c c|c|c}
{\includegraphics[width=0.134\textwidth]{images/gt/159.jpg}}&
{\includegraphics[width=0.134\textwidth]{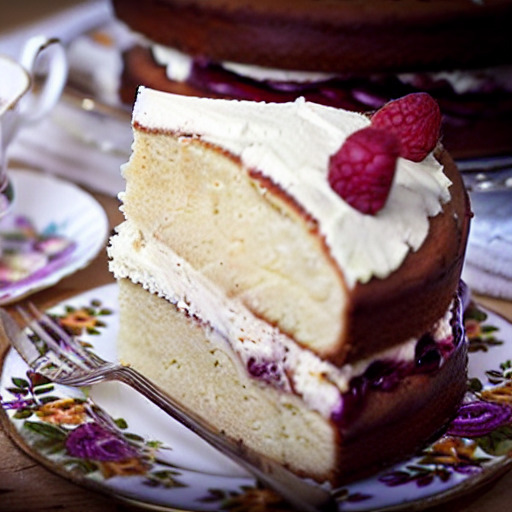}}&
{\includegraphics[width=0.134\textwidth]{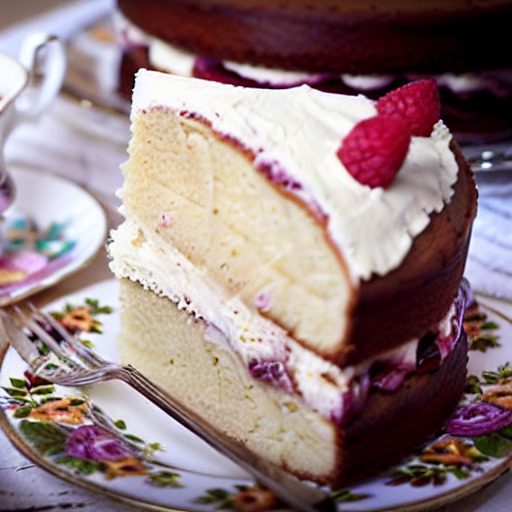}} &
{\includegraphics[width=0.134\textwidth]{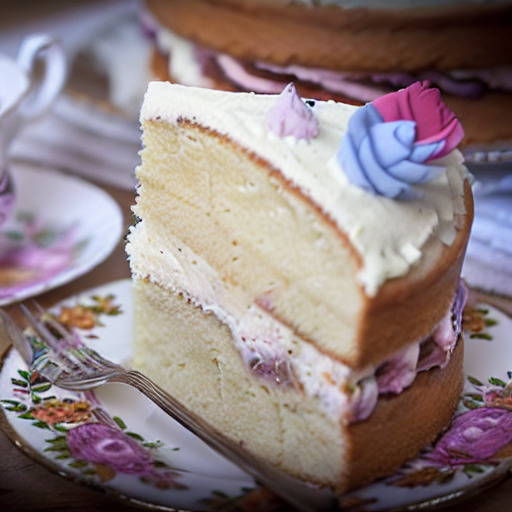}} &
{\includegraphics[width=0.134\textwidth]{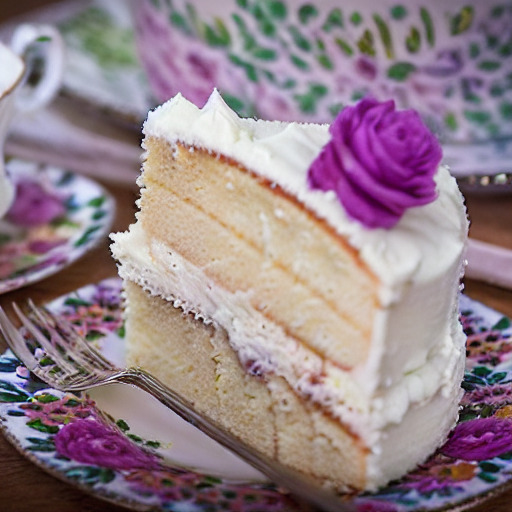}} &
{\includegraphics[width=0.134\textwidth]{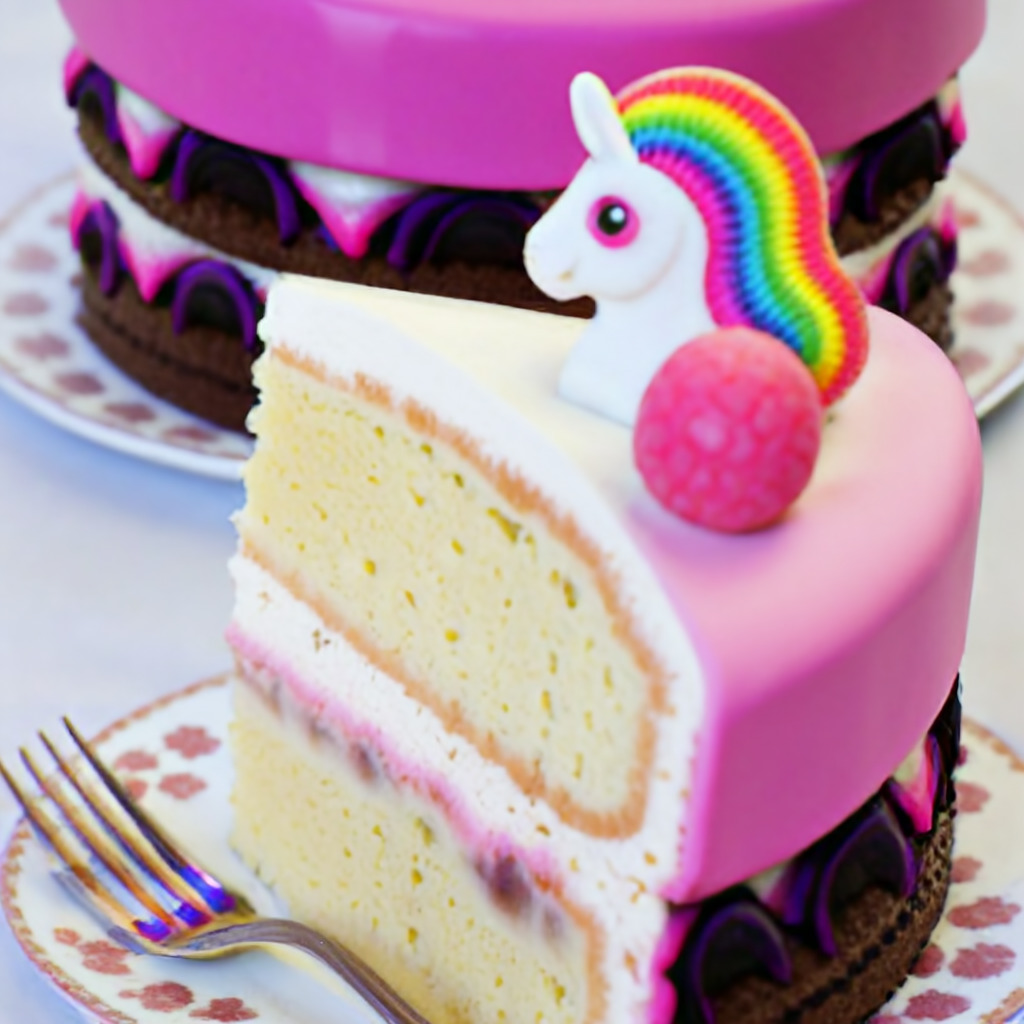}} &
{\includegraphics[width=0.134\textwidth]{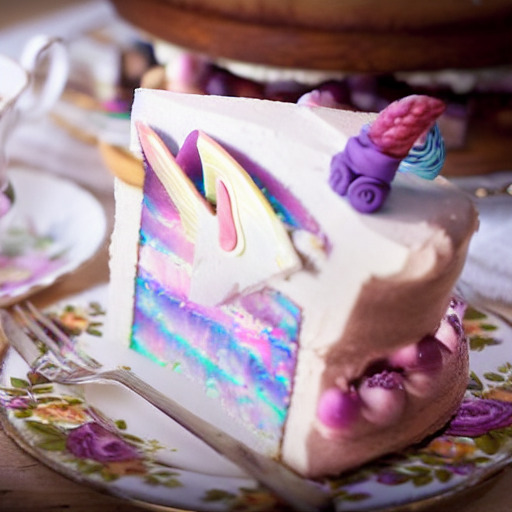}} \end{tabular}
\\
{"A piece of {\color{RoyalPurple} \bf unicorn} cake"}

\end{tabular}
}

\vspace{-0.2cm}
\caption{{\bf Comparison to Imagic \cite{Kawar2022ImagicTR}. {\it We first employ the unofficial Imagic implementation for Stable Diffusion and present the results for different values of the interpolation parameter $\alpha = 0.6,0.7,0.8,0.9$ (left to right). %This parameter is used to interpolate between the target text embedding and the optimized one \cite{Kawar2022ImagicTR}, where a larger value of $\alpha$ increases the fidelity to the target text. 
In addition, Imagic authors applied their method using the Imagen model over the same images, using the parameters $\alpha=0.93, 0.86, 1.08$ (from top to bottom row).
As can be seen, Imagic produces highly meaningful editing, especially when the Imagen model is involved.
However, Imagic struggles to preserve the original details, such as the identity of the baby ($1$st row) or cups in the background ($2$nd row). Furthermore, we observe that each example requires a separate tuning of the $\alpha$ parameter. Lastly, recall that each Imagic editing requires a separate tuning of the model.} }}
\vspace{-0.5cm}
% \vspace{-0.15cm}
% \ron{The gap is best viewed in the supplementary videos.}}
\label{fig:imagic} %\vspace{-7pt}
\end{figure*}

Quantitative comparison to Imagic is presented in \cref{fig:exp_graph_imagic}, using the unofficial Stable Diffusion implementation. According to these measures, our method achieves better preservation of the original details (lower LPIPS). This is also supported by the visual results in \cref{fig:imagic}, as Imagic struggles to accurately retain the background. Furthermore, we observe that Imagic is quite sensitive to the interpolation parameter $\alpha$, as a high value reduces the fidelity to the image and a low value reduces the fidelity to the text, while a single value cannot be applied to all examples.
In addition, the authors of Imagic applied their method on the same three images, presented in \cref{fig:imagic}, using $\alpha=0.93, 0.86, 1.08$. This results in much better quality, however, still the background is not preserved. %, the model is sensitive to $\alpha$, and fine-tuning per editing operation is required.

\section{Implementation details}
\label{sup_implementation}

In all of our experiments, we employ the Stable Diffusion \cite{rombach2021highresolution} using a DDIM sampler with the default hyperparameters: number of diffusion steps $T=50$  and guidance scale $w=7.5$.
Stable diffusion utilizes a pre-trained CLIP network as the language model $\psi$. The null-text is tokenized into \textit{start-token}, \textit{end-token}, and $75$ non-text padding tokens. Notice that the padding tokens are also used in CLIP and the diffusion model since both models do not use masking. 
%In addition, the paddings have different representations because of the added positional embedding.

All inversion results except the ones in the ablation study were obtained using $N=10$ (See Algorithm 1 in the main paper) and a learning rate of $0.01$. We have used an early stop parameter of $\epsilon=1e-5$ such that the total inversion for an input image and caption took $40s - 120s$ on a single A100 GPU. Namely, for each timestamp $t$, we stop the optimization when the loss function value reaches $\epsilon=1e-5$.

\paragraph{Baseline Implementations.}
For the comparisons in section 5, we use the official implementation of Text2Live\footnote{https://github.com/omerbt/Text2LIVE}~\cite{bar2022text2live} and  VQGAN+CLIP\footnote{https://github.com/nerdyrodent/VQGAN-CLIP}~\cite{katherine2021vqganclip}.
We have implemented the SDEdit \cite{meng2021sdedit} method over Stable Diffusion based on the official implementation\footnote{https://github.com/ermongroup/SDEdit}.
We also compare our method to Imagic \cite{Kawar2022ImagicTR} using an unofficial implementation\footnote{https://github.com/ShivamShrirao/diffusers/tree/main/examples/imagic} (see \cref{sec:supp_results}).

\paragraph{Global null-text Inversion.}
The algorithm for optimizing only a single null-text embedding $\varnothing$ for all timestamps is presented in algorithm \ref{alg:null_global}.
In this case, since the optimization of $\varnothing$ in a single timestamp affects all other timestamps, we change the order of the iterations in Algorithm 1. That is, we perform $N$ iterations in each we optimize $\varnothing$ for all the diffusion timestamps by iterating over $t$. As shown in Section 4, the convergence of this optimization is much slower than our final method. More specifically, we found that only after $~7500$ optimization steps (about 30 minutes) the global null-text inversion accurately reconstruct the input image.

\setcounter{algocf}{1}
\begin{algorithm}

\SetAlgoLined
\textbf{Input:} A source prompt $\mathcal{P}$ and input image $\mathcal{I}$.\\ %intermediate results of DDIM inversion $z^*_T,\ldots,z^*_0$ (using $w=1$).\\
\textbf{Output:} Noise vector $z_T$ and an optimized embedding $\varnothing$ .\\
 %Let $(s_{Normal},s_T,s_{T-1},\ldots,s_1)$ be a sequence of random seeds\; 
 \vspace{1mm} \hrule \vspace{1mm}
 Set guidance scale $w=1$; \\
 Compute the intermediate results  $z^*_T,\ldots,z^*_0$ of DDIM inversion for image $\mathcal{I}$; \\
 Set guidance scale $w=7.5$; \\
 Initialize $\varnothing \leftarrow  \psi("")$; \\
 \For{$j=0,\ldots,N-1$}{
 Set $\bar{z_T} \leftarrow z^*_T$; \\
 \For{$t=T,T-1,\ldots,1$}{
        $\varnothing \leftarrow  \varnothing - \eta\nabla_{\varnothing} \norm{z^*_{t-1}-z_{t-1}(\bar{z_t}, \varnothing, \mathcal{C})}^2$;
        Set $\bar{z}_{t-1} \leftarrow z_{t-1}(\bar{z_t}, \varnothing, \mathcal{C})$;
    }
 }
 \textbf{Return} $\bar{z_T}$, $\varnothing$
 
 \caption{Global null-text inversion}

\label{alg:null_global}
\end{algorithm}

\vspace{-0.2cm}
\section{Additional Background - Diffusion Models}\label{sec:supp_background}

Diffusion Denoising Probabilistic Models (DDPM)~\cite{sohl2015deep,ho2020denoising} are generative latent variable models that aim to model a distribution $p_\theta(x_0)$ that approximates the data distribution $q(x_0)$ and easy to sample from. DDPMs model a ``forward process'' in the space of $x_0$ from data to noise. This is called ``forward'' due to its procedure progressing from $x_0$ to $x_T$. Note that this process is a Markov chain starting from $x_0$, where we gradually add noise to the data to generate the latent variables $x_1,\ldots,x_T\in X$. The sequence of latent variables, therefore, follows $q(x_1,\ldots,x_t\mid x_0)=\prod_{i=1}^{t}q(x_t\mid x_{t-1})$, where a step in the forward process is defined as a Gaussian transition $q(x_t\mid x_{t-1}):=N(x_t;\sqrt{1-\beta_t}x_{t-1},\beta_t I)$ parameterized by a schedule $\beta_0,\ldots,\beta_T\in (0,1)$. When $T$ is large enough, the last noise vector $x_T$ nearly follows an isotropic Gaussian distribution.

An interesting property of the forward process is that one can express the latent variable $x_t$ directly as the following linear combination of noise and $x_0$ without sampling intermediate latent vectors: \\[-8pt]
\begin{equation}
  x_t = \sqrt{\alpha_t}x_0+\sqrt{1-\alpha_t}w,~~w\sim N(0,I),\label{eq:xtsamplefromx0}  
\end{equation} \\[-10pt]
where $\alpha_t:=\prod_{i=1}^{t}(1-\beta_i)$.

To sample from the distribution $q(x_0)$, we define the dual ``reverse process'' $p(x_{t-1}\mid x_t)$ from isotropic Gaussian noise $x_T$ to data by sampling the posteriors $q(x_{t-1} \mid x_t)$. Since the intractable reverse process $q(x_{t-1} \mid x_t)$ depends on the unknown data distribution $q(x_0)$, we approximate it with a parameterized Gaussian transition network $p_\theta(x_{t-1}\mid x_t):=N(x_{t-1}\mid \mu_\theta(x_t,t),\Sigma_\theta(x_t,t))$. The $\mu_\theta(x_t,t)$ can be replaced~\cite{ho2020denoising} by predicting the noise $\eps_\theta(x_t,t)$ added to $x_0$ using equation~\ref{eq:xtsamplefromx0}. 

We use Bayes' theorem to approximate \\[-8pt]
\begin{equation} 
\mu_\theta(x_t,t)=\frac{1}{\sqrt{\alpha_t}}\left(x_t-\frac{\beta_t}{\sqrt{1-\alpha_t}}\eps_\theta(x_t,t)\right).
\end{equation} \\[-5pt]
Once we have a trained $\eps_\theta(x_t,t)$, we can using the following sample method  \\[-8pt]
\begin{equation} 
x_{t-1} = \mu_\theta(x_t,t)+\sigma_t z,~~z\sim N(0,I).
\end{equation} \\[-12pt]
We can control $\sigma_t$ of each sample stage, and in DDIMs~\cite{song2020denoising} the sampling process can be made deterministic using $\sigma_t=0$ in all the steps. The reverse process can finally be trained by solving the following optimization problem: \\[-8pt]
$$\min_\theta L(\theta):=\min_\theta E_{x_0\sim q(x_0),w\sim N(0,I),t} \norm{w-\eps_\theta(x_t,t)}^2,$$ \\[-10pt]
teaching the parameters $\theta$ to fit $q(x_0)$ by maximizing a variational lower bound.

\section{User-Study}
\label{sup:user_study}
%\vspace{-0.2cm}

\begin{figure*}

% \begin{tabular}{c c}

% \includegraphics[width=\columnwidth]{sup_figures/user_study1.jpg} & 
% \includegraphics[width=\columnwidth]{sup_figures/user_study2.jpg} \\

% \end{tabular}
\centering
\includegraphics[width=\columnwidth]{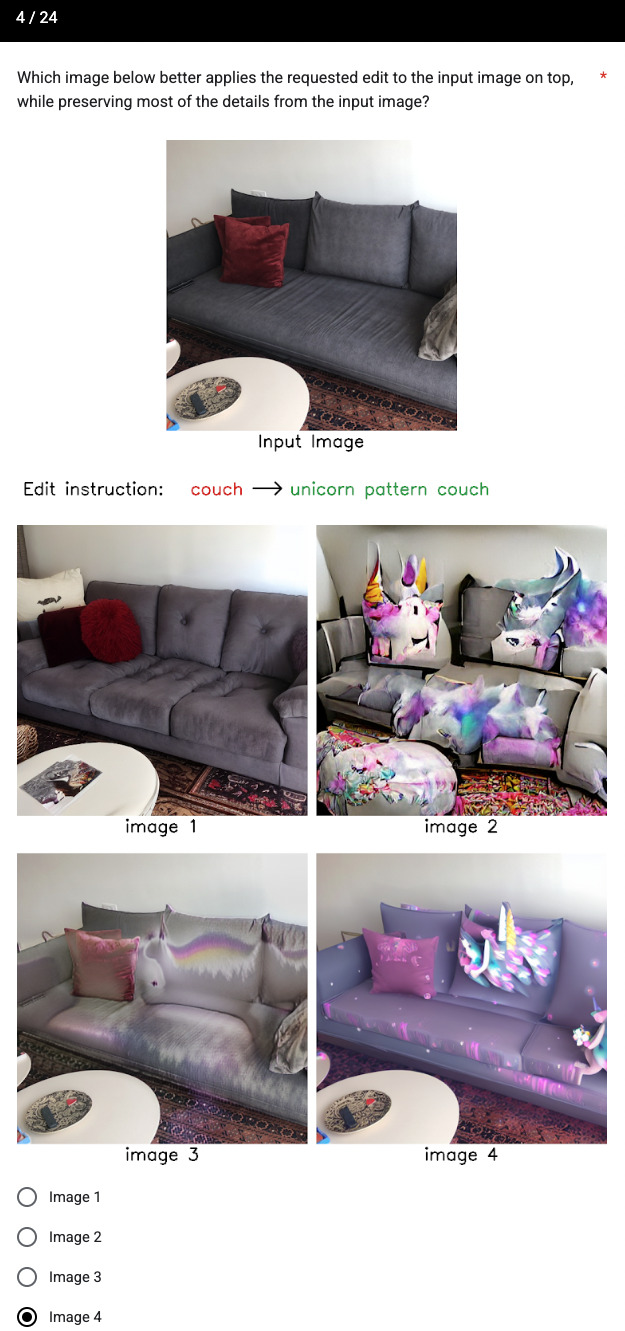}

\caption{{\bf User study print screen.}}

\label{fig:user_study} 
\end{figure*}

An illustration of our user study is provided in \cref{fig:user_study}

\section{Image Attribution}
%\vspace{-0.2cm}

{ \small
Girl in a field: ~\url{https://unsplash.com/photos/1pCpWipo_jM} \newline
Birds on a branch: ~\url{https://pixabay.com/photos/sparrows-birds-perched-sperlings-3434123/} \newline
Basket with apples:~\url{https://unsplash.com/photos/4Bj27zMqNSE} \newline
Bicycle: ~\url{https://unsplash.com/photos/vZAk_n9Plfc}\newline
Child climbing:~\url{https://unsplash.com/photos/oLZViCDG-dk} \newline
Mountains:~\url{https://pixabay.com/photos/desert-mountains-sky-clouds-peru-4842264/} \newline
Giraffe:~\url{https://www.flickr.com/photos/tambako/30850708538/} \newline
Blue-haired woman in the forest: ~\url{https://unsplash.com/photos/I3oRtzyBIFg} \newline
Dining table:~\url{https://cocodataset.org/#explore?id=360849} \newline
Elephants:~\url{https://cocodataset.org/#explore?id=345520} \newline
Man with a doughnut:~\url{https://cocodataset.org/#explore?id=360849} \newline
Cake on a table:~\url{https://cocodataset.org/#explore?id=413699} \newline
Piece of cake:~\url{https://cocodataset.org/#explore?id=133063} \newline
}


\end{document}